\documentclass[journal]{IEEEtran}
\DeclareUnicodeCharacter{2009}{\,}%
\usepackage{graphicx}
\usepackage{tikz}
\usetikzlibrary{arrows.meta, positioning}
\usepackage[utf8]{inputenc}
\usepackage{booktabs}
\usetikzlibrary{mindmap, shadows}
\usepackage{lipsum}  
\usepackage{amsmath,amssymb}
\usepackage{ragged2e}
\usepackage{tabularx}
\usepackage{adjustbox}
\usepackage{caption}
\usepackage{amssymb}    
\usepackage{pifont}     
\usepackage{xcolor}     
\newcommand*\emptycirc[1][1ex]{\tikz\draw (0,0) circle (#1);} 
\newcommand*\halfcirc[1][1ex]{%
  \begin{tikzpicture}
  \draw[fill] (0,0)-- (90:#1) arc (90:270:#1) -- cycle ;
  \draw (0,0) circle (#1);
  \end{tikzpicture}}
\newcommand*\fullcirc[1][1ex]{\tikz\fill (0,0) circle (#1);} 

\usepackage[
  colorlinks=false,
  linkbordercolor={0 1 0},   
  citebordercolor={0 1 0},
  urlbordercolor={0 1 0}
]{hyperref}

\ifCLASSINFOpdf
\else
\fi
\begin{document}
%
\title{From LLM Reasoning to Autonomous AI Agents: A Comprehensive Review }

\author{
\IEEEauthorblockN{Mohamed~Amine~Ferrag\IEEEauthorrefmark{1}\IEEEauthorrefmark{5}, Norbert Tihanyi\IEEEauthorrefmark{2}\IEEEauthorrefmark{3}, and  Merouane Debbah\IEEEauthorrefmark{4}}

\IEEEauthorblockA{\IEEEauthorrefmark{1}
Department of Computer and Network Engineering,  
United Arab Emirates University, UAE
} \\
\IEEEauthorblockA{\IEEEauthorrefmark{2}Technology Innovation Institute, UAE} \\
\IEEEauthorblockA{\IEEEauthorrefmark{3}Eötvös Loránd University, Hungary } \\
\IEEEauthorblockA{\IEEEauthorrefmark{4}Research Institute for Digital Future, Khalifa University, UAE} \\
\IEEEauthorblockA{\IEEEauthorrefmark{5}Corresponding author:mohamed.ferrag@uaeu.ac.ae}
 }


%
%

\markboth{ }%
{Shell \MakeLowercase{\textit{et al.}}: Bare Demo of IEEEtran.cls for IEEE Journals}
%



\maketitle


\begin{abstract}
Large language models and autonomous AI agents have evolved rapidly, resulting in a diverse array of evaluation benchmarks, frameworks, and collaboration protocols. Driven by the growing need for standardized evaluation and integration, we systematically consolidate these fragmented efforts into a unified framework. However, the landscape remains fragmented and lacks a unified taxonomy or comprehensive survey. Therefore, we present a side-by-side comparison of benchmarks developed between 2019 and 2025 that evaluate these models and agents across multiple domains. In addition, we propose a taxonomy of approximately 60 benchmarks that cover general and academic knowledge reasoning, mathematical problem-solving, code generation and software engineering, factual grounding and retrieval, domain-specific evaluations, multimodal and embodied tasks, task orchestration, and interactive assessments. Furthermore, we review AI-agent frameworks introduced between 2023 and 2025 that integrate large language models with modular toolkits to enable autonomous decision-making and multi-step reasoning. Moreover, we present real-world applications of autonomous AI agents in materials science, biomedical research, academic ideation, software engineering, synthetic data generation, chemical reasoning, mathematical problem-solving, geographic information systems, multimedia, healthcare, and finance. We then survey key agent-to-agent collaboration protocols, namely the Agent Communication Protocol (ACP), the Model Context Protocol (MCP), and the Agent-to-Agent Protocol (A2A). Finally, we discuss recommendations for future research, focusing on advanced reasoning strategies, failure modes in multi-agent LLM systems, automated scientific discovery, dynamic tool integration via reinforcement learning, integrated search capabilities, and security vulnerabilities in agent protocols.
\end{abstract}


%
\IEEEpeerreviewmaketitle

\section{Introduction}

Large Language Models (LLMs) such as OpenAI’s GPT-4 \cite{jaech2024openai}, Qwen2.5-Omni \cite{xu2025qwen25omnitechnicalreport}, DeepSeek-R1 \cite{guo2025deepseek}, and Meta’s LLaMA \cite{grattafiori2024llama} have transformed AI by enabling human-like text generation and advanced natural language processing, spurring innovation in conversational agents, automated content creation, and real-time translation \cite{huang2024understanding}. Recent enhancements have extended their utility to multimodal tasks, including text-to-image and text-to-video generation that broaden the scope of generative AI applications \cite{gu2024survey,bisztray2025know}. However, their dependence on static pre-training data can lead to outdated outputs and hallucinated responses \cite{wang2025vidorag,li2024benchmarking}, a limitation that Retrieval-Augmented Generation (RAG) addresses by incorporating real-time data from knowledge bases, APIs, or the web \cite{yu2025rag,ateia2024bioragent}. Building on this, the evolution of intelligent agents employing reflection, planning, and multi-agent collaboration has given rise to Agentic RAG systems, which dynamically orchestrate information retrieval and iterative refinement to manage complex workflows effectively \cite{shimadzu2025retrieval,xiong2025rag}.

Recent advances in large language models have paved the way for highly autonomous AI systems that can independently handle complex research tasks. These systems, often referred to as agentic AI, can generate hypotheses, conduct literature reviews, design experiments, analyze data, accelerate scientific discovery, and reduce research costs \cite{ferrag2025reasoninglimitsadvancesopen, achiam2023gpt, team2023gemini, touvron2023llama}. Several frameworks, such as LitSearch, ResearchArena, and Agent Laboratory, have been developed to automate various research tasks, including citation management and academic survey generation \cite{schmidgall2025agent, ajith2024litsearch,kang2024researcharena}. However, challenges persist, especially in executing domain-specific literature reviews and ensuring the reproducibility and reliability of automated processes \cite{baek2024researchagent,gridach2025agentic}. Parallel to these developments in research automation, large language model-based agents have also begun to transform the medical field \cite{kim2024mdagents}. These agents are increasingly used for diagnostic support, patient communication, and medical education by integrating clinical guidelines, medical knowledge bases, and healthcare systems \cite{chen2024agent, li2024agent, durante2024agent, zhu2024knowagent, he2024webvoyager}. Despite their promise, these applications face significant hurdles, including concerns over reliability, reproducibility, ethical governance, and safety \cite{mukherjee2024polaris,yuan2024r,yan2025application}. Addressing these issues is crucial for ensuring that LLM-based agents can be effectively and responsibly incorporated into clinical practice, underscoring the need for comprehensive evaluation frameworks that can reliably measure their performance across various healthcare tasks \cite{yu2024aipatient,schmidgall2024agentclinic,wang2025survey,ferrag2025prompt}.

LLM-based agents are emerging as a promising frontier in AI, combining reasoning and action to interact with complex digital environments \cite{ wang2024executable, shinn2023reflexion}. Therefore, various approaches have been explored to enhance LLM-based agents, from combining reasoning and acting using techniques like React \cite{yao2023react} and Monte Carlo Tree Search \cite{zhou2023language} to synthesizing high-quality data with methods like Learn-by-Interact \cite{su2025learn}, which sidestep assumptions such as state reversals. Other strategies involve training on human-labeled or GPT-4 distilled data with systems like AgentGen \cite{hu2024agentgen} and AgentTuning \cite{zeng2023agenttuning} to generate trajectory data. At the same time, reinforcement learning methods utilize offline algorithms and iterative refinement through reward models and feedback to enhance efficiency and performance in realistic environments \cite{gulcehre2023reinforced,aksitov2023rest}.

LLM-based Multi-Agents harness the collective intelligence of multiple specialized agents, enabling advanced capabilities over single-agent systems by simulating complex real-world environments through collaborative planning, discussion, and decision-making. This approach leverages the communicative strengths and domain-specific expertise of LLMs, allowing distinct agents to interact effectively, much like human teams tackling problem-solving tasks \cite{guo2024large,goldie2025synthetic}. Recent research highlights promising applications across various fields, including software development \cite{hong2023metagpt,qian2023communicative}, multi-robot systems \cite{mandi2024roco,zhang2023building}, society simulation \cite{park2023generative}, policy simulation \cite{xiao2023simulating}, and game simulation \cite{wang2023avalon}.

The main contributions of this study are:

\begin{itemize}
    \item We present a comparative table of benchmarks developed between 2019 and 2025 that rigorously evaluate large language models and autonomous AI agents across multiple domains.
    \item We propose a taxonomy of approximately 60 LLM and AI-agent benchmarks, including general and academic knowledge reasoning, mathematical problem solving, code generation and software engineering, factual grounding and retrieval, domain-specific evaluations, multimodal and embodied tasks, task orchestration, and interactive and agentic assessments.
    \item We present prominent AI-agent frameworks from 2023 to 2025 that integrate large language models with modular toolkits, enabling autonomous decision-making and multi-step reasoning.
    \item We provide applications of autonomous AI agents in various fields, including materials science and biomedical research, academic ideation and software engineering, synthetic data generation and chemical reasoning, mathematical problem-solving and geographic information systems, as well as multimedia, healthcare, and finance.
    \item We survey agent-to-agent collaboration protocols, namely the Agent Communication Protocol (ACP), the Model Context Protocol (MCP), and the Agent-to-Agent Protocol (A2A).
    \item We outline recommendations for future research on autonomous AI agents, specifically advanced reasoning strategies, failure modes in multi-agent large language model (LLM) systems, automated scientific discovery, dynamic tool integration via reinforcement learning, integrated search capabilities, and security vulnerabilities in agent protocols.
\end{itemize}

\begin{figure*}
    \centering
    \includegraphics[width=0.9\textwidth]{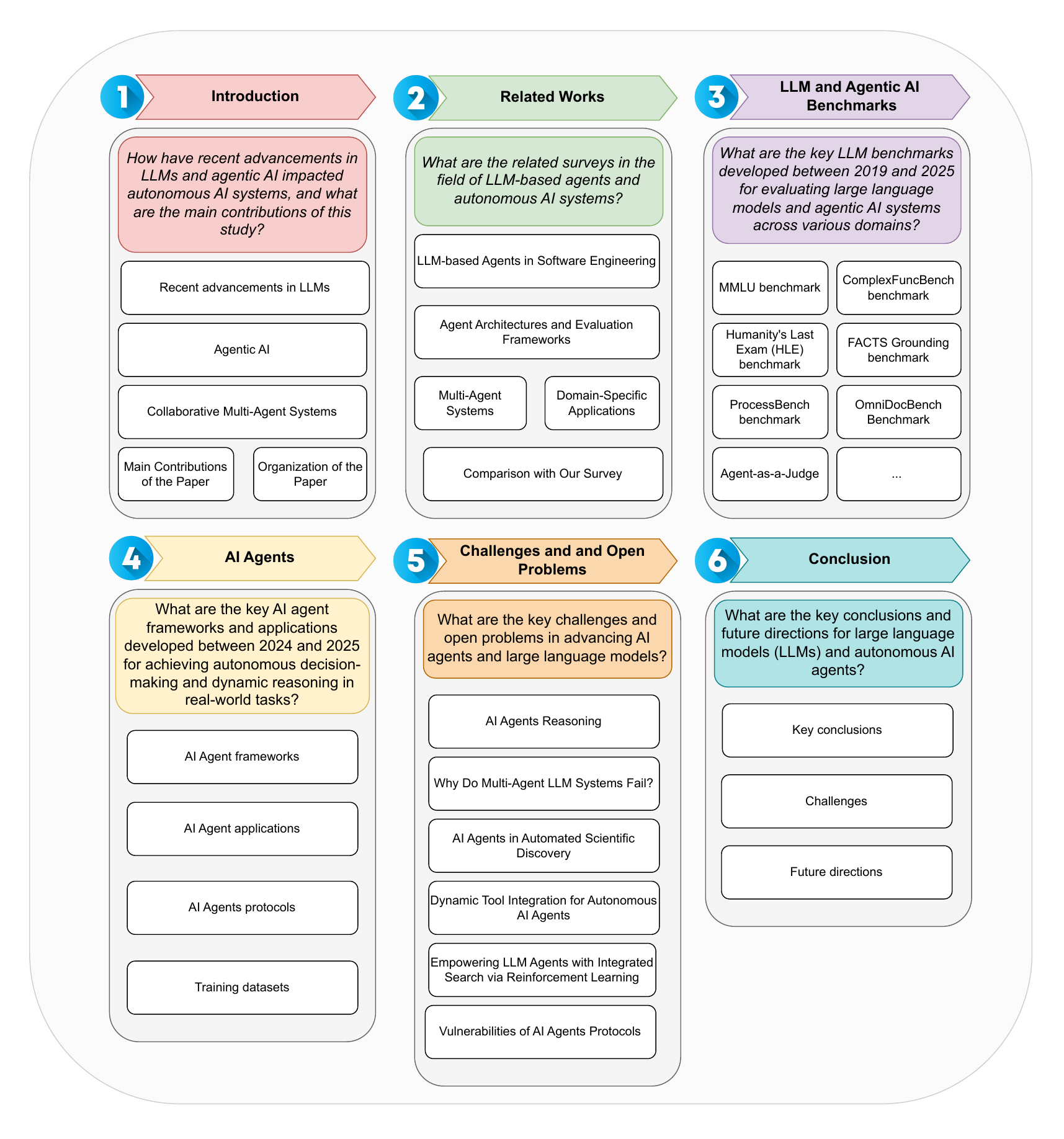}
    \caption{\textcolor{black}{Survey Structure.}}
    \label{fig:survey}
\end{figure*}

Fig. \ref{fig:survey} illustrates the structure of this survey. Section \ref{sec:2} presents the related works. Section \ref{sec:3} provides a side-by-side tabular comparison of state-of-the-art LLM and Agentic AI benchmarks. Section \ref{sec:4} reviews AI agent frameworks, AI agent applications, AI agent protocols, and training datasets across various domains. Section \ref{sec:5} highlights several critical research directions. Finally, Section \ref{sec:6} concludes the paper.

\begin{table*}[ht]
  \centering
  \scriptsize
  \caption{An overview of selected surveys on AI Agents.}
  \label{tab:feature-coverage}
  \begin{tabular}{p{2cm} p{1.3cm} c p{4.7cm} p{1cm}p{1cm}p{1cm}p{1cm}p{1cm}}
    \toprule
    \textbf{Theme} & \textbf{Reference} & \textbf{Year} & \textbf{Key Contribution} & \textbf{Benchmark} & \textbf{AI Agent Frameworks} & \textbf{AI Agent Applications} & \textbf{AI Agents Protocols} & \textbf{Challenges \& Open Problems} \\
    \midrule
    LLM‑based Agents in Software Engineering
      & Wang et al.~\cite{wang2024agents}
      & 2024
      & Survey of LLM‑based agent technologies in SE; proposes a perception–memory–action framework.
      & \emptycirc & \emptycirc & \fullcirc & \emptycirc & \halfcirc \\ \hline

    LLM‑based Agents in Software Engineering
      & Jin et al.~\cite{jin2024llms}
      & 2024
      & Reviews LLM vs. autonomous‑agent capabilities across six SE domains; highlights autonomy gaps.
      & \emptycirc & \emptycirc & \fullcirc & \emptycirc & \fullcirc \\ \hline

    Agent Architectures \& Evaluation
      & Singh et al.~\cite{singh2025agentic}
      & 2025
      & Introduces Agentic RAG: embedding autonomous agents in RAG with planning, reflection, tool use, and collaboration.
      & \emptycirc & \fullcirc & \fullcirc & \halfcirc & \halfcirc \\ \hline

    Agent Architectures \& Evaluation
      & Yehudai et al.~\cite{yehudai2025surveyevaluationllmbasedagents}
      & 2025
      & Surveys evaluation methodologies and benchmarks for LLM agents, covering cost, safety, and robustness.
      & \fullcirc & \emptycirc & \fullcirc & \halfcirc & \fullcirc \\ \hline

    Agent Architectures \& Evaluation
      & Chen et al.~\cite{chen2025towards}
      & 2025
      & Analyzes 1,676 RPAs, identifies core attributes, and proposes standardized evaluation guidelines.
      & \emptycirc & \emptycirc & \emptycirc & \emptycirc & \fullcirc \\ \hline

    Multi‑Agent Systems
      & Yan et al.~\cite{yan2025beyond}
      & 2025
      & Comprehensive survey of LLM‑powered MAS; focuses on communication, scalability, security, and multimodality.
      & \emptycirc & \emptycirc & \halfcirc & \fullcirc & \fullcirc \\ \hline

    Multi‑Agent Systems
      & Guo et al.~\cite{guo2024large}
      & 2024
      & Traces evolution from single‑agent LLM reasoning to collaborative MAS; examines profiling and communication.
      & \emptycirc & \emptycirc & \halfcirc & \emptycirc & \halfcirc \\ \hline

    Healthcare
      & Wang et al.~\cite{wang2025survey}
      & 2025
      & Reviews LLM‑agent architectures for clinical decision support, documentation, training; discusses ethics.
      & \emptycirc & \emptycirc & \fullcirc & \emptycirc & \fullcirc \\ \hline

    Social Agents in Game Theory
      & Feng et al.~\cite{feng2024survey}
      & 2024
      & Surveys LLM‑based social agents in game theory; categorizes frameworks, agent attributes, and evaluation protocols.
      & \emptycirc & \emptycirc & \fullcirc & \emptycirc & \fullcirc \\ \hline

    GUI Agents
      & Zhang et al.~\cite{zhang2024large}
      & 2024
      & Chronicles evolution of LLM‑driven GUI agents; covers multimodal understanding and large‑action models.
      & \emptycirc & \emptycirc & \fullcirc & \emptycirc & \fullcirc \\ \hline

    Personal LLM Agents
      & Li et al.~\cite{li2024personal}
      & 2024
      & Examines personal LLM agents integrating user data/devices; surveys architectures and security challenges.
      & \emptycirc & \emptycirc & \fullcirc & \emptycirc & \fullcirc \\ \hline

    Scientific Discovery
      & Gridach et al.~\cite{gridach2025agentic}
      & 2025
      & Explores Agentic AI in automating research workflows across domains; highlights reliability and ethics.
      & \emptycirc & \emptycirc & \fullcirc & \emptycirc & \fullcirc \\ \hline

    Chemistry
      & Ramos et al.~\cite{ramos2025review}
      & 2025
      & Reviews LLM roles in molecule design and synthesis planning; introduces agents for lab control.
      & \emptycirc & \emptycirc & \fullcirc & \emptycirc & \fullcirc \\ \hline

    Our Survey
      & Ferrag et al.
      & 2025
      & Unified end‑to‑end survey covering benchmarks, frameworks, applications, protocols, and challenges.
      & \fullcirc & \fullcirc & \fullcirc & \fullcirc & \fullcirc \\
    \bottomrule
  \end{tabular} \\
  Not Considered (\emptycirc); Partial discussion (\halfcirc); Considered (\fullcirc);
\end{table*}

\section{Related Works}\label{sec:2}

The growing field of autonomous AI agents powered by large language models has inspired a wide range of research efforts across multiple domains. In this section, we review the most relevant studies that investigate the integration of LLM-based agents into software engineering, propose agent architectures and evaluation frameworks, explore the development of multi-agent systems, and examine domain-specific applications, including healthcare, game-theoretic scenarios, GUI interactions, personal assistance, scientific discovery, and chemistry.

\subsection{LLM-based Agents in Software Engineering}
Wang et al. \cite{wang2024agents} present a survey that bridges Large Language Model (LLM)-based agent technologies with software engineering (SE). It highlights how LLMs have achieved significant success in various domains and have been integrated into SE tasks, often under the agent paradigm, whether explicitly or implicitly. The study presents a structured framework for LLM-based agents in SE, comprising three primary modules: perception, memory, and action. Jin et al. \cite{jin2024llms} investigate the use of large language models (LLMs) and LLM-based agents in software engineering, distinguishing between the traditional capabilities of LLMs and the enhanced functionalities offered by autonomous agents. It highlights the significant success of LLMs in tasks such as code generation and vulnerability detection, while also addressing their limitations, specifically the issues of autonomy and self-improvement that LLM-based agents aim to overcome. The paper provides an extensive review of current practices across six key domains: requirement engineering, code generation, autonomous decision-making, software design, test generation, and software maintenance. In a complementary study, Jin et al. \cite{jin2024llms} investigate the use of large language models (LLMs) and LLM-based agents in software engineering, distinguishing between the traditional capabilities of LLMs and the enhanced functionalities offered by autonomous agents. It highlights the significant success of LLMs in tasks such as code generation and vulnerability detection, while also addressing their limitations, specifically, issues of autonomy and self-improvement that LLM-based agents aim to overcome. The paper provides an extensive review of current practices across six key domains: requirement engineering, code generation, autonomous decision-making, software design, test generation, and software maintenance.

\subsection{Agent Architectures and Evaluation Frameworks}
Singh et al. \cite{singh2025agentic} delves into Agentic Retrieval-Augmented Generation (Agentic RAG), a sophisticated evolution of traditional Retrieval-Augmented Generation systems that enhances the capabilities of large language models (LLMs). While LLMs have transformed AI through human-like text generation and language understanding, their dependence on static training data often results in outdated or imprecise responses. The paper addresses these limitations by embedding autonomous agents within the RAG framework, enabling dynamic, real-time data retrieval and adaptive workflows. It details how agentic design patterns such as reflection, planning, tool utilization, and multi-agent collaboration equip these systems to manage complex tasks and support multi-step reasoning. The survey offers a comprehensive taxonomy of Agentic RAG architectures, highlights key applications across various sectors, including healthcare, finance, and education, and outlines practical implementation strategies.

Complementing this architectural perspective, Yehudai et al. \cite{yehudai2025surveyevaluationllmbasedagents} mark a significant milestone in artificial intelligence by surveying evaluation methodologies for agents powered by large language models (LLMs). It thoroughly reviews the capabilities of these agents, focusing on core functions such as planning, tool utilization, self-reflection, and memory, while assessing specialized applications ranging from web interactions to software engineering and conversational tasks. The authors uncover a clear trend toward developing more rigorous, dynamically updated evaluation frameworks by examining both targeted benchmarks for domain-specific applications and those designed for more generalist agents. Moreover, the paper critically highlights existing deficiencies in the field, notably the need for metrics that more effectively capture cost efficiency, safety, and robustness. In doing so, it maps the current landscape of agent evaluation and sets forth compelling directions for future inquiry, underscoring the importance of scalable and fine-grained evaluation techniques in the rapidly evolving AI domain. 

Similarly, Chen et al. \cite{chen2025towards} focus on Role-Playing Agents (RPAs), a growing class of LLM-based agents that mimic human behavior across various tasks. Recognizing the inherent challenges in evaluating such diverse systems, the authors systematically reviewed 1,676 papers published between January 2021 and December 2024. Their extensive analysis identifies six key agent attributes, seven task attributes, and seven evaluation metrics that are prevalent in the current literature. Based on these insights, the paper proposes an evidence-based, actionable, and generalizable evaluation guideline designed to standardize the assessment of RPAs.

\subsection{Multi-Agent Systems}
Yan et al. \cite{yan2025beyond} provides a comprehensive survey on integrating LLMs into multi-agent systems (MAS). Their work emphasizes the communication-centric aspects that enable agents to engage in both cooperative and competitive interactions, thereby tackling tasks that are unmanageable for individual agents. The paper examines system-level features, internal communication mechanisms, and challenges, including scalability, security, and multimodal integration. In a related study, Guo et al. \cite{guo2024large} offer an extensive overview of LLM-based multi-agent systems, charting the evolution from single-agent decision-making to collaborative frameworks that enhance collective problem-solving and world simulation. In a related study, Guo et al. \cite{guo2024large} provide an extensive overview of large language model (LLM)-based multi-agent systems, building on the success of LLMs in autonomous planning and reasoning. The authors detail how the evolution from single-agent decision-making to collaborative multi-agent frameworks has enabled significant advances in complex problem-solving and world simulation. Key aspects of these systems are examined, including the domains and environments they simulate, the profiling and communication strategies employed by individual agents, and the mechanisms that underpin the enhancement of their collective capacities.

\subsection{Domain-Specific Applications}

\subsubsection{Healthcare}
Wang et al.  \cite{wang2025survey} explores the transformative impact of LLM-based agents on healthcare, presenting a detailed review of their architectures, applications, and inherent challenges. It dissects the core components of medical agent systems, such as system profiles, clinical planning mechanisms, and medical reasoning frameworks, while also discussing methods to enhance external capacities. Major application areas include clinical decision support, medical documentation, training simulations, and overall healthcare service optimization. The survey further evaluates the performance of these agents using established frameworks and metrics, identifying persistent challenges such as hallucination management, multimodal integration, and ethical considerations.

\subsubsection{Social Agents in Game-Theoretic Scenarios}
Feng et al. \cite{feng2024survey} provide a review of research on LLM-based social agents in game-theoretic scenarios. This area has gained prominence for assessing social intelligence in AI systems. The authors categorize the literature into three main components. First, the game framework is examined, highlighting various choice- and communication-focused scenarios. Second, the paper explores the attributes of social agents, examining their preferences, beliefs, and reasoning capabilities. Third, it discusses evaluation protocols incorporating game-agnostic and game-specific metrics to assess performance. By synthesizing current studies and outlining future research directions, the survey offers valuable insights to further the development and systematic evaluation of social agents within game-theoretic contexts.

\subsubsection{GUI Agents}
Zhang et al. \cite{zhang2024large} review LLM-brained GUI agents, marking a paradigm shift in human-computer interaction through integrating multimodal LLMs. It traces the historical evolution of GUI automation, detailing how advancements in natural language understanding, code generation, and visual processing have enabled these agents to interpret complex graphical user interface (GUI) elements and execute multi-step tasks from conversational commands. The survey systematically examines the core components of these systems, including existing frameworks, data collection and utilization methods for training, and the development of specialized large-scale action models for GUI tasks.

\subsubsection{Personal LLM Agents}
Li et al. \cite{li2024personal} explore the evolution of intelligent personal assistants (IPAs) by focusing on Personal LLM Agents, LLM-based agents that deeply integrate personal data and devices to provide enhanced personal assistance. The authors outline the limitations of traditional IPAs, including insufficient understanding of user intent, task planning, and tool utilization, which have hindered their practicality and scalability. In contrast, the emergence of foundation models like LLMs offers new possibilities by leveraging advanced semantic understanding and reasoning for autonomous problem-solving. The survey systematically reviews the architecture and design choices underlying Personal LLM Agents, informed by expert opinions, and examines key challenges related to intelligence, efficiency, and security. Furthermore, it comprehensively analyzes representative solutions addressing these challenges, laying the groundwork for Personal LLM Agents to become a major paradigm in next-generation end-user software.

\subsubsection{Scientific Discovery}
Gridach et al. \cite{gridach2025agentic} explore the transformative role of Agentic AI in scientific discovery, underscoring its potential to automate and enhance research processes. It reviews how these systems, endowed with reasoning, planning, and autonomous decision-making capabilities, are revolutionizing traditional research activities, including literature reviews, hypothesis generation, experimental design, and data analysis. The paper highlights recent advancements across multiple scientific domains, such as chemistry, biology, and materials science, by categorizing existing Agentic AI systems and tools. It provides a detailed discussion on key evaluation metrics, implementation frameworks, and datasets used in the field, offering valuable insights into current practices. Moreover, the paper critically addresses significant challenges, including automating comprehensive literature reviews, ensuring system reliability, and addressing ethical concerns. It outlines future research directions, emphasizing the importance of human-AI collaboration and improved system calibration.

\subsubsection{Chemistry}
Ramos et al. \cite{ramos2025review} examine the transformative impact of large language models (LLMs) in chemistry, focusing on their roles in molecule design, property prediction, and synthesis optimization. It highlights how LLMs not only accelerate scientific discovery through automation but also discuss the advent of LLM-based autonomous agents. These agents extend the functionality of LLMs by interfacing with their environment and performing tasks such as literature scraping, automated laboratory control, and synthesis planning. Expanding the discussion beyond chemistry, the review also considers applications across other scientific domains.

\subsection{Comparison with Our Survey}
Table~\ref{tab:feature-coverage} presents a consolidated view of how existing works cover key themes, benchmarks, AI agent frameworks, AI agent applications, AI agents protocols, and challenges \& open problems against our survey. While prior studies typically focus on one or two aspects (e.g., Yehudai et al.~\cite{yehudai2025surveyevaluationllmbasedagents} on evaluation benchmarks, Singh et al.~\cite{singh2025agentic} on RAG architectures, Yan et al.~\cite{yan2025beyond} on multi-agent communication, or Wang et al.~\cite{wang2025survey} on domain-specific applications), none integrate the full spectrum of developments in a single, unified treatment. In contrast, our survey is the first to systematically combine state-of-the-art benchmarks, framework design, application domains, communication protocols, and a forward-looking discussion of challenges and open problems, thereby providing researchers with a comprehensive roadmap for advancing LLM-based autonomous AI agents.

\begin{table*}[htbp]
\centering
\scriptsize
\caption{Summary of LLM Benchmarks (Part 1)}
\label{tab:llm_benchmarks_group1}
\begin{adjustbox}{max width=\textwidth}
\begin{tabularx}{\textwidth}{@{}%
>{\raggedright\arraybackslash}p{1.5cm}
>{\centering\arraybackslash}p{1cm}
>{\raggedright\arraybackslash}p{2cm}
>{\raggedright\arraybackslash}X
>{\raggedright\arraybackslash}X
>{\raggedright\arraybackslash}X
@{}}
\toprule
\textbf{Benchmark / Dataset} & \textbf{Year} & \textbf{Evaluation Focus} & \textbf{Key Features / Metrics} & \textbf{Innovations/Techniques} & \textbf{Observations} \\
\midrule
ENIGMAEVAL \cite{wang2025enigmaeval} & 2025 & Multimodal Reasoning & Contains 1,184 puzzles combining text and images; state-of-the-art systems score only $\sim$7\% on standard puzzles and fail on the hardest ones. & Evaluates multimodal and long-context reasoning using challenging puzzles from global competitions. & Pushes models into unstructured, creative problem-solving scenarios requiring integration of visual and semantic clues. \\
\midrule
MMLU Benchmark \cite{hendrycks2020measuring} & 2021 & Multitask Knowledge & Comprises 57 diverse tasks (from elementary math to professional law) testing zero-shot and few-shot performance. & Assesses broad world knowledge and problem-solving skills; uncovers calibration challenges and imbalances between procedural and declarative knowledge. & Designed for general multitask language understanding without task-specific fine-tuning. \\
\midrule
ComplexFuncBench \cite{zhong2025complexfuncbench} & 2025 & Function Calling & Evaluates complex function calling tasks with multi-step operations and input lengths up to 128k tokens over more than 1,000 scenarios. & Introduces an automatic evaluation framework (ComplexEval) for function calling, testing reasoning over implicit parameters and constraints. & Highlights performance differences between closed models (e.g., Claude 3.5, GPT-4) and open models (e.g., Qwen 2.5, Llama 3.1). \\
\midrule
Humanity's Last Exam (HLE) \cite{phan2025humanity} & 2025 & Academic Reasoning & Features 3,000 questions spanning over 100 subjects, including multi-modal challenges. & Developed through a global collaborative effort with nearly 1,000 experts; includes both multiple-choice and short-answer formats with verifiable answers. & Exposes significant performance gaps as state-of-the-art LLMs score below 10\%, serving as a critical tool for assessing academic reasoning. \\
\midrule
FACTS Grounding \cite{deepmind2023facts} & 2023 & Factual Grounding & Contains 1,719 examples requiring detailed responses grounded in source documents, with inputs reaching up to 32,000 tokens. & Uses a two-phase evaluation (eligibility and factual grounding) with assessments from frontier LLM judges. & Focuses on factual accuracy and information synthesis while excluding creative or complex reasoning tasks. \\
\midrule
ProcessBench \cite{zheng2024processbench} & 2024 & Error Detection & Comprises 3,400 math problem cases with step-by-step solutions and human-annotated error locations. & Evaluates models’ ability to detect the earliest error in reasoning; compares process reward models with LLM-based critics. & Targets granular error detection in mathematical problem solving. \\
\midrule
OmniDocBench \cite{ouyang2024omnidocbench} & 2024 & Document Understanding & A multi-source dataset spanning nine document types with 19 layout categories and 14 attribute labels. & Provides a detailed, multi-level evaluation framework for document content extraction, contrasting modular pipelines with end-to-end methods. & Addresses challenges such as fuzzy scans, watermarks, and complex layouts in document processing. \\
\midrule
Agent-as-a-Judge \cite{zhuge2024agent} & 2024 & Evaluation Methodology & Evaluated on 55 code generation tasks with 365 hierarchical user requirements. & Leverages agentic systems to provide granular, intermediate feedback; achieves up to 90\% alignment with human judgments. & Reduces evaluation cost and time for agentic systems, particularly in code generation tasks. \\
\midrule
JudgeBench \cite{tan2024judgebench} & 2024 & Judgment Evaluation & Consists of 350 challenging response pairs across knowledge, reasoning, math, and coding domains. & Transforms existing datasets into paired comparisons with objective correctness, mitigating positional bias through double evaluation. & Aims to objectively assess LLM-based judges; fine-tuning can boost judge accuracy significantly. \\
\midrule
SimpleQA \cite{openai2023simpleqa} & 2023 & Factual QA & Contains 4,326 fact-seeking questions across domains; uses a strict three-tier grading system. & Focuses on evaluating factual accuracy and reveals models’ overconfidence in incorrect responses through repeated testing. & Highlights current limitations in handling straightforward, factual queries. \\
\midrule
FineTasks \cite{huggingfacefw2023finetasks} & 2023 & Multilingual Task Selection & Evaluates 185 candidate tasks across nine languages, ultimately selecting 96 reliable tasks; supports over 550 tasks overall. & Employs metrics such as monotonicity, low noise, non-random performance, and model ordering consistency to assess task quality. & Provides a scalable, multilingual evaluation platform that highlights the impact of task formulation. \\
\midrule
FRAMES \cite{krishna2024fact} & 2024 & Retrieval \& Reasoning & Consists of 824 multi-hop questions requiring integration of 2–15 Wikipedia articles. & Unifies evaluations of factual accuracy, retrieval, and reasoning; labels questions with specific reasoning types (e.g., numerical, tabular). & Baseline experiments show improvements from 40\% (without retrieval) to 66\% (with multi-step retrieval). \\
\midrule
DABStep \cite{huggingface2025dabstep} & 2025 & Step-Based Reasoning & A step-based approach for multi-step reasoning tasks; the best model achieves only a 16\% success rate. & Decomposes complex problem solving into discrete steps with iterative refinement and self-correction. & Highlights the significant challenges in training models for complex, iterative reasoning. \\
\bottomrule
\end{tabularx}
\end{adjustbox}
\end{table*}

\begin{table*}[htbp]
\centering
\scriptsize
\caption{Summary of LLM Benchmarks (Part 2)}
\label{tab:llm_benchmarks_group2}
\begin{adjustbox}{max width=\textwidth}
\begin{tabularx}{\textwidth}{@{}%
>{\raggedright\arraybackslash}p{1.5cm}
>{\centering\arraybackslash}p{1cm}
>{\raggedright\arraybackslash}p{2cm}
>{\raggedright\arraybackslash}X
>{\raggedright\arraybackslash}X
>{\raggedright\arraybackslash}X
@{}}
\toprule
\textbf{Benchmark / Dataset} & \textbf{Year} & \textbf{Evaluation Focus} & \textbf{Key Features / Metrics} & \textbf{Innovations/Techniques} & \textbf{Observations} \\
\midrule
BFCL v2 \cite{mao2025bfcl} & 2025 & Function Calling & Contains 2,251 question-function-answer pairs covering simple to parallel function calls. & Leverages real-world, user-contributed data to address issues like data contamination and bias in function calling evaluation. & Demonstrates that models such as Claude 3.5 and GPT-4 outperform others, while some open models struggle. \\
\midrule
SWE-Lancer \cite{miserendino2025swelancerfrontierllmsearn} & 2025 & Software Engineering & Consists of over 1,400 freelance software engineering tasks, including independent and managerial tasks with real-world payout data. & Uses triple-verified tests for independent tasks and benchmarks managerial decisions against hiring manager selections. & Indicates that even advanced models (e.g., Claude 3.5 Sonnet) have low pass rates (26.2\%) on implementation tasks. \\
\midrule
CRAG Benchmark \cite{yang2024crag} & 2024 & Retrieval-Augmented Generation & Comprises 4,409 question-answer pairs across 5 domains; simulates retrieval with mock APIs. & Evaluates the generative component of RAG pipelines; shows improvement from 34\% to 63\% accuracy with advanced RAG methods. & Highlights performance drops for questions involving highly dynamic or less popular facts. \\
\midrule
OCCULT Benchmark \cite{kouremetis2025occultevaluatinglargelanguage} & 2025 & Cybersecurity & A lightweight framework for operational evaluation of cybersecurity risks; includes three distinct OCO benchmarks. & Simulates real-world threat scenarios to assess LLM capabilities in offensive cyber operations. & Preliminary results indicate models like DeepSeek-R1 achieve over 90\% in Threat Actor Competency Tests. \\
\midrule
DIA Benchmark \cite{tihanyi2024dynamic} & 2024 & Dynamic Problem Solving & Uses dynamic question templates with mutable parameters across domains (math, cryptography, cybersecurity, computer science). & Introduces innovative metrics for reliability and confidence over multiple attempts; emphasizes adaptive intelligence. & Reveals gaps in handling complex tasks and compares models’ self-assessment abilities. \\
\midrule
CyberMetric Benchmark \cite{tihanyi2024cybermetric} & 2024 & Cybersecurity Knowledge & A suite of multiple-choice Q\&A datasets (CyberMetric-80, -500, -2000, -10000) validated over 200 human expert hours. & Generated using GPT-3.5 and RAG, it benchmarks cybersecurity knowledge against human performance. & Demonstrates that larger, domain-specific models outperform smaller ones in cybersecurity understanding. \\
\midrule
BIG-Bench Extra Hard \cite{kazemi2025big} & 2025 & Challenging Reasoning & An elevated-difficulty variant of BIG-Bench Hard; average accuracy is 9.8\% for general models and 44.8\% for reasoning-specialized models. & Replaces each BBH task with a more challenging variant to probe reasoning capabilities robustly. & Emphasizes substantial room for improvement in general-purpose reasoning skills. \\
\midrule
MultiAgentBench \cite{zhu2025multiagentbench}
& 2025
& Multi-Agent
& Encompasses six domains: research proposal writing, Minecraft structure building, database error analysis, collaborative coding, competitive Werewolf gameplay, and resource bargaining. 
& Investigates various coordination protocols (star, chain, tree, graph); peer-to-peer communication plus cognitive planning yields a 3\% improvement in milestone achievement. Graph-based protocols outperform others in research tasks.
& GPT-4o-mini achieves the highest average task score; highlights synergy vs. complexity trade-offs in multi-agent LLM settings. \\
\midrule
GAIA \cite{mialon2023gaia}
& 2024
& General AI Assistants
& 466 curated questions with reference answers; humans achieve 92\% accuracy while GPT-4 with plugins only reaches 15\%.
& Emphasizes everyday reasoning tasks involving multi-modality, web browsing, and tool use. Targets AI robustness over specialized skills.
& Highlights the large performance gap between humans and SOTA models; aims to measure truly general-purpose AI capabilities. \\ \hline
CASTLE \cite{dubniczky2025castle} & 2025 & Vulnerability detection in source code & 250 hand-crafted micro-benchmark programs covering 25 common CWEs; introduces the novel CASTLE Score metric & Integrates evaluations across 13 static analysis tools, 10 LLMs, and two formal verification tools; provides a unified framework for comparing diverse methods & Formal verification tools (e.g., ESBMC) minimize false positives but miss vulnerabilities beyond model checking; static analyzers generate excessive false positives; LLMs perform well on small code snippets, but accuracy declines and hallucinations increase as code size grows \\ \hline
SPIN-Bench \cite{yao2025spin} & 2025 & Strategic Planning, Interaction, and Negotiation & Evaluates reasoning and strategic behavior in diverse social settings by combining classical PDDL tasks, competitive board games, cooperative card games, and multi-agent negotiation scenarios. & Systematically varies action spaces, state complexity, and the number of interacting agents to simulate realistic social interactions, providing both a benchmark and an arena for multi-agent evaluation. & Reveals that while LLMs perform basic fact retrieval and short-range planning reasonably well, they struggle with deep multi-hop reasoning and socially adept coordination, highlighting a significant gap in robust multi-agent planning and human–AI teaming. \\
\midrule
\(\tau\)-bench \cite{yao2024tau}& 2024 & Conversational Agent Evaluation & Evaluates dynamic, multi-turn conversations by comparing the final database state with an annotated goal state using a novel \(\text{pass}^k\) metric. & Integrates domain-specific API tool usage and strict policy adherence within simulated user interactions to assess agent reliability over multiple trials. & Reveals that even state-of-the-art agents (e.g., GPT-4o) succeed on less than 50\% of tasks, with marked inconsistency (e.g., \(\text{pass}^8 < 25\%\) in retail), highlighting the need for improved consistency and rule-following. \\
\bottomrule
\end{tabularx}
\end{adjustbox}
\end{table*}

\section{LLM and Agentic AI Benchmarks}\label{sec:3}

This section provides a comprehensive overview of benchmarks developed between 2019 and 2025 that rigorously evaluate large language models (LLMs) across diverse and challenging domains. For instance, ENIGMAEVAL \cite{wang2025enigmaeval} assesses complex multimodal puzzle-solving by requiring the synthesis of textual and visual clues, while ComplexFuncBench \cite{zhong2025complexfuncbench} challenges models with multi-step function-calling tasks that mirror real-world scenarios. Humanity’s Last Exam (HLE) \cite{phan2025humanity} further raises the bar by presenting expert-level academic questions across a broad spectrum of subjects, thereby reflecting the growing demand for deeper reasoning and domain-specific proficiency. Additional frameworks such as FACTS Grounding \cite{deepmind2023facts} and ProcessBench \cite{zheng2024processbench} scrutinize the models’ capacities for generating factually accurate long-form responses and detecting errors in multi-step reasoning. Meanwhile, innovative evaluation paradigms like Agent-as-a-Judge \cite{zhuge2024agent}, JudgeBench \cite{tan2024judgebench}, and CyberMetric \cite{tihanyi2024cybermetric} provide granular insights into cybersecurity competencies and error-detection capabilities. Tables \ref{tab:llm_benchmarks_group2}, \ref{tab:llm_benchmarks_group1} present a comprehensive overview of benchmarks developed between 2024 and 2025.

\subsection{ENIGMAEVAL benchmark}

ENIGMAEVAL \cite{wang2025enigmaeval} is a benchmark designed to rigorously evaluate advanced language models' multimodal and long-context reasoning capabilities using challenging puzzles derived from global competitions. The dataset comprises 1,184 complex puzzles that combine text and images, requiring models to synthesize disparate clues, perform multi-step deductive reasoning, and integrate visual and semantic information to arrive at unambiguous, verifiable solutions. Unlike conventional benchmarks focusing on well-structured academic tasks, ENIGMAEVAL pushes models into unstructured, creative problem-solving scenarios where even state-of-the-art systems achieve only about 7\% accuracy on standard puzzles and fail on the hardest ones. 

\subsection{MMLU Benchmark}

Measuring Massive Multitask Language Understanding (MMLU) \cite{hendrycks2020measuring} is a comprehensive benchmark designed by Hendrycks et al. (2021) to evaluate large language models across a diverse range of subjects, from elementary mathematics to professional law. The benchmark comprises 57 tasks that test models' ability to apply broad world knowledge and problem-solving skills in zero-shot and few-shot settings, emphasizing generalization without task-specific fine-tuning. The study also uncovers challenges related to model calibration and the imbalance between procedural and declarative knowledge, highlighting critical areas where current models fall short of expert-level proficiency.

\subsection{ComplexFuncBench Benchmark} 

Zhong et al. \cite{zhong2025complexfuncbench} introduced ComplexFuncBench, a novel benchmark designed to evaluate large language models (LLMs) on complex function calling tasks in real-world settings. Unlike previous benchmarks, ComplexFuncBench challenges models with multi-step operations within a single turn, adherence to user-imposed constraints, reasoning over implicit parameter values, and managing extensive input lengths that can exceed 500 tokens, including scenarios with a context window of up to 128k tokens. Complementing the benchmark, the authors present an automatic evaluation framework, ComplexEval, which quantitatively assesses performance across over 1,000 scenarios derived from five distinct aspects of function calling. Experimental results reveal significant limitations in current state-of-the-art LLMs, with closed models like Claude 3.5 and OpenAI's GPT-4 outperforming open models such as Qwen 2.5 and Llama 3.1. Notably, the study identifies prevalent issues, including value errors and premature termination in multi-step function calls, underscoring the need for further research to enhance the function-calling capabilities of LLMs in practical applications.

\subsection{Humanity's Last Exam (HLE) Benchmark}
Phan et al. \cite{phan2025humanity} introduced Humanity's Last Exam (HLE), a benchmark designed to push the limits of LLMs by challenging them with expert-level academic tasks. Unlike traditional benchmarks such as MMLU, where LLMs have achieved over 90\% accuracy, HLE presents a significantly more demanding test, featuring 3,000 questions spanning over 100 subjects including mathematics, humanities, and the natural sciences. This benchmark is the product of a global collaborative effort, with nearly 1,000 subject matter experts from over 500 institutions contributing questions that are both multi-modal and resistant to quick internet retrieval, ensuring that only genuine deep academic understanding can lead to success. The tasks, which include both multiple-choice and short-answer formats with clearly defined, verifiable answers, expose a substantial performance gap: current state-of-the-art LLMs, such as DeepSeek R1, OpenAI's models, Google DeepMind Gemini Thinking, and Anthropic Sonnet 3.5, perform at less than 10\% accuracy and suffer from high calibration errors, indicating overconfidence in incorrect responses. The results underscore that while existing benchmarks may no longer provide a meaningful measure of progress, HLE serves as a critical tool for assessing the true academic reasoning capabilities of LLMs, potentially heralding a new era in benchmark design as the field moves toward more challenging and nuanced evaluations in the pursuit of artificial general intelligence.

\subsection{FACTS Grounding benchmark}

Google DeepMind introduced FACTS Grounding \cite{deepmind2023facts}, a comprehensive benchmark designed to evaluate how accurately LLMs ground their long-form responses in provided source documents while avoiding hallucinations. The benchmark comprises 1,719 meticulously crafted examples split into 860 public and 859 private cases that require models to generate detailed answers strictly based on a corresponding context document, with inputs reaching up to 32,000 tokens. Covering diverse domains such as medicine, law, technology, finance, and retail, FACTS Grounding excludes tasks that require creativity, mathematics, or complex reasoning, focusing squarely on factual accuracy and information synthesis. To ensure robust and unbiased evaluation, responses are assessed in two phases: eligibility and factual grounding using a panel of three frontier LLM judges (Gemini 1.5 Pro, GPT-4o, and Claude 3.5 Sonnet), with final scores derived from the aggregation of these assessments. With an online leaderboard hosted on Kaggle already populated with initial results where, for instance, Gemini 2.0 Flash leads with 83.6\% accuracy FACTS Grounding aims to drive industry-wide advancements in grounding and factuality, ultimately fostering greater trust and reliability in LLM applications.

\subsection{ProcessBench benchmark}

Qwen team \cite{zheng2024processbench} introduced ProcessBench, a novel benchmark specifically designed to evaluate the ability of language models to detect errors within the reasoning process for mathematical problem solving. ProcessBench comprises 3,400 test cases, primarily drawn from competition- and Olympiad-level math problems, where each case includes a detailed, step-by-step solution with human-annotated error locations. Models are tasked with identifying the earliest erroneous step or confirming that all steps are correct, thereby providing a granular assessment of their reasoning accuracy. The benchmark is employed to evaluate two classes of models: process reward models (PRMs) and critic models, the latter involving general large language models (LLMs) that are prompted to critique each solution step. Experimental results reveal two key findings. First, existing PRMs generally fail to generalize to more challenging math problems beyond standard datasets like GSM8K and MATH, often underperforming relative to both prompted LLM-based critics and a PRM fine-tuned on a larger, more complex PRM800K dataset. Second, the best open-source model tested, QwQ-32B-Preview, demonstrates error detection capabilities that rival those of the proprietary GPT-4o, although it still falls short compared to reasoning-specialized models like o1-mini.

\subsection{OmniDocBench Benchmark}

Ouyang et al. \cite{ouyang2024omnidocbench} introduced OmniDocBench, a comprehensive multi-source benchmark designed to advance automated document content extraction a critical component for high-quality data needs in LLMs and RAG systems. OmniDocBench features a meticulously curated and annotated dataset spanning nine diverse document types including academic papers, textbooks, slides, notes, and financial documents and utilizes a detailed evaluation framework with 19 layout categories and 14 attribute labels to facilitate multi-level assessments. Through extensive comparative analysis of existing modular pipelines and multimodal end-to-end methods, the benchmark reveals that while specialized models (e.g., Nougat) outperform general vision-language models (VLMs) on standard documents, general VLMs exhibit superior resilience and adaptability in challenging scenarios, such as those involving fuzzy scans, watermarks, or colorful backgrounds. Moreover, fine-tuning general VLMs with domain-specific data leads to enhanced performance, as evidenced by high accuracy scores in tasks like formula recognition (with models such as GPT-4o, Mathpix, and UniMERNet achieving around 85–86.8\% accuracy) and table recognition (RapidTable at 82.5\%). Nonetheless, the findings also highlight persistent challenges, notably that complex column layouts continue to degrade reading order accuracy across all evaluated models.

\subsection{Agent-as-a-Judge}

Meta team proposed the Agent-as-a-Judge framework \cite{zhuge2024agent}, an innovative evaluation approach explicitly designed for agentic systems that overcome the limitations of traditional methods, which either focus solely on outcomes or require extensive manual labor. This framework provides granular, intermediate feedback throughout the task-solving process by leveraging agentic systems to evaluate other agentic systems. The authors demonstrate its effectiveness on code generation tasks using DevAI, a new benchmark comprising 55 realistic automated AI development tasks annotated with 365 hierarchical user requirements. Their evaluation shows that Agent-as-a-Judge not only dramatically outperforms the conventional LLM-as-a-Judge approach (which typically achieves a 60–70\% alignment rate with human assessment) but also reaches an impressive 90\% alignment with human judgments. Additionally, this method offers substantial cost and time savings, reducing evaluation costs to approximately 2.29\% (\$30.58 vs. \$1,297.50) and cutting evaluation time down to 118.43 minutes compared to 86.5 hours for human assessments.

\begin{table*}[htbp]
  \centering
  \caption{LLM Benchmark Comparison: Multimodal, Task Diversity, Reasoning \& Agentic AI Evaluation}
  \label{tab:llm_benchmark}
  \footnotesize
  \begin{tabular}{l c c c c c c}
    \toprule
    Benchmark & Year & Multimodal & Task & Diversity & Reasoning & Agentic AI \\
    \midrule
    DROP \cite{dua2019drop} & 2019 & No & English discrete reasoning comprehension & High & High & No \\
    MMLU \cite{hendrycks2020measuring} & 2020 & No  & Academic/general knowledge & High & Moderate & No \\
    MATH \cite{hendrycks2021measuring} & 2021 & No & Evaluating mathematical reasoning & High & High & No \\
    Codex \cite{chen2021evaluating} & 2021 & No & Evaluating LLMs trained on code & Medium & Medium & No \\
    MGSM \cite{shi2022language} & 2022 & No & Multilingual grade-school math problems & High & High & No \\
    FACTS Grounding \cite{deepmind2023facts} & 2023 & No  & Factual grounding in long responses & High & Low & No \\
    SimpleQA \cite{openai2023simpleqa} & 2023 & No  & Factual Q\&A & High & Low & No \\
    PersonaGym \cite{samuel2024personagym} & 2024 & No & Dynamic evaluation framework for persona agents & High & High & Yes \\
    FineTasks \cite{huggingfacefw2023finetasks} & 2023 & No  & Multilingual task selection & High & Medium & No \\
    GAIA \cite{mialon2023gaia} & 2023 & Yes & General AI assistant tasks & High & High & No \\
    \midrule
    OmniDocBench \cite{ouyang2024omnidocbench} & 2024 & Yes & Document content extraction & High & Medium & No \\
    ProcessBench \cite{zheng2024processbench} & 2024 & No  & Error detection in math solutions & Low  & High & No \\
    MIRAI \cite{ye2024mirai} & 2024 & No & Evaluating llm agents for event forecasting & High & High & Yes \\
    AppWorld \cite{trivedi2024appworld} & 2025 & No & Benchmarking Interactive Coding Agents & High & High & Yes \\
    VisualAgentBench\cite{liu2024visualagentbench} & 2024 & Yes & Benchmark for evaluating Large Multimodal Models & High & High & Yes \\
    ScienceAgentBench \cite{chen2024scienceagentbench} & 2024 & No & Evaluation of language agents for Scientific Discovery & High & High & Yes \\
    Agent-SafetyBench \cite{zhang2024agent} & 2024 & No & Safety evaluation of LLM agents & High & High & Yes \\
    DiscoveryBench \cite{majumder2024discoverybench} & 2024 & No & Data-Driven Discovery & High & High & Yes \\
    BLADE \cite{gu2024blade} & 2024 & No & Benchmark for data-driven scientific discovery & High & High & Yes \\
    Dyn-VQA \cite{li2024benchmarking} & 2024 & Yes & Adaptive VQA multimodal benchmark  & High & High & Yes \\
    Agent-as-a-Judge \cite{zhuge2024agent} & 2024 & No  & Code generation evaluation & Low  & Low  & Yes \\
    JudgeBench \cite{tan2024judgebench} & 2024 & No  & Evaluation of LLM-based judges & High & High & No \\
    FRAMES \cite{krishna2024fact} & 2024 & No  & Factuality \& retrieval for RAG & High & High & No \\
    MedChain \cite{liu2024medchain} & 2024 & No & Interactive clinical decision adaptation & High & High & Yes \\
    CRAG \cite{yang2024crag} & 2024 & No  & Factual Q\&A for RAG systems & High & High & No \\
    DIA \cite{tihanyi2024dynamic} & 2024 & Yes & Dynamic problem solving & High & High & No \\
    CyberMetric \cite{tihanyi2024cybermetric} & 2024 & No  & Cybersecurity Q\&A & Low  & Low & No \\
    TeamCraft \cite{long2024teamcraft} & 2024 & Yes & Collaborative Minecraft multimodal evaluation & High & High & Yes \\
    AgentHarm \cite{andriushchenko2024agentharm} & 2024 & No & LLM jailbreak robustness evaluation  & High & High & Yes \\
    \(\tau\)-bench \cite{yao2024tau}& 2024 & No & Conversational Agent Evaluation & High & High & Yes \\
    LegalAgentBench \cite{li2024legalagentbench} & 2024 & No & Evaluating LLM Agents in Legal Domain & High & High & Yes \\
    GPQA \cite{rein2024gpqa} & 2024 & No & Biology, physics, and chemistry & High & High & No \\
    \midrule
    ENIGMAEVAL \cite{wang2025enigmaeval} & 2025 & Yes & Complex multimodal puzzles & Low  & High & No \\
    ComplexFuncBench \cite{zhong2025complexfuncbench} & 2025 & No  & Function calling tasks & Medium & High & No \\
    MedAgentsBench \cite{tang2025medagentsbench} & 2025 & No  & Complex medical reasoning \& treatment planning & High & High & Yes \\
    Humanity's Last Exam \cite{phan2025humanity} & 2025 & Yes & Expert-level academic tasks & High & High & No \\
    DABStep \cite{huggingface2025dabstep} & 2025 & No  & Step-based multi-step reasoning & Low  & High & No \\
    BFCL v2 \cite{mao2025bfcl} & 2025 & No  & Function calling evaluation & High & High & No \\
    SWE-Lancer \cite{miserendino2025swelancerfrontierllmsearn} & 2025 & No  & Freelance software engineering tasks & High & Moderate & No \\
    OCCULT \cite{kouremetis2025occultevaluatinglargelanguage} & 2025 & No  & Cyber security operational tasks & Medium & High & No \\
    BIG-Bench Extra Hard \cite{kazemi2025big} & 2025 & No  & Challenging reasoning tasks & High & High & No \\
    MultiAgentBench \cite{zhu2025multiagentbench} & 2025 & Yes & Multi-agent coordination tasks & High & High & Yes \\
    CASTLE \cite{dubniczky2025castle} & 2025 & No  & Software vulnerability detection & Low  & Medium & No \\
    EmbodiedEval \cite{cheng2025embodiedeval} & 2025 & Yes & 3D embodied tasks benchmark & Medium & High & Yes \\
    SPIN-Bench \cite{yao2025spin} & 2025 & Yes & Strategic planning \& social reasoning & High & High & Yes \\
    OlympicArena \cite{huang2024olympicarena} & 2025 & Yes  & Olympic competition problems & Medium & High & No \\
    SciReplicate-Bench \cite{xiang2025scireplicate} & 2025 & No & Algorithm-driven code generation & High & High & Yes \\
    EconAgentBench \cite{fish2025econevals} & 2025 & No & Decision-making tasks in economic environments & High & High & Yes \\
    VeriLA \cite{sung2025verila} & 2025 & No & Human-centered LLM failure verification  & High & High & Yes \\
    CapaBench \cite{yang2025s} & 2025 & No & Evaluation of modular contributions in LLM agents & High & High & Yes \\
    AgentOrca \cite{li2025agentorca} & 2025 & No & Dual-system agent compliance evaluation & High & High & Yes \\
    ProjectEval \cite{liu2025projecteval} & 2025 & No & Project-level code generation evaluation & Medium & High & Yes \\
    RefactorBench \cite{gautam2025refactorbench} & 2025 & No & Autonomous multi-file refactoring evaluation & High & High & Yes \\
    BEARCUBS \cite{song2025bearcubs} & 2025 & Yes & Multimodal web agents evaluation & High & Medium & Yes \\
    Robotouille \cite{gonzalez2025robotouille} & 2025 & No & Asynchronous Planning Benchmark & High & High & Yes \\
    DSGBench \cite{tang2025dsgbench} & 2025 & No & Strategic games decision evaluation & Medium & High & Yes \\
    TheoremExplainBench \cite{ku2025theoremexplainagent} & 2025 & Yes & STEM theorem animation videos & Medium & High & Yes \\
    RefuteBench 2.0 \cite{yan2025refutebench} & 2025 & No & Multi-turn LLM feedback evaluation & High & High & Yes \\
    MLGym \cite{nathani2025mlgym} & 2025 & Yes & ML agents automate research & High & High & Yes  \\
    DataSciBench \cite{zhang2025datascibench} & 2025 & No & LLM Data Science Benchmark & High & High & Yes \\
    EmbodiedBench \cite{yang2025embodiedbench} & 2025 & Yes & Vision-driven embodied agent evaluation & High & High & Yes \\
    BrowseComp \cite{wei2025browsecomp} & 2025 & No & Benchmark for Browsing Agents & High & High & Yes \\
    Vending-Bench \cite{backlund2025vending} & 2025 & No   & Long-horizon business simulation          & Medium & High     & Yes \\
    MLE-bench \cite{chan2025mle} & 2025 & No & ML engineering-related competitions from Kaggle & Medium & High & Yes \\
    SWE-PolyBench \cite{rashid2025swepolybenchmultilanguagebenchmarkrepository} & 2025 & No & Evaluation of coding agents & High & High & Yes \\
    Multi-SWE-bench \cite{zan2025multi} & 2025 & No & Multilingual Benchmark for Issue Resolving & High & High & No \\
    \bottomrule
  \end{tabular}
\end{table*}

\subsection{JudgeBench Benchmark}

Tan et al. \cite{tan2024judgebench} proposed JudgeBench, a novel benchmark designed to objectively evaluate LLM-based judges models that are increasingly employed to assess and improve the outputs of large language models by focusing on their ability to accurately discern factual and logical correctness rather than merely aligning with human stylistic preferences. Unlike prior benchmarks that rely primarily on crowdsourced human evaluations, JudgeBench leverages a carefully constructed set of 350 challenging response pairs spanning knowledge, reasoning, math, and coding domains. The benchmark employs a novel pipeline to transform challenging existing datasets into paired comparisons with preference labels based on objective correctness while mitigating positional bias through double evaluation with swapped order. Comprehensive testing across various judge architectures, including prompted, fine-tuned, multi-agent judges, and reward models, reveals that even strong models, such as GPT-4o, often perform only marginally better than random guessing, particularly on tasks requiring rigorous error detection in intermediate reasoning steps. Moreover, fine-tuning can significantly boost performance, as evidenced by a 14\% improvement observed in Llama 3.1 8B, and reward models achieve accuracies in the 59–64\% range.

\subsection{SimpleQA Benchmark}

SimpleQA \cite{openai2023simpleqa} is a benchmark introduced by OpenAI to assess and improve the factual accuracy of large language models on short, fact-seeking questions. Comprising 4,326 questions spanning domains such as science/tech, politics, art, and geography, SimpleQA challenges models to deliver a single correct answer under a strict three-tier grading system ("correct," "incorrect," or "not attempted"). While built on foundational datasets such as TriviaQA and Natural Questions, SimpleQA presents a more challenging task for LLMs. Early results indicate that even advanced models, such as OpenAI o1-preview, achieve only 42.7\% accuracy (with Claude 3.5 Sonnet trailing at 28.9\%), and models tend to exhibit overconfidence in their incorrect responses. Moreover, experiments that repeated the same question 100 times revealed a strong correlation between higher answer frequency and overall accuracy. This benchmark thus provides critical insights into the current limitations of LLMs in handling straightforward, factual queries. It underscores the need for further improvements in grounding model outputs in reliable, factual data.

\subsection{FineTasks}

FineTasks \cite{huggingfacefw2023finetasks} is a data-driven evaluation framework designed to systematically select reliable tasks for assessing LLMs across diverse languages. Developed as the first step toward the broader FineWeb Multilingual initiative, FineTasks evaluates candidate tasks based on four critical metrics: monotonicity, low noise, non-random performance, and model ordering consistency to ensure robustness and reliability. In an extensive study, the Hugging Face team tested 185 candidate tasks across nine languages (including Chinese, French, Arabic, Russian, Thai, Hindi, Turkish, Swahili, and Telugu), ultimately selecting 96 final tasks that cover domains such as reading comprehension, general knowledge, language understanding, and reasoning. The work further reveals that the formulation of tasks has a significant impact on performance; for instance, Cloze format tasks are more effective during early training phases, while multiple-choice formats yield better evaluation results. Recommended evaluation metrics include length normalization for most tasks and pointwise mutual information (PMI) for complex reasoning challenges. Benchmarking 35 open and closed-source LLMs demonstrated that open models are narrowing the gap with their proprietary counterparts, with Qwen 2 models excelling in high- and mid-resource languages and Gemma-2 particularly strong in low-resource settings. Moreover, the FineTasks framework supports over 550 tasks across various languages, providing a scalable and comprehensive platform for advancing multilingual large language model (LLM) evaluation.

\subsection{FRAMES benchmark}

Google team \cite{krishna2024fact} propose FRAMES (Factuality, Retrieval, and Reasoning MEasurement Set), a comprehensive evaluation dataset specifically designed to assess the capabilities of retrieval-augmented generation (RAG) systems built on LLMs. FRAMES addresses a critical need by unifying evaluations of factual accuracy, retrieval effectiveness, and reasoning ability in an end-to-end framework, rather than assessing these facets in isolation. The dataset comprises 824 challenging multi-hop questions spanning diverse topics, including history, sports, science, and health, each requiring the integration of information from between two and fifteen Wikipedia articles. By labeling questions with specific reasoning types, such as numerical or tabular. FRAMES provides a nuanced benchmark to identify the strengths and weaknesses of current RAG implementations. Baseline experiments reveal that state-of-the-art models like Gemini-Pro-1.5-0514 achieve only 40\% accuracy when operating without retrieval mechanisms, but their performance increases significantly to 66\% with a multi-step retrieval pipeline, representing a greater than 50\% improvement.

\subsection{DABStep benchmark}

DabStep \cite{huggingface2025dabstep} is a new framework from Hugging Face that pioneers a step-based approach to enhance the performance and efficiency of language models on multi-step reasoning tasks. DabStep addresses the challenges of traditional end-to-end inference by decomposing complex problem-solving into discrete, manageable steps, enabling models to refine their outputs through step-level feedback and iterative dynamic adjustments. This method is designed to enable models to self-correct and navigate the complexities of multi-step reasoning processes more effectively. However, despite these innovative improvements, experimental results reveal that even the best-performing model under this framework only achieves a 16\% success rate on the evaluated tasks. This modest accuracy underscores the significant challenges that remain in effectively training models for complex, iterative reasoning and highlights the need for further research and optimization.

\subsection{BFCL v2 benchmark}

Mao et al. \cite{mao2025bfcl} propose BFCL v2, a novel benchmark and leaderboard designed to evaluate large language models' function-calling abilities using real-world, user-contributed data. The benchmark comprises 2,251 question-function-answer pairs, enabling comprehensive assessments across a range of scenarios from multiple and straightforward function calls to parallel executions and irrelevance detection. By leveraging authentic user interactions, BFCL v2 addresses prevalent issues such as data contamination, bias, and limited generalization in previous evaluation methods. Initial evaluations reveal that models like Claude 3.5 and GPT-4 consistently outperform others, with Mistral, Llama 3.1 FT, and Gemini following in performance. However, some open models, such as Hermes, struggle due to potential prompting and formatting challenges. Overall, BFCL v2 offers a rigorous and diverse platform for benchmarking the practical capabilities of LLMs in interfacing with external tools and APIs, thereby providing valuable insights for future advancements in function calling and interactive AI systems.

\subsection{SWE-Lancer benchmark}

OpenAI team \cite{miserendino2025swelancerfrontierllmsearn} presents SWE-Lancer, an innovative benchmark comprised of over 1,400 freelance software engineering tasks collected from Upwork, representing more than \$1 million in real-world payouts. This benchmark encompasses both independent engineering tasks, ranging from minor bug fixes to substantial feature implementations valued up to \$32,000, and managerial tasks, where models must select the best technical proposals. Independent tasks are rigorously evaluated using end-to-end tests that have been triple-verified by experienced engineers. At the same time, managerial decisions are benchmarked against the selections made by the original hiring managers. Experimental results indicate that state-of-the-art models, such as Claude 3.5 Sonnet, still struggle with the majority of these tasks, achieving a 26.2\% pass rate on independent tasks and 44.9\% on managerial tasks, which translates to an estimated earning of \$403K a figure well below the total available value. Notably, the analysis highlights that while models tend to perform better in evaluative managerial roles than in direct code implementation, increasing inference-time computing can enhance performance. 

\subsection{Comprehensive RAG Benchmark (CRAG)}

Yang et al. \cite{yang2024crag} propose the Comprehensive RAG Benchmark (CRAG), a novel dataset designed to evaluate the factual question-answering capabilities of Retrieval-Augmented Generation systems rigorously. CRAG comprises 4,409 question-answer pairs across five domains and eight distinct question categories. It incorporates mock APIs to simulate web and Knowledge Graph retrieval, thereby reflecting the varied levels of entity popularity and temporal dynamism encountered in real-world scenarios. Empirical results show that state-of-the-art large language models without grounding achieve only around 34\% accuracy on CRAG, and that incorporating simple RAG methods improves this to just 44\%, whereas industry-leading RAG systems can reach 63\% accuracy without hallucination. The benchmark also highlights significant performance drops for questions involving highly dynamic, lower-popularity, or more complex facts. Notably, CRAG focuses solely on evaluating the generative component of the RAG pipeline, and early findings indicate that Llama 3 70B nearly matches GPT-4 Turbo across these tasks.

\subsection{OCCULT Benchmark}

Kouremetis et al. \cite{kouremetis2025occultevaluatinglargelanguage} present OCCULT, a novel and lightweight operational evaluation framework that rigorously measures the cybersecurity risks associated with using large language models (LLMs) for offensive cyber operations (OCO). Traditionally, evaluating AI in cybersecurity has relied on simplistic, all-or-nothing tests such as capture-the-flag exercises, which fail to capture the nuanced threats faced by modern infrastructure. In contrast, OCCULT enables cybersecurity experts to craft repeatable and contextualized benchmarks by simulating real-world threat scenarios. The authors detail three distinct OCO benchmarks designed to assess the capability of LLMs to execute adversarial tactics, providing preliminary evaluation results that indicate a significant advancement in AI-enabled cyber threats. Most notably, the DeepSeek-R1 model correctly answered over 90\%  of questions in the Threat Actor Competency Test for LLMs (TACTL). 

\subsection{DIA benchmark}

Dynamic Intelligence Assessment (DIA) \cite{tihanyi2024dynamic} is introduced as a novel methodology to more rigorously test and compare the problem-solving abilities of AI models across diverse domains such as mathematics, cryptography, cybersecurity, and computer science. Unlike traditional benchmarks that rely on static question-answer pairs, often allowing models to perform uniformly well or rely on memorization, DIA employs dynamic question templates with mutable parameters, presented in various formats including text, PDFs, compiled binaries, visual puzzles, and CTF-style challenges. This framework also introduces four innovative metrics to evaluate a model’s reliability and confidence across multiple attempts, revealing that even simple questions are frequently answered incorrectly when posed in different forms. Notably, the evaluation shows that while API models like GPT‑4o may overestimate their mathematical capabilities, models such as ChatGPT‑4o perform better due to practical tool usage, and OpenAI’s o1-mini excels in self-assessment of task suitability. Testing 25 state-of-the-art LLMs with DIA-Bench reveals significant gaps in handling complex tasks and in adaptive intelligence, establishing a new standard for evaluating both problem-solving performance and a model’s ability to recognize its own limitations.

\subsection{CyberMetric benchmark}

Tihanyi et al. \cite{tihanyi2024cybermetric} introduce a suite of novel multiple-choice Q\&A benchmark datasets, CyberMetric-80, CyberMetric-500, CyberMetric-2000, and CyberMetric-10000, designed to evaluate the cybersecurity knowledge of LLMs rigorously. By leveraging GPT-3.5 and Retrieval-Augmented Generation (RAG), the authors generated questions from diverse cybersecurity sources such as NIST standards, research papers, publicly accessible books, and RFCs. Complete with four possible answers, each question underwent extensive rounds of error checking and refinement, with over 200 hours of human expert validation to ensure accuracy and domain relevance. Evaluations were conducted on 25 state-of-the-art large language models (LLMs), and the results were further benchmarked against human performance on CyberMetric-80 in a closed-book scenario. Findings reveal that models like GPT-4o, GPT-4-turbo, Mixtral-8x7 B-Instruct, Falcon-180 B-Chat, and GEMINI-pro 1.0 exhibit superior cybersecurity understanding, outperforming humans on CyberMetric-80, while smaller models such as Llama-3-8B, Phi-2, and Gemma-7b lag behind, underscoring the value of model scale and domain-specific data in this challenging field.

\subsection{BIG-Bench Extra Hard}

A team from Google DeepMind \cite{kazemi2025big} addresses a critical gap in evaluating large language models by tackling the limitations of current reasoning benchmarks, which have primarily focused on mathematical and coding tasks. While the BIG-Bench dataset \cite{srivastava2022beyond} and its more complex variant, BIG-Bench Hard (BBH) \cite{suzgun2022challenging}, have provided comprehensive assessments of general reasoning abilities, recent advances in LLMs have led to saturation, with state-of-the-art models achieving near-perfect scores on many BBH tasks. To overcome this, the authors introduce BIG-Bench Extra Hard (BBEH). This novel benchmark replaces each BBH task with a more challenging variant designed to probe similar reasoning capabilities at an elevated difficulty level. Evaluations on BBEH reveal that even the best general-purpose models only achieve an average accuracy of 9.8\%, while reasoning-specialized models reach 44.8\%, highlighting substantial room for improvement and underscoring the ongoing challenge of developing LLMs with robust, versatile reasoning skills.

\subsection{MultiAgentBench benchmark}

Zhu et al. \cite{zhu2025multiagentbench} introduce MultiAgentBench, a benchmark specifically designed to evaluate the capabilities of multi-agent systems powered by LLMs in dynamic, interactive environments. Unlike traditional benchmarks that focus on single-agent performance or narrow domains, MultiAgentBench encompasses six diverse domains, including research proposal writing, Minecraft structure building, database error analysis, collaborative coding, competitive Werewolf gameplay, and resource bargaining to measure both task completion and the quality of agent coordination using milestone-based performance indicators. The study investigates various coordination protocols, such as star, chain, tree, and graph topologies, and finds that direct peer-to-peer communication and cognitive planning are particularly effective evidenced by a 3\% improvement in milestone achievement when planning is employed while also noting that adding more agents can decrease performance. Among the models evaluated (GPT-4o-mini, 3.5, and Llama), GPT-4o-mini achieved the highest average task score, and graph-based coordination protocols outperformed other structures in research scenarios.

\subsection{GAIA Benchmark}

GAIA \cite{mialon2023gaia} is a groundbreaking benchmark designed to assess General AI Assistants on real-world questions that tap into fundamental abilities like reasoning, multi-modality handling, web browsing, and tool use. Unlike traditional benchmarks that focus on increasingly specialized tasks, GAIA features conceptually simple questions solvable by humans at 92\% accuracy that current systems, such as GPT-4 with plugins, struggle with, achieving only 15\%. Comprising 466 meticulously curated questions with reference answers, GAIA shifts the evaluation paradigm toward measuring AI robustness in everyday reasoning tasks, a critical step toward achieving true Artificial General Intelligence (AGI). This substantial performance gap between humans and state-of-the-art models emphasizes the need for AI systems that can mimic the general-purpose, resilient reasoning exhibited by average human problem solvers.

\subsection{CASTLE Benchmark}

Dubniczky et al. \cite{dubniczky2025castle} introduce CASTLE, a novel benchmarking framework for evaluating software vulnerability detection methods, addressing existing approaches' critical weaknesses. CASTLE assesses 13 static analysis tools, 10 LLMs, and two formal verification tools using a meticulously curated dataset of 250 micro-benchmark programs that cover 25 common CWEs. The framework proposes a new evaluation metric, the CASTLE Score, to enable fair comparisons across different methods. Results reveal that while formal verification tools like ESBMC minimize false positives, they struggle with vulnerabilities beyond the scope of model checking. Static analyzers often generate excessive false positives, which burden developers with manual validation. LLMs perform strongly on small code snippets; however, their accuracy declines, and hallucinations increase as code size grows. These findings suggest that, despite current limitations, LLMs hold significant promise for integration into code completion frameworks, providing real-time vulnerability prevention and marking an important step toward more secure software systems.

\subsection{SPIN-Bench Benchmark}

Yao et al. \cite{yao2025spin} introduce a comprehensive evaluation framework, SPIN-Bench, highlighting the challenges of strategic planning and social reasoning in AI agents. Unlike traditional benchmarks focused on isolated tasks, SPIN-Bench combines classical planning, competitive board games, cooperative card games, and negotiation scenarios to simulate real-world social interactions. This multifaceted approach reveals significant performance bottlenecks in current large language models (LLMs), which, while adept at factual retrieval and short-range planning, struggle with deep multi-hop reasoning, spatial inference, and socially coordinated decision-making. For instance, models perform reasonably well on simple tasks like Tic-Tac-Toe but falter in complex environments such as Chess or Diplomacy, and even the best models achieve only around 58.59\% accuracy on classical planning tasks.

\subsection{$\tau$-bench} 

Yao et al. \cite{yao2024tau} present \(\tau\)-bench, a benchmark designed to evaluate language agents in realistic, dynamic, multi-turn conversational settings that emulate real-world environments. In \(\tau\)-bench, agents are challenged to interact with a simulated user to understand needs, utilize domain-specific API tools (such as booking flights or returning items), and adhere to provided policy guidelines, while performance is measured by comparing the final database state with an annotated goal state. A novel metric, \(\text{pass}^k\), is introduced to assess reliability over multiple trials. Experimental findings reveal that even state-of-the-art function-calling agents like GPT-4o succeed on less than 50\% of tasks, with significant inconsistency (for example, pass\(^8\) scores below 25\% in retail domains) and markedly lower success rates for tasks requiring multiple database writes. These results underscore the need for enhanced methods that improve consistency, adherence to rules, and overall reliability in language agents for real-world applications.

\begin{figure*}[htbp]
\centering
\resizebox{0.8\textwidth}{!}{%
\begin{tikzpicture}[
  mindmap,
  every node/.style={concept, circular drop shadow, minimum size=0.1cm},
  grow cyclic, align=flush center, concept color=black!50,
  level 1/.append style={
    sibling angle=45,
    level distance=12cm,
    font=\large
},
level 2/.append style={
    sibling angle=18,
    level distance=6cm,
    font=\small
  }
]
\node[concept color=black!50] {\Large LLM Benchmark}
  child[concept color=red!50] { node[align=center] {Academic \&\\General Knowledge Reasoning}
    child { node[align=center] {DROP\\\cite{dua2019drop}} }
    child { node[align=center] {MMLU\\\cite{hendrycks2020measuring}} }
    child { node[align=center] {BIG-Bench Extra Hard\\\cite{kazemi2025big}} }
    child { node[align=center] {Humanity's Last Exam\\\cite{phan2025humanity}} }
    child { node[align=center] {DABStep\\\cite{huggingface2025dabstep}} }
  }
  child[concept color=orange!50] { node[align=center] {Mathematical\\Problem Solving}
    child { node[align=center] {MATH\\\cite{hendrycks2021measuring}} }
    child { node[align=center] {MGSM\\\cite{shi2022language}} }
    child { node[align=center] {ProcessBench\\\cite{zheng2024processbench}} }
  }
  child[concept color=green!50] { node[align=center] {Code \& Software\\Engineering}
    child { node[align=center] {Codex\\\cite{chen2021evaluating}} }
    child { node[align=center] {Agent-as-a-Judge\\\cite{zhuge2024agent}} }
    child { node[align=center] {AppWorld\\\cite{trivedi2024appworld}} }
    child { node[align=center] {SciReplicate-Bench\\\cite{xiang2025scireplicate}} }
    child { node[align=center] {ProjectEval\\\cite{liu2025projecteval}} }
    child { node[align=center] {RefactorBench\\\cite{gautam2025refactorbench}} }
    child { node[align=center] {SWE-Lancer\\\cite{miserendino2025swelancerfrontierllmsearn}} }
    child { node[align=center] {CASTLE\\\cite{dubniczky2025castle}} }
    child { node[align=center] {SWE-PolyBench\\\cite{rashid2025swepolybenchmultilanguagebenchmarkrepository}} }
    child { node[align=center] {MLE-bench\\\cite{chan2025mle}} }
    child { node[align=center] {ComplexFuncBench\\\cite{zhong2025complexfuncbench}} }
    child { node[align=center] {BFCL v2\\\cite{mao2025bfcl}} }
  }
  child[concept color=yellow!50] { node[align=center] {Factual Grounding \&\\Retrieval}
    child { node[align=center] {FACTS Grounding\\\cite{deepmind2023facts}} }
    child { node[align=center] {SimpleQA\\\cite{openai2023simpleqa}} }
    child { node[align=center] {FRAMES\\\cite{krishna2024fact}} }
    child { node[align=center] {CRAG\\\cite{yang2024crag}} }
    child { node[align=center] {GPQA\\\cite{rein2024gpqa}} }
  }
  child[concept color=teal!50] { node[align=center] {Domain-Specific\\Evaluations}
    child { node[align=center] {MedChain\\\cite{liu2024medchain}} }
    child { node[align=center] {LegalAgentBench\\\cite{li2024legalagentbench}} }
    child { node[align=center] {MedAgentsBench\\\cite{tang2025medagentsbench}} }
    child { node[align=center] {CyberMetric\\\cite{tihanyi2024cybermetric}} }
    child { node[align=center] {OCCULT\\\cite{kouremetis2025occultevaluatinglargelanguage}} }
    child { node[align=center] {EconAgentBench\\\cite{fish2025econevals}} }
  }
  child[concept color=purple!50] { node[align=center] {Multimodal, Visual \&\\Embodied Evaluations}
    child { node[align=center] {GAIA\\\cite{mialon2023gaia}} }
    child { node[align=center] {OmniDocBench\\\cite{ouyang2024omnidocbench}} }
    child { node[align=center] {Dyn-VQA\\\cite{li2024benchmarking}} }
    child { node[align=center] {DIA\\\cite{tihanyi2024dynamic}} }
    child { node[align=center] {OlympicArena\\\cite{huang2024olympicarena}} }
    child { node[align=center] {BEARCUBS\\\cite{song2025bearcubs}} }
    child { node[align=center] {EmbodiedEval\\\cite{cheng2025embodiedeval}} }
    child { node[align=center] {EmbodiedBench\\\cite{yang2025embodiedbench}} }
    child { node[align=center] {ENIGMAEVAL\\\cite{wang2025enigmaeval}} }
    child { node[align=center] {TheoremExplainBench\\\cite{ku2025theoremexplainagent}} }
    child { node[align=center] {VisualAgentBench\\\cite{liu2024visualagentbench}} }
  }
  child[concept color=gray!50] { node[align=center] {Task Selection}
    child { node[align=center] {FineTasks\\\cite{huggingfacefw2023finetasks}} }
    child { node[align=center] {Multi-SWE-bench\\\cite{zan2025multi}} }
  }
  child[concept color=pink!50] { node[align=center] {Agentic \& Interactive\\Evaluations}
    child { node[align=center] {PersonaGym\\\cite{samuel2024personagym}} }
    child { node[align=center] {MIRAI\\\cite{ye2024mirai}} }
    child { node[align=center] {ScienceAgentBench\\\cite{chen2024scienceagentbench}} }
    child { node[align=center] {Agent-SafetyBench\\\cite{zhang2024agent}} }
    child { node[align=center] {DiscoveryBench\\\cite{majumder2024discoverybench}} }
    child { node[align=center] {BLADE\\\cite{gu2024blade}} }
    child { node[align=center] {JudgeBench\\\cite{tan2024judgebench}} }
    child { node[align=center] {TeamCraft\\\cite{long2024teamcraft}} }
    child { node[align=center] {AgentHarm\\\cite{andriushchenko2024agentharm}} }
    child { node[align=center] {\(\tau\)-bench\\\cite{yao2024tau}} }
    child { node[align=center] {MultiAgentBench\\\cite{zhu2025multiagentbench}} }
    child { node[align=center] {SPIN-Bench\\\cite{yao2025spin}} }
    child { node[align=center] {VeriLA\\\cite{sung2025verila}} }
    child { node[align=center] {CapaBench\\\cite{yang2025s}} }
    child { node[align=center] {AgentOrca\\\cite{li2025agentorca}} }
    child { node[align=center] {Robotouille\\\cite{gonzalez2025robotouille}} }
    child { node[align=center] {DSGBench\\\cite{tang2025dsgbench}} }
    child { node[align=center] {RefuteBench 2.0\\\cite{yan2025refutebench}} }
    child { node[align=center] {MLGym\\\cite{nathani2025mlgym}} }
    child { node[align=center] {DataSciBench\\\cite{zhang2025datascibench}} }
    child { node[align=center] {BrowseComp\\\cite{wei2025browsecomp}} }
  }
;
\end{tikzpicture}
}
\caption{Classification of LLM Benchmarks for AI Agents Applications}
\label{fig:llm_benchmark_diagram}
\end{figure*}
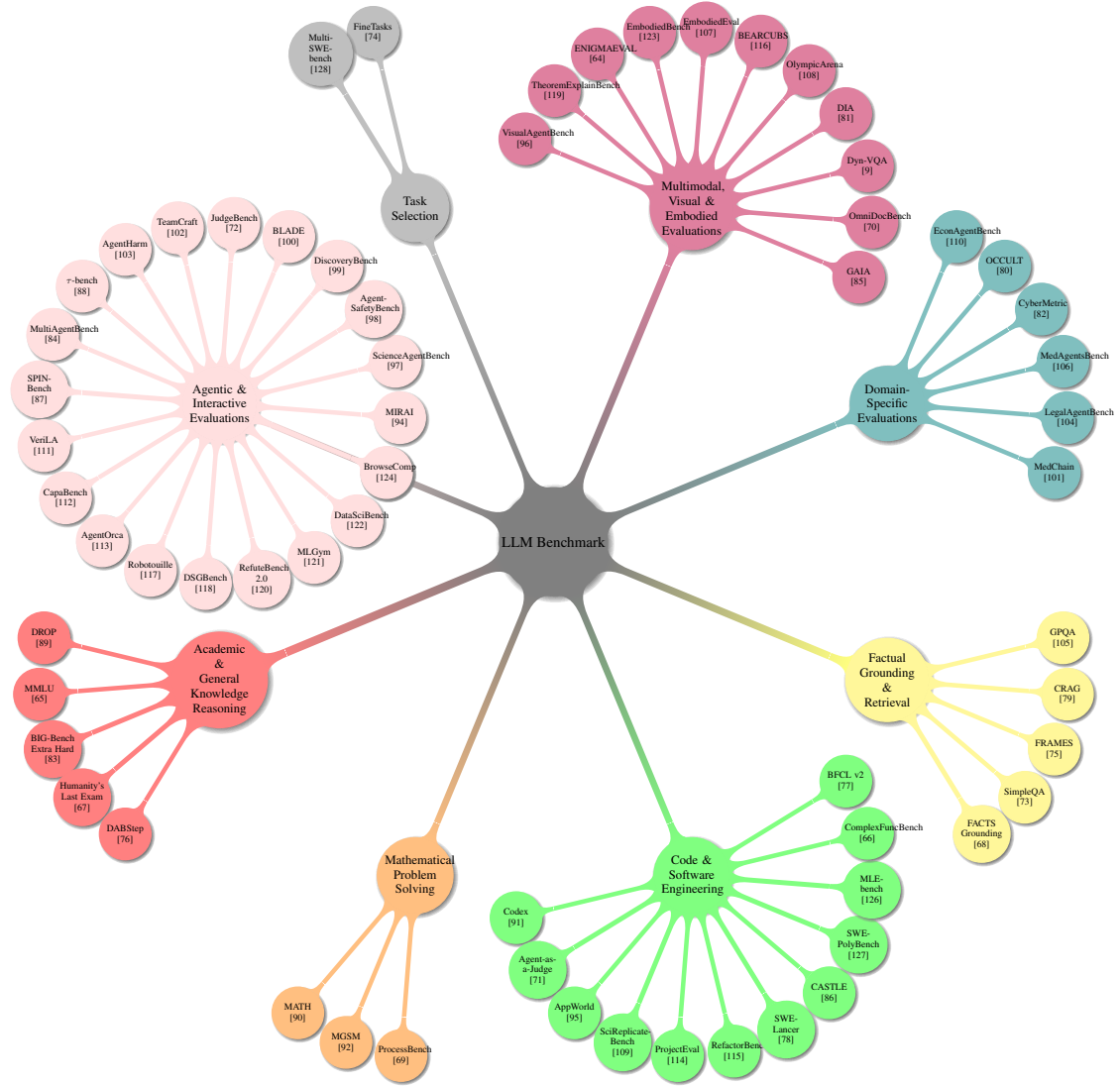

\subsection{Discussion and Comparison of LLM Benchmarks}

Table~\ref{tab:llm_benchmark} presents an extensive overview of benchmarks developed from 2019 to 2025 for evaluating large language models (LLMs) concerning multimodal capabilities, task scope, diversity, reasoning, and agentic behaviors. Early benchmarks, such as DROP~\cite{dua2019drop}, MMLU~\cite{hendrycks2020measuring}, MATH~\cite{hendrycks2021measuring}, Codex~\cite{chen2021evaluating}, MGSM~\cite{shi2022language}, FACTS Grounding~\cite{deepmind2023facts}, and SimpleQA~\cite{openai2023simpleqa}, concentrated on core competencies like discrete reasoning, academic knowledge, mathematical problem solving, and factual grounding. These pioneering efforts lay the groundwork for performance evaluation in language understanding and reasoning tasks, setting a baseline against which later, more sophisticated benchmarks have been compared.

A notable progression in benchmark design is observed with the emergence of frameworks that target more complex agentic and multimodal tasks. For instance, PersonaGym~\cite{samuel2024personagym} and FineTasks~\cite{huggingfacefw2023finetasks} introduce dynamic persona evaluation and multilingual task selection. GAIA~\cite{mialon2023gaia} expands the evaluative scope to general AI assistant tasks while OmniDocBench~\cite{ouyang2024omnidocbench} and ProcessBench~\cite{zheng2024processbench} address document extraction and error detection in mathematical solutions. Further, MIRAI~\cite{ye2024mirai}, AppWorld~\cite{trivedi2024appworld}, VisualAgentBench~\cite{liu2024visualagentbench}, and ScienceAgentBench~\cite{chen2024scienceagentbench} explore various facets of multimodal and scientific discovery tasks. This decade-spanning evolution is complemented by additional evaluations focusing on safety (Agent-SafetyBench~\cite{zhang2024agent}), discovery (DiscoveryBench~\cite{majumder2024discoverybench}), code generation (BLADE~\cite{gu2024blade}, Dyn-VQA~\cite{li2024benchmarking}, and Agent-as-a-Judge~\cite{zhuge2024agent}), judicial reasoning (JudgeBench~\cite{tan2024judgebench}), and clinical decision making (MedChain~\cite{liu2024medchain}), among others including FRAMES~\cite{krishna2024fact}, CRAG~\cite{yang2024crag}, DIA~\cite{tihanyi2024dynamic}, CyberMetric~\cite{tihanyi2024cybermetric}, TeamCraft~\cite{long2024teamcraft}, AgentHarm~\cite{andriushchenko2024agentharm}, \(\tau\)-bench~\cite{yao2024tau}, LegalAgentBench~\cite{li2024legalagentbench}, and GPQA~\cite{rein2024gpqa}.

Recent benchmarks from 2025 further indicate a substantial expansion in the depth and breadth of large language model (LLM) evaluations. ENIGMAEVAL~\cite{wang2025enigmaeval} and ComplexFuncBench~\cite{zhong2025complexfuncbench} target complex puzzles and function calling tasks, while MedAgentsBench~\cite{tang2025medagentsbench} and Humanity's Last Exam~\cite{phan2025humanity} focus on advanced medical reasoning and expert-level academic tasks. Additional benchmarks such as DABStep~\cite{huggingface2025dabstep}, BFCL v2~\cite{mao2025bfcl}, SWE-Lancer~\cite{miserendino2025swelancerfrontierllmsearn}, and OCCULT~\cite{kouremetis2025occultevaluatinglargelanguage} further diversify evaluative criteria by incorporating multi-step reasoning, cybersecurity, and freelance software engineering challenges. The table also includes BIG-Bench Extra Hard~\cite{kazemi2025big}, MultiAgentBench~\cite{zhu2025multiagentbench}, CASTLE~\cite{dubniczky2025castle}, EmbodiedEval~\cite{cheng2025embodiedeval}, SPIN-Bench~\cite{yao2025spin}, OlympicArena~\cite{huang2024olympicarena}, SciReplicate-Bench~\cite{xiang2025scireplicate}, EconAgentBench~\cite{fish2025econevals}, VeriLA~\cite{sung2025verila}, CapaBench~\cite{yang2025s}, AgentOrca~\cite{li2025agentorca}, ProjectEval~\cite{liu2025projecteval}, RefactorBench~\cite{gautam2025refactorbench}, BEARCUBS~\cite{song2025bearcubs}, Robotouille~\cite{gonzalez2025robotouille}, DSGBench~\cite{tang2025dsgbench}, TheoremExplainBench~\cite{ku2025theoremexplainagent}, RefuteBench 2.0~\cite{yan2025refutebench}, MLGym~\cite{nathani2025mlgym}, DataSciBench~\cite{zhang2025datascibench}, EmbodiedBench~\cite{yang2025embodiedbench}, BrowseComp~\cite{wei2025browsecomp}, and MLE-bench~\cite{chan2025mle}. Collectively, these benchmarks exemplify the field’s shift towards more comprehensive and nuanced evaluation metrics, supporting the development of LLMs that can tackle increasingly multifaceted, real-world challenges.

Fig. \ref{fig:llm_benchmark_diagram} groups benchmarks into categories such as Academic \& General Knowledge Reasoning, Mathematical Problem Solving, Code \& Software Engineering, Factual Grounding \& Retrieval, Domain‐Specific Evaluations, Multimodal/Visual \& Embodied Evaluations, Task Selection, and Agentic \& Interactive Evaluations, illustrating the full range of tasks used to assess LLMs in AI agent settings.

\begin{table*}[htbp]
\centering
\scriptsize
\caption{Overview of AI Agent Frameworks: Core Concepts, Workflow, and Advantages}
\label{tab:ai_agent_frameworks}
\begin{adjustbox}{max width=\textwidth}
\begin{tabularx}{\textwidth}{@{}%
>{\raggedright\arraybackslash}p{3.2cm}
>{\raggedright\arraybackslash}X
>{\raggedright\arraybackslash}X
>{\raggedright\arraybackslash}X
@{}}
\toprule
\textbf{Agent Framework} & \textbf{Core Idea} & \textbf{Workflow \& Components} & \textbf{Key Advantages} \\
\midrule
LangChain \cite{langchain_agents} & Integrates LLMs with diverse tools to build autonomous agents. & Combines conversational LLMs, search integrations, and utility functions into iterative workflows. & Customizable roles and streamlined agent prototyping. \\
\midrule
LlamaIndex \cite{llamaindex_agent} & Enables autonomous agent creation via external tool integration. & Wraps functions into \texttt{FunctionTool} objects and employs a ReActAgent for stepwise tool selection. & Simplifies agent development with a dynamic, modular pipeline. \\
\midrule
CrewAI \cite{crewai} & Orchestrates teams of specialized AI agents for complex tasks. & Structures systems into Crew (oversight), AI Agents (specialized roles), Process (collaboration), and Tasks (assignments). & Mimics human team collaboration with flexible, parallel workflows. \\
\midrule
Swarm \cite{openai_swarm2023} & Provides a lightweight, stateless abstraction for multi-agent systems. & Defines multiple agents with specific instructions and roles; enables dynamic handoffs and context management. & Fine-grained control and compatibility with various backends. \\
\midrule
GUI Agent \cite{hu2024dawn} & Facilitates computer control via natural language and visual inputs. & Translates user instructions and screenshots into desktop actions (e.g., cursor movements, clicks). & Demonstrates end-to-end performance in real-world desktop workflows. \\
\midrule
Agentic Reasoning \cite{wu2025agentic} & Enhances reasoning by integrating specialized external tool-using agents. & Leverages web-search, coding, and Mind Map agents to iteratively refine multi-step reasoning. & Achieves improved multi-step problem-solving and structured knowledge synthesis. \\
\midrule
OctoTools \cite{lu2025octotools} & Empowers LLMs for complex reasoning via training-free tool integration. & Combines standardized tool cards, a strategic planner, and an executor for effective tool usage. & Outperforms similar frameworks by up to 10.6\% on varied tasks. \\
\midrule
Agents SDK \cite{openai_agents_sdk} & Provides a modular framework for building autonomous agent applications that integrate LLMs with external tools and advanced features. & Offers core primitives such as Agents (LLMs with instructions, tools, handoffs, and guardrails), Tools (wrapped functions/APIs), Context for state management, along with support for Streaming, Tracing, and Guardrails to manage multi-turn interactions. & Streamlines development with an extensible, robust architecture that enhances debuggability and scalability, enabling rapid prototyping and seamless integration of complex, multi-agent workflows. \\
\bottomrule
\end{tabularx}
\end{adjustbox}
\end{table*}

\section{AI Agents}\label{sec:4}

This section presents a comprehensive overview of AI agent frameworks and applications developed between 2024 and 2025, highlighting transformative approaches that integrate large language models with modular tools to achieve autonomous decision-making and dynamic multi-step reasoning. The frameworks discussed include LangChain \cite{langchain_agents}, LlamaIndex \cite{llamaindex_agent}, CrewAI \cite{crewai}, and Swarm \cite{openai_swarm2023}, which abstract complex functionalities into reusable components that enable context management, tool integration, and iterative refinement of outputs. Additionally, pioneering efforts in GUI control \cite{hu2024dawn} and agentic reasoning \cite{wu2025agentic, lu2025octotools} demonstrate the increasing capabilities of these systems to interact with external environments and tools in real-time.

In parallel, this section presents a diverse range of AI agent applications that span materials science, biomedical research, academic ideation, software engineering, synthetic data generation, and chemical reasoning. Systems such as the StarWhisper Telescope System \cite{wang2024starwhisper} and HoneyComb \cite{zhang2024honeycomb} have revolutionized operational workflows by automating observational and analytical tasks in materials science. In the biomedical domain, platforms like GeneAgent \cite{wang2024geneagent} and frameworks such as PRefLexOR \cite{buehler2024preflexor} demonstrate enhanced reliability through self-verification and iterative refinement. Moreover, innovative solutions for research ideation, exemplified by SurveyX \cite{liang2025surveyx} and Chain-of-Ideas \cite{li2024chain}, as well as specialized frameworks for synthetic data generation \cite{mitra2024agentinstruct} and chemical reasoning \cite{tang2025chemagent}, collectively underscore the significant strides made in leveraging autonomous AI agents for complex, real-world tasks. Table \ref{tab:ai_agent_frameworks} presents an overview of AI Agent frameworks.

\begin{figure}[t]
    \centering
    \includegraphics[width=1\linewidth]{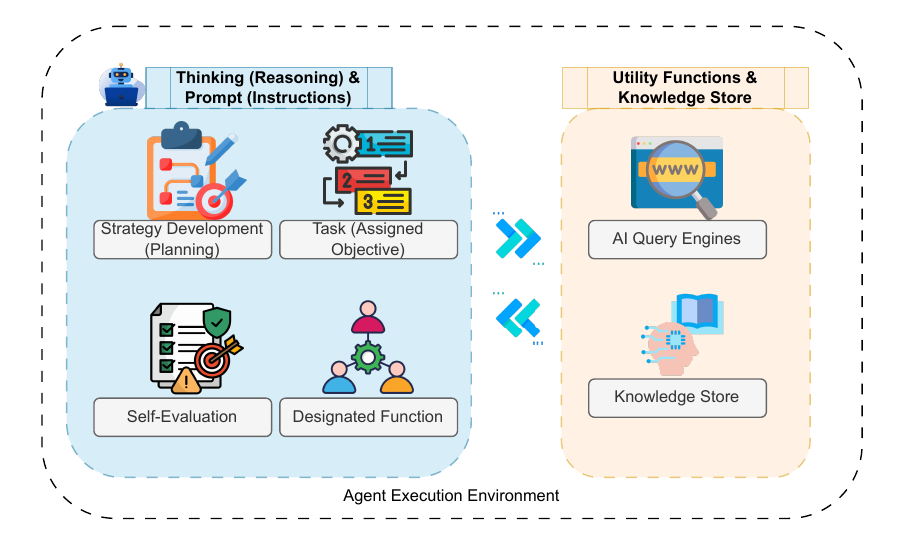}
    \caption{Core Elements of AI Agents.}
    \label{fig:agentelment}
\end{figure}

\subsection{AI Agent frameworks}

AI agent frameworks represent a transformative paradigm in developing intelligent systems, combining the power of large language models with modular tools and utilities to build autonomous software agents. These frameworks abstract complex functionalities such as natural language understanding, multi-step reasoning, and dynamic decision-making into reusable components that streamline prototyping, iterative refinement, and deployment. By integrating advanced LLMs with external tools and specialized functions, developers can create agents that process and generate language and adapt to complex workflows and diverse operational contexts \cite{liu2025textit}.

Fig. \ref{fig:agentelment} illustrates a comprehensive AI agent framework where each component plays a crucial role in achieving adaptive, autonomous decision-making. An assigned task is first approached through a designated function that defines the agent's role, followed by strategy development, essentially the planning phase, where the agent breaks down complex objectives into actionable steps. This is supported by an iterative thinking process, driven by reasoning and guided by prompts, which enables the agent to reflect on its actions and refine its approach. Core operational support comes from AI query engines and utility functions that interface with an integrated knowledge store, ensuring that both static and real-time information is readily accessible. Ultimately, these elements operate within an agent execution environment, seamlessly combining planning, reasoning, and execution into a responsive and self-evolving system.

\begin{figure}
    \centering
    \includegraphics[width=1\linewidth]{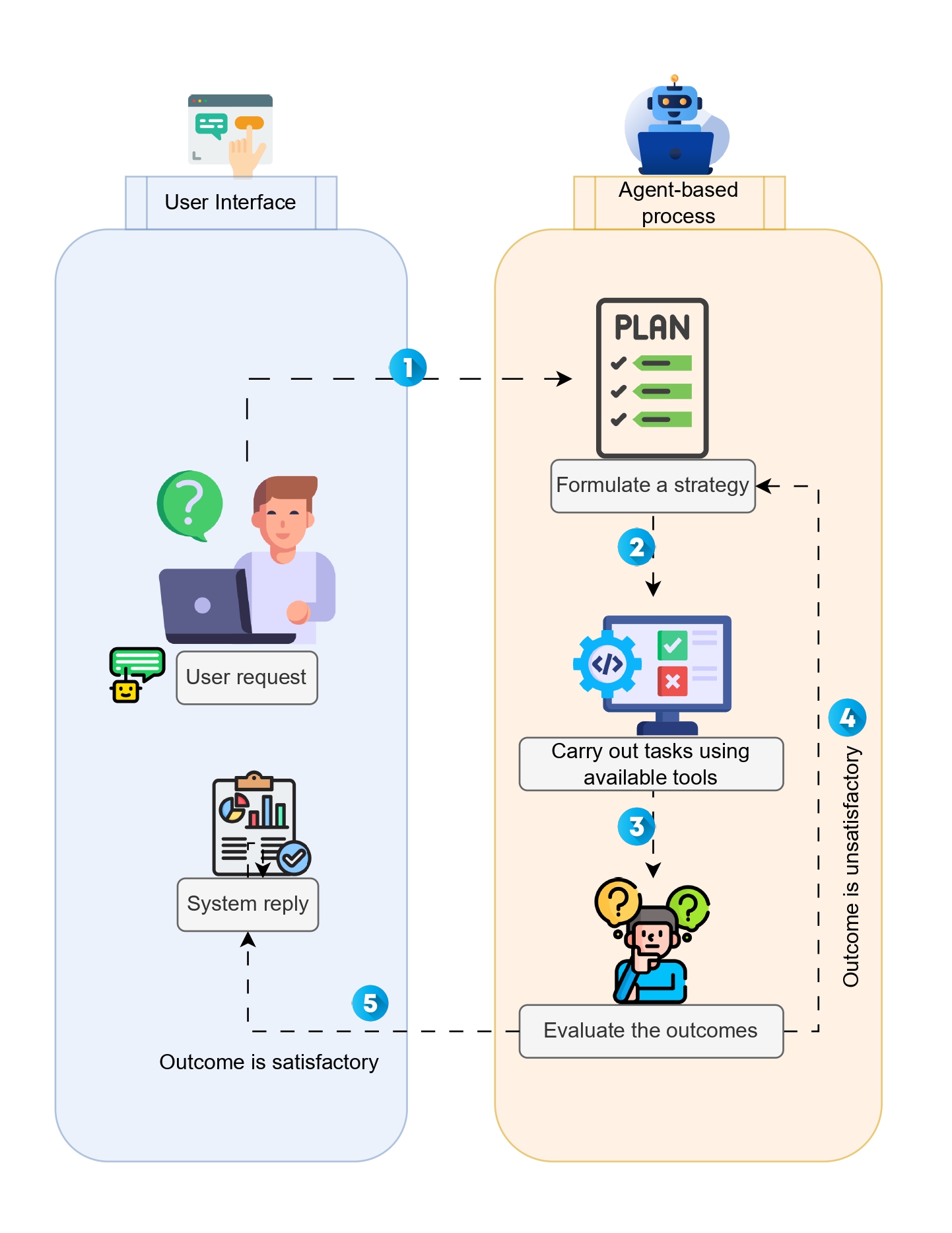}
    \caption{What are Agentic Workflows?.}
    \label{fig:agentic}
\end{figure}

Agentic workflows transform traditional, rigid processes into dynamic, adaptive systems. As illustrated in Fig. \ref{fig:agentic}, these workflows begin at the user interface, where a user query is submitted and receives a system reply. Unlike deterministic workflows that follow fixed, unchanging rules, an agent-based process involves AI agents who actively formulate a strategy, carry out tasks using available tools, and evaluate the outcomes. This cycle, ranging from planning to execution and ultimately to assessment, where outcomes are marked as either satisfactory or unsatisfactory, empowers the system to respond to real-world challenges more flexibly and autonomously \cite{zhang2024survey}.

\begin{figure}[t]
    \centering
    \includegraphics[width=1\linewidth]{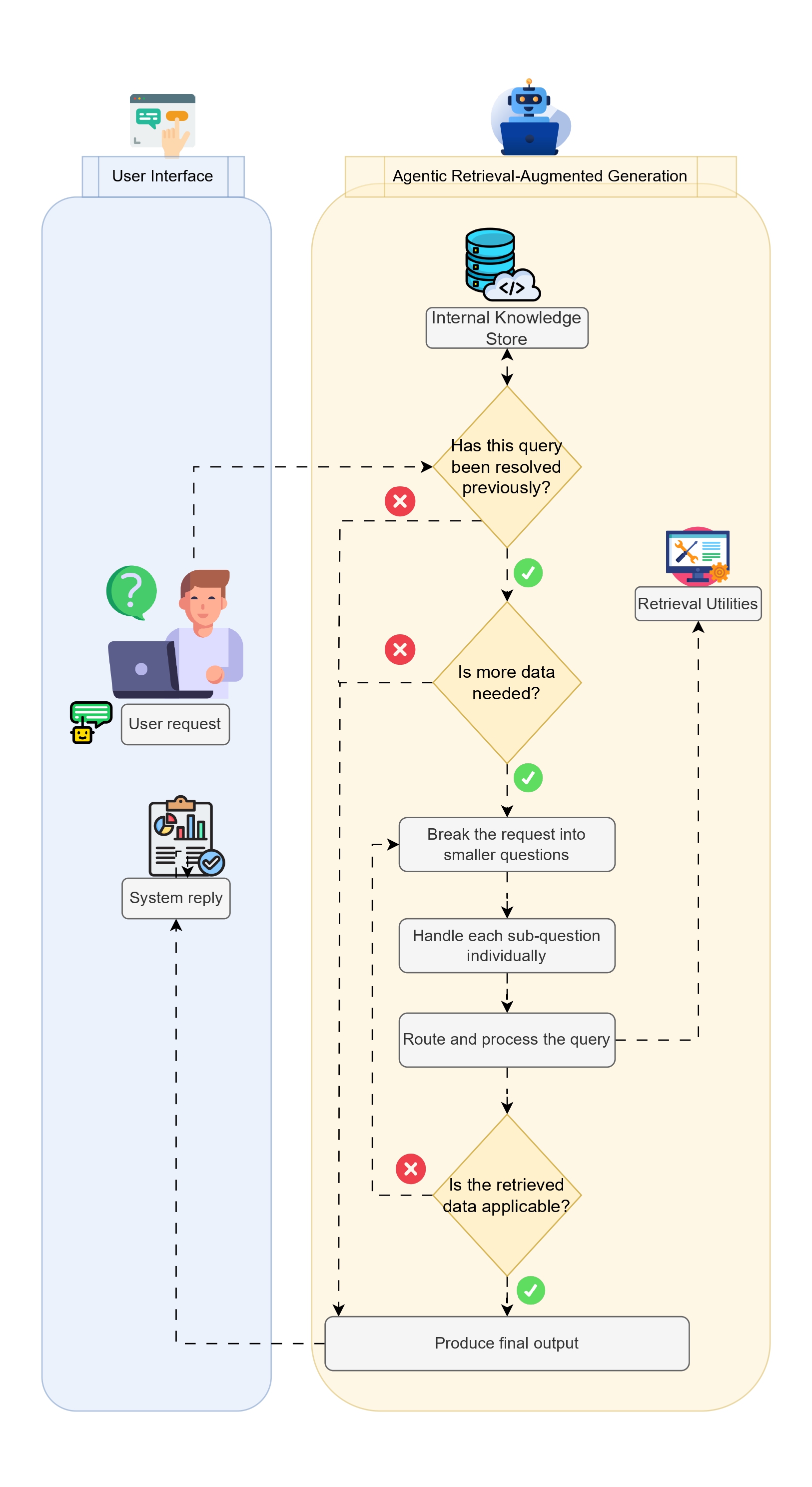}
    \caption{Agent-Driven RAG Framework.}
    \label{fig:agenticrag}
\end{figure}

\begin{table*}[htbp]
\centering
\caption{Comparative Analysis of LLM Strategies in RAG, AI Agents, and Agentic RAG}
\label{tab:comLLM}
\scriptsize
\begin{tabular}{>{\raggedright\arraybackslash}p{2.5cm}%
                >{\raggedright\arraybackslash}p{2.5cm}%
                >{\raggedright\arraybackslash}p{2.5cm}%
                >{\raggedright\arraybackslash}p{2.5cm}%
                >{\raggedright\arraybackslash}p{2.5cm}%
                >{\raggedright\arraybackslash}p{2.5cm}}
\toprule
\textbf{Feature} &
\textbf{LLM Pre-trained} &
\textbf{LLM Post Training \& Fine Tuning} &
\textbf{RAG} &
\textbf{AI Agents} &
\textbf{Agentic RAG} \\
\midrule
Core Function   & Uses LLM for text generation.  & Applies task-specific tuning.  & Retrieves data and generates text.  & Automates tasks and decisions.  & Integrates retrieval with adaptive reasoning. \\[0.5em] \hline
Autonomy        & Basic language understanding.      & Enhances autonomy through tuning.  & Limited; user-driven.       & Moderately autonomous.   & Highly autonomous. \\[0.5em] \hline
Learning        & Relies on pre-training.            & Uses fine tuning for precision.  & Static pre-trained knowledge.  & 
Incorporates user feedback.  & Adapts using real-time data. \\[0.5em] \hline
Use Cases       & General applications.              & Domain-specific enhancements.  & Q\&A, summaries, guidance.  & Chatbots, automation, workflow.  & Complex decision-making tasks. \\[0.5em] \hline
Complexity      & Provides baseline complexity.     & Adds refined capabilities.  & Simple integration.  & More sophisticated.  & Highly complex. \\[0.5em] \hline
Reliability     & Depends on static training data.   & Improves consistency with updates.  & Consistent for known queries.  & May vary with dynamic inputs.  & Reliability boosted by adaptive methods. \\[0.5em] \hline
Scalability     & Scales with model size.            & Scales with domain-specific tuning.  & Easily scalable for static tasks.  & Scales moderately with added features.  & Scalable for complex systems (with extra resources). \\[0.5em] \hline
Integration     & Easily integrable with various apps.  & Requires domain customization.  & Integrates well with retrieval systems.  & Connects with operational workflows.  & Supports advanced decision frameworks. \\
\bottomrule
\end{tabular}
\end{table*}

Agentic Retrieval-Augmented Generation (RAG) integrates a language model's advanced capabilities with dynamic data retrieval and processing. As shown in Fig. \ref{fig:agenticrag}, the process begins at the user interface, where a query is submitted and a system reply is generated. The system first checks its internal knowledge store to determine whether the query has been addressed or needs more data. When necessary, the query is decomposed into smaller, manageable sub-questions that are individually routed and processed through retrieval utilities \cite{zhao2024retrieval}. These utilities fetch relevant external data, and the system evaluates whether the retrieved information is applicable before producing a final output. This layered, agentic approach ensures that responses are accurate, context-aware, and continuously refined throughout the process \cite{liu2025agent}.

Tab. \ref{tab:comLLM} demonstrates that retrieval-augmented generation (RAG) is highly effective at producing up-to-date, accurate responses, making it ideal for fields like healthcare or law, where precise, domain-specific information is critical. In contrast, AI Agents distinguish themselves with their continuous learning and autonomous decision-making capabilities, which make them adaptable to evolving contexts. When these two approaches are combined into Agentic RAG, the model benefits from RAG's fact-based grounding and AI Agents' dynamic adaptability, resulting in a system that minimizes errors and remains current by leveraging the best aspects of each methodology.

\begin{figure}[t]
    \centering
    \includegraphics[width=1\linewidth]{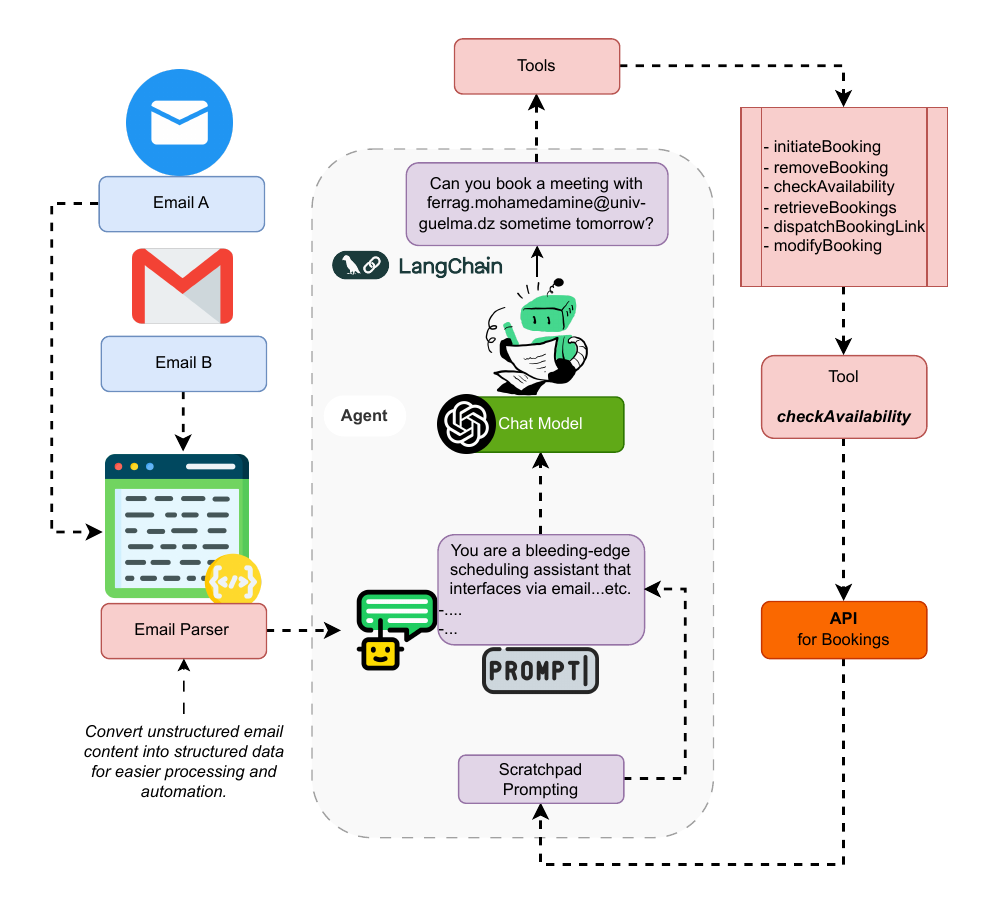}
    \caption{Agent architecture using Langchain framework.}
    \label{fig:langchain}
\end{figure}

\subsubsection{LangChain}
LangChain \cite{langchain_agents} is a robust framework designed to simplify the development of autonomous AI agents by seamlessly integrating large language models with a diverse array of tools and data sources. In LangChain, agents combine prepackaged components, such as conversational large language models (LLMs), search engine integrations, and specialized utility functions, into coherent workflows that enable multi-step reasoning and decision-making. Developers can build custom agents by defining specific roles, tasks, and tools, allowing the agent to analyze a given prompt, select the appropriate tool for each subtask, and iteratively refine its response until a final answer is produced. Fig. \ref{fig:langchain} illustrates the architecture of a LangChain-powered scheduling agent that processes email requests to perform calendar-related operations \cite{langchain2025}. Incoming emails are first parsed to extract relevant content and convert unstructured text into structured data. This data is then passed to the chat model, guided by a contextual prompt that defines the assistant's role. The agent uses a scratchpad to reason through the request and determine the appropriate tool from a predefined set (such as checkAvailability, initiateBooking, or modifyBooking). These tools interact with the backend booking API to execute the requested actions, enabling seamless AI-driven scheduling. 

\subsubsection{LlamaIndex}

The LlamaIndex framework \cite{llamaindex_agent} provides a powerful and flexible platform for building autonomous AI agents by seamlessly integrating large language models with external tools. In this framework, a basic AI agent is defined as a semi-autonomous software component that receives a task and a set of tools ranging from simple Python functions to complete query engines and iteratively selects the appropriate tool to process each step of the task. To build such an agent, developers first set up a clean Python environment and install LlamaIndex along with necessary dependencies, then configure an LLM (for example, GPT‑4 via an API key). Next, they wrap simple functions (such as addition and multiplication) into FunctionTool objects that the agent can call, and instantiate a ReActAgent with these tools. When prompted with a task, the agent evaluates its reasoning process, chooses a tool to execute the necessary operations, and loops through these steps until the final answer is generated. This structured yet dynamic approach allows for the creation of customizable, agentic workflows capable of tackling complex tasks.

\subsubsection{CrewAI}
CrewAI \cite{crewai} is a framework designed to orchestrate autonomous teams of AI agents, each with specialized roles, tools, and objectives, to collaboratively tackle complex tasks. The system is organized around four key components: the Crew, which oversees the overall operation and workflow; AI Agents, which serve as specialized team members such as researchers, writers, and analysts that make autonomous decisions and delegate tasks; the Process, which manages collaboration patterns and task assignments to ensure efficient execution; and Tasks, which are individual assignments with clear objectives that contribute to a larger goal. Key features of CrewAI include role-based agent specialization, flexible integration of custom tools and APIs, intelligent collaboration that mimics natural human interaction, and robust task management supporting both sequential and parallel workflows. Together, these elements enable the creation of dynamic, production-ready AI teams capable of achieving sophisticated, multi-step objectives in real-world applications.

\subsubsection{Swarm}

Swarm \cite{openai_swarm2023} is a lightweight, experimental library from OpenAI designed to build and manage multi-agent systems without relying on the Assistants API. Swarm provides a stateless abstraction that orchestrates a continuous loop of agent interactions, function calls, and dynamic handoffs, offering fine-grained control and transparency. Key features include: 
\begin{itemize}
    \item Agent Definition: Developers can define multiple agents, each equipped with its own set of instructions, designated role (e.g., "Sales Agent"), and available functions, which are converted into standardized JSON structures.
    \item Dynamic Handoffs: Agents can transfer control to one another based on the conversation flow or specific function criteria, simply by returning the next agent to call.
    \item Context Management: Context variables are used to initialize and update state throughout the conversation, ensuring continuity and effective information sharing across agents.
    \item Client Orchestration: The Client.run() function initiates and manages the multi-agent dialogue by taking an initial agent, user messages, and context, and then returning updated messages, context variables, and the last active agent.
    \item Direct Function Calling \& Streaming: Swarm supports direct Python function calls within agents and provides streaming responses for real-time interactions.
    \item Flexibility: The framework is designed to be agnostic to the underlying OpenAI client, working seamlessly with tools such as Hugging Face TGI or vLLM hosted models.

\end{itemize}

\subsubsection{GUI Agent}

Hu et al. \cite{hu2024dawn} introduced Claude 3.5 Computer Use, marking a significant milestone as the first frontier AI model to offer computer control via a graphical user interface in a public beta setting. The study assembles a diverse set of tasks, ranging from web search and productivity workflows to gaming and file management, to rigorously evaluate the model's ability to translate natural language instructions and screenshots into precise desktop actions, such as cursor movements, clicks, and keystrokes. The evaluation framework not only demonstrates Claude 3.5's unprecedented end-to-end performance, with a success rate of 16 out of 20 test cases, but also highlights critical areas for future refinement, including improved planning, action execution, and self-critique capabilities. Moreover, the performance is shown to be influenced by factors like screen resolution, and the study reveals that while the model can perform a wide range of operations, it still struggles with replicating subtle human-like behaviors such as natural scrolling and browsing. Overall, this preliminary exploration underscores the potential of LLMs to control computers via GUI, while also identifying the need for more comprehensive, multimodal datasets to capture real-world complexities.

The paper by Sun et al. \cite{sun2024genesis} tackles a major challenge in training GUI agents powered by Vision-Language Models (VLMs): collecting high-quality trajectory data. Traditional methods relying on human supervision or synthetic data generation via pre-defined tasks are either resource-intensive or fail to capture the complexity and diversity of real-world environments. The authors propose OS-Genesis, a novel data synthesis pipeline that reverses the conventional trajectory collection process to overcome these limitations. Rather than starting with fixed tasks, OS-Genesis enables agents to explore environments through step-by-step interactions and then derive high-quality tasks retrospectively, with a trajectory reward model ensuring data quality.

\subsubsection{Agentic Reasoning}

Wu et al. \cite{wu2025agentic} presents a novel framework that significantly enhances the reasoning capabilities of large language models by integrating external tool-using agents into the inference process. The approach leverages three key agents: a web-search agent for real-time retrieval of pertinent information, a coding agent for executing computational tasks, and a Mind Map agent that constructs structured knowledge graphs to track and organize logical relationships during reasoning. By dynamically engaging these specialized agents, the framework enables LLMs to perform multi-step, expert-level problem solving and deep research, addressing limitations in conventional internal reasoning approaches. Evaluations on challenging benchmarks such as the GPQA dataset and domain-specific deep research tasks demonstrate that Agentic Reasoning substantially outperforms traditional retrieval-augmented generation systems and closed-source models, highlighting its potential for improved knowledge synthesis, test-time scalability, and structured problem-solving.

OctoTools \cite{lu2025octotools} is a robust, training-free, and user-friendly framework designed to empower large language models to tackle complex reasoning tasks across diverse domains. By integrating standardized tool cards that encapsulate various tool functionalities, a planner for orchestrating both high-level and low-level strategies, and an executor for effective tool usage, OctoTools overcomes the limitations of prior methods that were confined to specialized domains or required extra training data. Validated across 16 varied tasks including MathVista, MMLU-Pro, MedQA, and GAIA-Text OctoTools achieves an average accuracy improvement of 9.3\% over GPT‑4o and outperforms frameworks like AutoGen, GPT‑Functions, and LangChain by up to 10.6\% when using the same toolset. Comprehensive analysis and ablation studies demonstrate its advantages in task planning, effective tool integration, and multi-step problem solving, positioning it as a significant advancement for general-purpose, complex reasoning applications.

\subsubsection{Agents SDK}

The OpenAI Agents SDK \cite{openai_agents_sdk} provides a comprehensive framework for building autonomous, multi-step agent applications that harness the power of large language models alongside external tools. This SDK abstracts the core components necessary for agentic workflows, including agents themselves which are LLMs configured with instructions, tools, handoffs, and guardrails as well as the tools that enable these agents to perform external actions (such as API calls or computations). It also supports context management to maintain state over multi-turn interactions, structured output types for reliable data exchange, and advanced features like streaming, tracing, and guardrails to ensure safety and debugability. 


\begin{table*}[htbp]
\centering
\scriptsize
\caption{Overview of AI Agent Applications for Healthcare}
\label{tab:ai_agent_apps_partII}
\begin{adjustbox}{max width=\textwidth}
\begin{tabularx}{\textwidth}{@{}%
>{\raggedright\arraybackslash}p{1.2cm}   
>{\centering\arraybackslash}p{0.3cm}      
>{\centering\arraybackslash}p{1cm}      
>{\raggedright\arraybackslash}X          
>{\raggedright\arraybackslash}X          
>{\raggedright\arraybackslash}X          
>{\centering\arraybackslash}p{0.8cm}      
>{\centering\arraybackslash}p{0.8cm}      
>{\centering\arraybackslash}p{0.8cm}      
@{}}
\toprule
\textbf{Application} & \textbf{Year} & \textbf{Category} & \textbf{Core Objective} & \textbf{Workflow \& Components} & \textbf{Key Benefits/Results} & \textbf{C} & \textbf{W} & \textbf{R} \\
\midrule
DiagnosisGPT \cite{chen2024cod} 
& 2024 
& Medical Diagnostics 
& Enhance interpretability via a transparent, step-by-step chain. 
& Implements CoD to yield confidence scores and entropy reduction. 
& Diagnoses 9,604 diseases; outperforms existing LLMs. 
& \halfcirc 
& \emptycirc 
& \emptycirc \\
\midrule
ZODIAC \cite{zhou2024zodiac}  
& 2024 
& Cardiology 
& Deliver expert-level cardiological diagnostics. 
& Multi-agent LLM fine-tuned on adjudicated patient data. 
& Outperforms leading models; integrated into ECG devices. 
& \fullcirc 
& \fullcirc 
& \fullcirc \\
\midrule
MedAgent-Pro \cite{wang2025medagent} 
& 2025 
& Medical Diagnosis 
& Enhance multi-modal diagnosis by addressing visual and reasoning gaps. 
& Hierarchical workflow with knowledge-based reasoning and multi-modal agents. 
& Outperforms existing methods on 2D/3D tasks with improved reliability. 
& \halfcirc 
& \emptycirc 
& \emptycirc \\
\midrule
Steenstra et al. \cite{steenstra2025scaffolding} 
& 2025 
& Therapeutic Counseling 
& Improve counseling training with continuous, detailed feedback. 
& LLM-powered simulated patients with turn-by-turn visualizations. 
& High usability and satisfaction; enhances learning vs. traditional methods. 
& \halfcirc 
& \halfcirc 
& \emptycirc \\
\midrule
M3Builder \cite{feng2025m} 
& 2025 
& Medical Imaging ML 
& Automate ML workflows in medical imaging. 
& Four agents manage data processing, configuration, debugging, and training. 
& Achieves 94.29\% success with state-of-the-art LLM cores. 
& \halfcirc 
& \halfcirc 
& \emptycirc \\
\midrule
MEDDxAgent \cite{rose2025meddxagent} 
& 2025 
& Differential Diagnosis 
& Enable iterative, interactive differential diagnosis. 
& Integrates a DDxDriver, history simulator, and specialized retrieval/diagnosis agents. 
& Boosts diagnostic accuracy by over 10\% with enhanced explainability. 
& \halfcirc 
& \emptycirc 
& \emptycirc \\
\midrule
PathFinder \cite{ghezloo2025pathfinder} 
& 2025 
& AI-assisted Diagnostics 
& Replicate holistic WSI analysis as done by expert pathologists. 
& Four agents collaboratively generate importance maps and diagnoses. 
& Outperforms state-of-the-art by 8\%, exceeding average pathologist performance by 9\%. 
& \fullcirc 
& \halfcirc 
& \halfcirc \\
\midrule
HamRaz \cite{abbasi2025hamraz} 
& 2025 
& Therapeutic Counseling 
& Provide the first Persian PCT dataset for LLMs with culturally adapted therapy sessions. 
& Combines scripted dialogues and adaptive LLM role‑play. 
& Produces more empathetic, nuanced, and realistic counseling interactions. 
& \emptycirc 
& \emptycirc 
& \emptycirc \\
\midrule
CAMI \cite{yang2025cami} 
& 2025 
& Therapeutic Counseling 
& Automate MI-based counseling with client state inference, topic exploration, and empathetic response generation. 
& STAR framework with three LLM modules for state, topic, and response. 
& Outperforms baselines in MI competency and counseling realism. 
& \halfcirc 
& \halfcirc 
& \emptycirc \\
\midrule
AutoCBT \cite{xu2025autocbt} 
& 2025 
& Therapeutic Counseling 
& Deliver dynamic CBT via multi-agent routing and supervision. 
& Uses single-turn agents and dynamic supervisory routing for tailored interventions. 
& Generates higher-quality CBT responses vs. fixed systems. 
& \halfcirc 
& \halfcirc 
& \emptycirc \\
\midrule
PSYCHE \cite{lee2025psyche} 
& 2025 
& Psychiatric Assessment 
& Benchmark PACAs with simulated patient profiles and multi-turn interactions. 
& Uses detailed psychiatric constructs and board-certified psychiatrist evaluations. 
& Validated for clinical appropriateness and safety. 
& \fullcirc 
& \halfcirc 
& \emptycirc \\
\midrule
PsyDraw \cite{zhang2024psydraw} 
& 2024 
& Mental Health Screening 
& Analyze HTP drawings with multimodal agents for early screening of LBCs. 
& Two-stage feature extraction and report generation; evaluated on 290 submissions; pilot deployment in schools. 
& 71.03\% high consistency with experts; scalable screening tool. 
& \fullcirc 
& \halfcirc 
& \emptycirc \\
\midrule
EvoPatient \cite{du2024llms} 
& 2024 
& Medical Training 
& Simulate patient–doctor dialogues for training via unsupervised LLM agents. 
& Iterative multi-turn consultations refine patient responses and physician questions over 200 case simulations. 
& Improves requirement alignment by $>$10\% and achieves higher human preference. 
& \halfcirc 
& \emptycirc 
& \emptycirc \\
\midrule
Scripted Therapy Agents \cite{wasenmuller2024script} 
& 2024 
& Therapeutic Counseling 
& Constrain LLM responses via expert-written scripts and finite conversational states. 
& Two prompting variants execute 100 simulated sessions following deterministic therapeutic scripts. 
& Demonstrates reliable script adherence and transparent decision paths. 
& \halfcirc 
& \halfcirc 
& \emptycirc \\
\midrule
LIDDiA \cite{averly2025liddia} 
& 2025 
& Drug Discovery 
& Automate end-to-end drug discovery from target selection to lead optimization. 
& Orchestrates LLM-driven reasoning across all pipeline steps; evaluated on 30 targets. 
& Generates valid candidates $>$70\% of cases; identifies novel EGFR inhibitors. 
& \emptycirc 
& \halfcirc 
& \emptycirc \\
\midrule
PatentAgent \cite{wang2024texttt} 
& 2024 
& Pharmaceutical Patents 
& Streamline patent analysis with LLM-driven QA, image-to-molecule, and scaffold ID. 
& PA‑QA, PA‑Img2Mol, PA‑CoreId modules for comprehensive patent insights. 
& Improves image-to-molecule accuracy by up to 8.37\% and scaffold ID by up to 7.62\%. 
& \emptycirc 
& \halfcirc 
& \emptycirc \\
\midrule
DrugAgent \cite{inoue2024drugagent} 
& 2024 
& Drug Repurposing 
& Accelerate drug repurposing via multi-agent ML and knowledge integration. 
& Combines DTI modeling, KG extraction, and literature mining agents. 
& Improves prediction accuracy and reduces discovery time/cost. 
& \halfcirc 
& \halfcirc 
& \emptycirc \\
\midrule
MAP \cite{chen2025map} 
& 2025 
& Inpatient Decision Support 
& Support complex inpatient pathways with specialized triage, diagnosis, and treatment agents. 
& Uses IPDS benchmark; coordinated by a chief agent for end-to-end care planning. 
& +25.10\% diagnostic accuracy vs. HuatuoGPT2‑13B; +10–12\% clinical compliance over clinicians. 
& \fullcirc 
& \halfcirc 
& \emptycirc \\
\midrule
SynthUserEval \cite{yun2025sleepless} 
& 2025 
& Health Coaching 
& Generate synthetic users for evaluating behavior-change agents. 
& Creates structured profiles and simulates interactions with coaching agents. 
& Enables realistic, health-grounded dialogues; validated by expert evaluations. 
& \emptycirc 
& \halfcirc 
& \emptycirc \\
\bottomrule
\end{tabularx}
\end{adjustbox}\\
C: Clinical Validation; W: Workflow Integration; R: Regulatory Compliance;\halfcirc : Partial; \emptycirc : Not Supported; \fullcirc: Supported. 
\end{table*}

\begin{figure*}[t]
    \centering
    \includegraphics[width=0.8\linewidth]{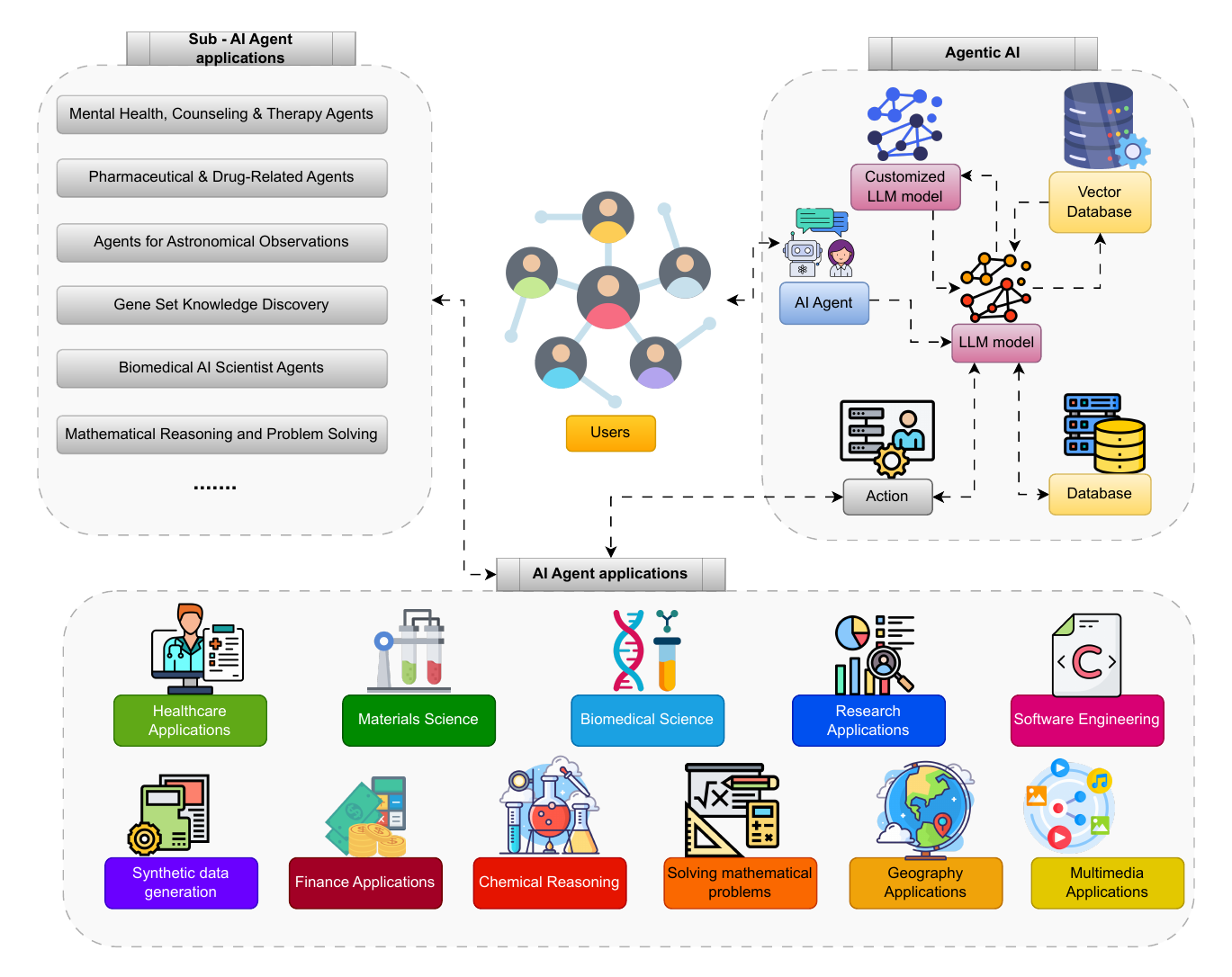}
    \caption{Architecture and Application Domains of AI Agents.}
    \label{fig:LLMapplication}
\end{figure*}

\subsection{AI Agent applications}

AI Agents are autonomous systems that combine large language models (LLMs), data retrieval mechanisms, and decision‑making pipelines to tackle a wide array of tasks across industries. In healthcare, they assist with clinical diagnosis and personalized treatment planning; in finance, they support forecasting and risk analysis; in scientific research, they automate literature review and experimental design; and in software engineering, they generate, analyze, and repair code. Using domain-specific fine-tuning and structured data sources, AI agents can also drive the generation of synthetic data, facilitate chemical reasoning, support mathematical problem-solving, and enable creative multimedia production, thereby expanding the reach of AI-powered automation and insight generation. Fig. \ref{fig:LLMapplication} presents both the architectural backbone and the application landscape of AI Agents.

\subsubsection{Healthcare Applications}

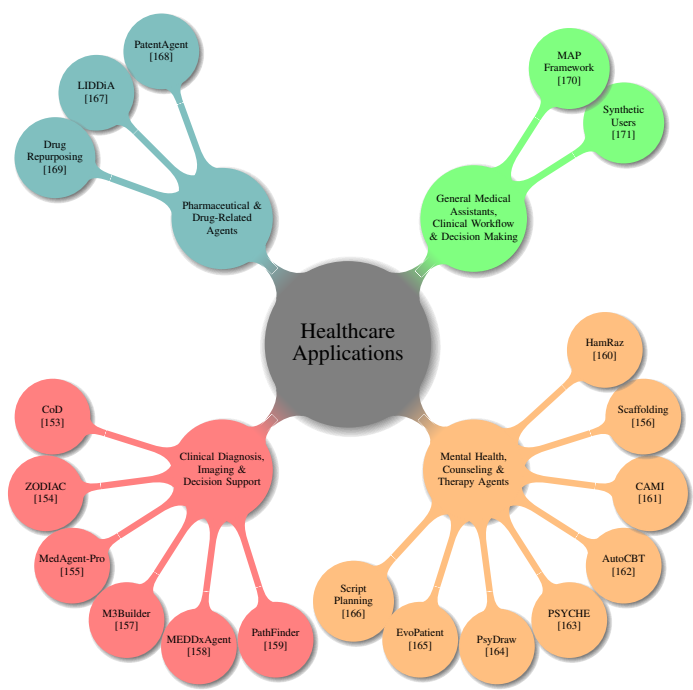
\begin{figure}[htbp]
\centering
\resizebox{0.5\textwidth}{!}{%
\begin{tikzpicture}[
  mindmap,
  every node/.style={concept, circular drop shadow, minimum size=0.1cm},
  grow cyclic, align=flush center, concept color=black!50,
  level 1/.append style={
    sibling angle=90,
    level distance=4cm,
    font=\scriptsize
  },
  level 2/.append style={
    sibling angle=25,
    level distance=4cm,
    font=\scriptsize
  }
]
\node[concept color=black!50] {\Large Healthcare Applications}
  child[concept color=red!50] { node[align=center] {Clinical Diagnosis, Imaging \&\\Decision Support}
    child { node[align=center] {CoD\\\cite{chen2024cod}} }
    child { node[align=center] {ZODIAC\\\cite{zhou2024zodiac}} }
    child { node[align=center] {MedAgent‑Pro\\\cite{wang2025medagent}} }
    child { node[align=center] {M3Builder\\\cite{feng2025m}} }
    child { node[align=center] {MEDDxAgent\\\cite{rose2025meddxagent}} }
    child { node[align=center] {PathFinder\\\cite{ghezloo2025pathfinder}} }
  }
  child[concept color=orange!50] { node[align=center] {Mental Health, Counseling \&\\Therapy Agents}
    child { node[align=center] {Script Planning\\\cite{wasenmuller2024script}} }
    child { node[align=center] {EvoPatient\\\cite{du2024llms}} }
    child { node[align=center] {PsyDraw\\\cite{zhang2024psydraw}} }
    child { node[align=center] {PSYCHE\\\cite{lee2025psyche}} }
    child { node[align=center] {AutoCBT\\\cite{xu2025autocbt}} }
    child { node[align=center] {CAMI\\\cite{yang2025cami}} }
    child { node[align=center] {Scaffolding\\\cite{steenstra2025scaffolding}} }
    child { node[align=center] {HamRaz\\\cite{abbasi2025hamraz}} }
  }
  child[concept color=green!50] { node[align=center] {General Medical Assistants,\\Clinical Workflow \& Decision Making}
    child { node[align=center] {Synthetic Users\\\cite{yun2025sleepless}} }
    child { node[align=center] {MAP Framework\\\cite{chen2025map}} }
  }
  child[concept color=teal!50] { node[align=center] {Pharmaceutical \&\\Drug-Related Agents}
    child { node[align=center] {PatentAgent\\\cite{wang2024texttt}} }
    child { node[align=center] {LIDDiA\\\cite{averly2025liddia}} }
    child { node[align=center] {Drug Repurposing\\\cite{inoue2024drugagent}} }
  }
;
\end{tikzpicture}
}
\caption{Agent LLM Applications for Healthcare}
\label{fig:healthcare_applications}
\end{figure}

The healthcare sector has witnessed significant advancements through the integration of large language model-based agents across a wide range of applications. In this subsection, we present recent developments organized into key categories, as presented in Fig. \ref{fig:healthcare_applications}, including clinical diagnosis and decision support, mental health and therapy agents, general medical assistants for workflow optimization, and pharmaceutical and drug discovery agents. These works demonstrate how AI agents are increasingly supporting medical professionals, enhancing diagnostic accuracy, improving patient care, and accelerating research in diverse healthcare domains. Tab.\label{tab:ai_agent_apps_partII} reviews AI agent applications for Healthcare. 

\paragraph{Clinical Diagnosis, Imaging \& Decision Support}

Chen et al. \cite{chen2024cod} introduce Chain-of-Diagnosis (CoD), a novel approach designed to enhance the interpretability of LLM-based medical diagnostics. By transforming the diagnostic process into a transparent, step-by-step chain that mirrors a physician’s reasoning, CoD provides a clear reasoning pathway alongside a disease confidence distribution, which aids in identifying critical symptoms through entropy reduction. This transparent methodology not only makes the diagnostic process controllable but also boosts rigor in decision-making. Leveraging CoD, the authors developed DiagnosisGPT, an advanced system capable of diagnosing 9,604 diseases. Experimental results demonstrate that DiagnosisGPT outperforms existing large language models (LLMs) on diagnostic benchmarks, achieving both high diagnostic accuracy and enhanced interpretability.

Zhou et al. \cite{zhou2024zodiac} present ZODIAC, an innovative LLM-powered framework that elevates cardiological diagnostics to a level of professionalism comparable to that of expert cardiologists. Designed to address the limitations of general-purpose large language models (LLMs) in clinical settings, ZODIAC leverages a multi-agent collaboration architecture to process patient data across multiple modalities. Each agent is fine-tuned using real-world patient data adjudicated by cardiologists, ensuring the system's diagnostic outputs, such as the extraction of clinically relevant characteristics, arrhythmia detection, and preliminary report generation, are accurate and reliable. Rigorous clinical validation, conducted by independent cardiologists and evaluated across eight metrics addressing clinical effectiveness and security, demonstrates that ZODIAC outperforms industry-leading models, including GPT-4o, Llama-3.1-405B, Gemini-pro, and even specialized medical LLMs like BioGPT. Notably, the successful integration of ZODIAC into electrocardiography (ECG) devices underscores its potential to transform healthcare delivery, exemplifying the emerging trend of embedding LLMs within Software-as-Medical-Device (SaMD) solutions.

Wang et al. \cite{wang2025medagent} introduce MedAgent-Pro, an evidence-based, agentic system designed to enhance multi-modal medical diagnosis by addressing key limitations of current Multi-modal Large Language Models (MLLMs). While MLLMs have demonstrated strong reasoning and task-performing capabilities, they often struggle with detailed visual perception and exhibit reasoning inconsistencies, both of which are critical in clinical settings. MedAgent-Pro employs a hierarchical workflow: at the task level, it leverages knowledge-based reasoning to generate reliable diagnostic plans grounded in retrieved clinical criteria, and at the case level, it utilizes multiple tool agents to process multi-modal inputs and analyze diverse indicators. The final diagnosis is derived from a synthesis of quantitative and qualitative evidence. Comprehensive experiments on both 2D and 3D medical diagnosis tasks demonstrate that MedAgent-Pro not only outperforms existing methods but also offers enhanced reliability and interpretability, marking a significant step forward in AI-assisted clinical diagnostics.

Feng et al. \cite{feng2025m} introduce M3Builder. This novel multi-agent system automates machine learning workflows in the medical imaging domain, a field that has traditionally needed specialized models and tools. M3Builder is structured around four specialized agents that collaboratively manage complex, multi-step ML tasks, including automated data processing, environment configuration, self-contained auto-debugging, and model training, all within a dedicated medical imaging ML workspace. To assess progress in this area, the authors propose M3Bench, a comprehensive benchmark featuring four general tasks across 14 training datasets, covering five anatomies, three imaging modalities, and both 2D and 3D data. Evaluations using seven state-of-the-art large language models as agent cores, such as the Claude series, GPT-4o, and DeepSeek-V3, demonstrate that M3Builder significantly outperforms existing ML agent designs, achieving a remarkable 94.29\% success rate with Claude-3.7-Sonnet.

Rose et al. \cite{rose2025meddxagent} tackles the complexities of differential diagnosis (DDx) by introducing the Modular Explainable DDx Agent (MEDDxAgent) framework, which facilitates interactive, iterative diagnostic reasoning rather than relying on complete patient profiles from the outset. Addressing limitations in previous approaches such as evaluations on single datasets, isolated component optimization, and single-attempt diagnoses MEDDxAgent integrates three modular components: an orchestrator (DDxDriver), a history-taking simulator, and two specialized agents for knowledge retrieval and diagnosis strategy. To ensure robust evaluation, the authors also present a comprehensive DDx benchmark covering respiratory, skin, and rare diseases. Their findings reveal that iterative refinement significantly enhances diagnostic accuracy, with MEDDxAgent achieving over a 10\% improvement across both large and small LLMs while providing critical explainability in its reasoning process.

Ghezloo et al. \cite{ghezloo2025pathfinder} introduce Pathfinder, a novel multimodal, multi-agent framework designed to replicate the holistic diagnostic process of expert pathologists when analyzing whole-slide images (WSIs). Recognizing that WSIs are characterized by their gigapixel scale and complex structure, PathFinder employs four specialized agents a Triage Agent, Navigation Agent, Description Agent, and Diagnosis Agent that collaboratively navigate and interpret the image data. The Triage Agent first determines whether a slide is benign or risky; if deemed risky, the Navigation and Description Agents iteratively focus on and characterize significant regions, generating importance maps and detailed natural language descriptions. Finally, the Diagnosis Agent synthesizes these findings to provide a comprehensive diagnostic classification that is inherently explainable. Experimental results indicate that PathFinder outperforms state-of-the-art methods in skin melanoma diagnosis by 8\% and, notably, surpasses the average performance of pathologists by 9\%, establishing a new benchmark for accurate, efficient, and interpretable AI-assisted diagnostics in pathology.

\paragraph{Mental Health, Counseling \& Therapy Agents}

Wasenm{\"u}ller et al. \cite{wasenmuller2024script} present a script-based dialog policy planning paradigm that enables LLM-powered conversational agents to function as AI therapists by adhering to expert-written therapeutic scripts and transitioning through a finite set of conversational states. By treating the script as a deterministic guide, the approach constrains the model’s responses to align with a defined therapeutic framework, making decision paths transparent for clinical evaluation and risk management. The authors implement two variants of this paradigm, utilizing different prompting strategies, and generate 100 simulated therapy sessions with LLM-driven patient agents. Experimental results demonstrate that both implementations can reliably follow the scripted policy, providing insights into their relative efficiency and effectiveness, and underscoring the feasibility of building inspectable, rule-aligned AI therapy systems.

Du et al. \cite{du2024llms} introduce EvoPatient, a framework for generating simulated patients using large language models to train medical personnel through multi-turn diagnostic dialogues. Existing approaches focus on data retrieval accuracy or prompt tuning, but EvoPatient emphasizes unsupervised simulation to teach patient agents standardized presentation patterns. In this system, a patient agent and doctor agents engage in iterative consultations, with each dialogue cycle serving to both train the agents and gather experience that refines patient responses and physician questions. Extensive experiments across diverse clinical scenarios show that EvoPatient improves requirement alignment by more than 10 percent compared to state‑of‑the‑art methods and achieves higher human preference ratings. After evolving through 200 case simulations over ten hours, the framework achieves an optimal balance between resource efficiency and performance, demonstrating strong generalizability for scalable medical training.

Zhang et al. \cite{zhang2024psydraw} present PsyDraw, a multimodal LLM-driven multi-agent system designed to support mental health professionals in analyzing House‑Tree‑Person (HTP) drawings for early screening of left‑behind children (LBCs) in rural China. Recognizing the acute shortage of clinicians, PsyDraw employs specialized agents for detailed feature extraction and psychological interpretation in two stages: comprehensive analysis of drawing elements and automated generation of professional reports. Evaluated on 290 primary‑school HTP submissions, PsyDraw achieved High Consistency with expert evaluations in 71.03\% of cases and Moderate Consistency in 26.21\%, flagging 31.03\% of children as needing further attention. Deployed in pilot schools, PsyDraw demonstrates strong potential as a scalable, preliminary screening tool that maintains high professional standards and addresses critical mental health gaps in resource‑limited settings.

Lee et al. \cite{lee2025psyche} introduce PSYCHE, a comprehensive framework for benchmarking psychiatric assessment conversational agents (PACAs) built on large language models. Recognizing that psychiatric evaluations rely on nuanced, multi-turn interactions between clinicians and patients, PSYCHE simulates patients using a detailed psychiatric construct that specifies their profiles, histories, and behavioral patterns. This approach enables clinically relevant assessments, ensures ethical safety checks, facilitates cost-efficient deployment, and provides quantitative evaluation metrics. The framework was validated in a study involving ten board-certified psychiatrists who reviewed and rated the simulated interactions, demonstrating PSYCHE’s ability to evaluate PACAs’ clinical appropriateness and safety rigorously. 

Xu et al. \cite{xu2025autocbt} addresses the limitations of existing LLM-based Cognitive Behavioral Therapy (CBT) systems, namely their rigid agent structures and tendency toward redundant, unhelpful suggestions, by proposing AutoCBT, a dynamic multi-agent framework for automated psychological counseling. Initially, the authors develop a general single-turn consultation agent using Quora-like and YiXinLi models, evaluated on a bilingual dataset to benchmark response quality in single-round interactions. Building on these insights, they introduce dynamic routing and supervisory mechanisms modeled after real-world counseling practices, enabling agents to self-optimize and tailor interventions more effectively. Experimental results demonstrate that AutoCBT generates higher-quality CBT-oriented responses compared to fixed-structure systems, highlighting its potential to deliver scalable, empathetic, and contextually appropriate psychological support for users who might otherwise avoid in-person therapy.

Yang et al. \cite{yang2025cami} present CAMI, an automated conversational counselor agent grounded in Motivational Interviewing (MI), a client-centered approach designed to resolve ambivalence and promote behavior change. CAMI’s novel STAR framework integrates three LLM-powered modules client State inference, motivation Topic exploration, and response gEneration to evoke “change talk” in line with MI principles. By accurately inferring a client’s emotional and motivational state, exploring relevant topics, and generating empathetic, directive responses, CAMI facilitates more effective counseling across diverse populations. The authors evaluate CAMI using both automated metrics and manual assessments with simulated clients, measuring MI skill competency, state inference accuracy, topic exploration proficiency, and overall counseling success. Results demonstrate that CAMI outperforms existing methods and exhibits counselor-like realism, while ablation studies highlight the essential contributions of the state inference and topic exploration modules to its superior performance.

Steenstra et al. \cite{steenstra2025scaffolding} address the challenges in therapeutic counseling training, confined mainly to an innovative LLM-powered system that provides continuous, detailed feedback during simulated patient interactions. Focusing on motivational interviewing a counseling approach emphasizing empathy and collaborative behavior change the framework features a simulated patient and visualizations of turn-by-turn performance to guide counselors through role-play scenarios. The system was evaluated with both professional and student counselors, who reported high usability and satisfaction, indicating that frequent and granular feedback can significantly enhance the learning process compared to traditional, intermittent methods.

Abbasi et al. \cite{abbasi2025hamraz} introduce HamRaz, the first Persian-language dataset tailored for Person-Centered Therapy (PCT) with large language models (LLMs), addressing a critical gap in culturally and linguistically appropriate mental health resources. Recognizing that existing counseling datasets are largely confined to Western and East Asian contexts, the authors design HamRaz by blending scripted therapeutic dialogues with adaptive LLM-driven role-playing to foster coherent, dynamic therapy sessions in Persian. To rigorously assess performance, they propose HamRazEval, a dual evaluation framework combining general dialogue quality metrics with the Barrett–Lennard Relationship Inventory (BLRI) to measure therapeutic rapport and effectiveness. Experimental comparisons demonstrate that LLMs trained on HamRaz generate more empathetic, contextually nuanced, and realistic counseling interactions than conventional Script Mode or Two-Agent Mode approaches.

\paragraph{General Medical Assistants, Clinical Workflow \& Decision Making}

Yun et al. \cite{yun2025sleepless} introduce an end‑to‑end framework for generating synthetic users to evaluate interactive agents aimed at promoting positive behavior change, focusing on sleep and diabetes management. The framework first generates structured data based on real‑world health and lifestyle factors, demographics, and behavioral attributes. Next, it creates complete user profiles conditioned on this structured data. Interactions between synthetic users and health coaching agents are simulated using generative agent models such as Concordia or by directly prompting a language model. Case studies with sleep and diabetes coaching agents demonstrate that the synthetic users enable realistic dialogue by accurately reflecting users’ needs and challenges. Blinded evaluations by human experts confirm that these health‑grounded synthetic users portray real human users more faithfully than generic synthetic users. This approach provides a scalable and realistic testing ground for developing and refining conversational agents in health and lifestyle coaching.

Chen et al. \cite{chen2025map} address the complexity of clinical decision-making in inpatient pathways by introducing both a new benchmark and a multi-agent AI framework. The authors construct the Inpatient Pathway Decision Support (IPDS) benchmark from the MIMIC‑IV database, comprising 51,274 cases across nine triage departments, 17 disease categories, and 16 standardized treatment options to capture the multifaceted nature of inpatient care. Building on this resource, they propose the Multi-Agent Inpatient Pathways (MAP) framework, which employs a triage agent for patient admission, a diagnosis agent for department-level decision-making, and a treatment agent for care planning, all coordinated by a chief agent that oversees the entire pathway. In extensive experiments, MAP achieves a 25.10\% improvement in diagnostic accuracy over the state-of-the-art LLM HuatuoGPT2‑13B and surpasses three board-certified clinicians in clinical compliance by 10–12\%. These results demonstrate the potential of multi‑agent systems to support complex inpatient workflows and lay the groundwork for future AI‑driven decision support in hospital settings.

\paragraph{Pharmaceutical \& Drug-Related Agents}

Wang et al. \cite{wang2024texttt} introduce PatentAgent, the first end-to-end intelligent agent designed to streamline pharmaceutical patent analysis by leveraging large language models. PatentAgent integrates three core modules: PA‑QA for patent question answering, PA‑Img2Mol for converting chemical structure images into molecular representations, and PA‑CoreId for identifying core chemical scaffolds. PA‑Img2Mol achieves accuracy gains of 2.46 to 8.37 percent across CLEF, JPO, UOB, and USPTO patent image benchmarks, while PA‑CoreId delivers improvements of 7.15 to 7.62 percent on the PatentNetML scaffold identification task. By combining these modules within a unified framework, PatentAgent addresses the full spectrum of patent analysis needs, from extracting detailed experimental insights to pinpointing key molecular structures, and offers a powerful tool to accelerate research and innovation in drug discovery.

Averly et al. \cite{averly2025liddia} introduce LIDDiA, an autonomous in silico agent designed to navigate the entire drug discovery pipeline by leveraging the reasoning capabilities of large language models. Unlike prior AI tools that address individual steps such as molecule generation or property prediction, LIDDiA orchestrates the end-to-end process from target selection through lead optimization. The authors evaluate LIDDiA on 30 clinically relevant targets and show that it generates candidate molecules satisfying key pharmaceutical criteria in over 70 percent of cases. Furthermore, LIDDiA demonstrates an intelligent balance between exploring novel chemical space and exploiting known scaffolds and successfully identifies promising new inhibitors for the epidermal growth factor receptor (EGFR), a major oncology target.

Inoue et al. \cite{inoue2024drugagent} present a multi‑agent framework designed to accelerate drug repurposing by combining machine learning and knowledge integration. The system includes three specialized agents: an AI Agent that trains robust drug–target interaction (DTI) models, a Knowledge Graph Agent that extracts DTIs from databases such as DGIdb, DrugBank, CTD and STITCH, and a Search Agent that mines biomedical literature to validate computational predictions. By integrating outputs from these agents, the framework leverages diverse data sources to identify promising candidates for repurposing. Preliminary evaluations indicate that this approach not only enhances the accuracy of drug–disease interaction predictions compared to existing methods but also reduces the time and cost associated with traditional drug discovery. The interpretable results and scalable architecture demonstrate the potential of multi‑agent systems to drive innovation and efficiency in biomedical research.

\subsubsection{Materials Science}

Materials science has recently benefited from the integration of LLM-based agents, which are helping to automate complex scientific workflows and enhance research efficiency. In this subsection, we highlight two notable developments, including the application of AI agents in astronomical observations to streamline data collection and analysis, and the creation of specialized agent systems tailored to address the unique challenges of materials science research. 

\paragraph{LLM-Based Agents for Astronomical Observations}
The StarWhisper Telescope System \cite{wang2024starwhisper} leverages LLM-based agents to streamline the complex workflow of astronomical observations within the Nearby Galaxy Supernovae Survey (NGSS) project. This innovative system automates critical tasks including generating customized observation lists, initiating telescope observations, real-time image analysis, and formulating follow-up proposals to reduce the operational burden on astronomers and lower training costs. By integrating these agents into the observation process, the system can efficiently verify and dispatch observation lists, analyze transient phenomena in near real-time, and seamlessly communicate results to observatory teams for subsequent scheduling.

\paragraph{Materials Science Research}
HoneyComb \cite{zhang2024honeycomb} is introduced as the first LLM-based agent system tailored explicitly for materials science, addressing the unique challenges posed by complex computational tasks and outdated implicit knowledge that often lead to inaccuracies and hallucinations in general-purpose LLMs. The system leverages a novel, high-quality materials science knowledge base (MatSciKB) curated from reliable literature and a sophisticated tool hub (ToolHub) that employs an Inductive Tool Construction method to generate, decompose, and refine specialized API tools. Additionally, the retriever module adaptively selects the most relevant knowledge sources and tools for each task, ensuring high accuracy and contextual relevance.

\subsubsection{Biomedical Science}

The biomedical field has seen important progress through the development of LLM-based agents designed to support knowledge discovery, enhance reasoning capabilities, and evaluate scientific literature. In this subsection, we review recent contributions that focus on gene set analysis, iterative learning for improved reasoning, and the evaluation of AI scientist agents through specialized biomedical benchmarks.

\paragraph{Gene Set Knowledge Discovery}
Gene set knowledge discovery is crucial for advancing human functional genomics, yet traditional LLM approaches often suffer from issues like hallucinations. To address this, Wang et al. \cite{wang2024geneagent} introduce GeneAgent a pioneering language agent with self-verification capabilities that autonomously interacts with biological databases and leverages specialized domain knowledge to enhance accuracy. Benchmarking on 1,106 gene sets from diverse sources, GeneAgent consistently outperforms standard GPT‑4, and a detailed manual review confirms that its self-verification module effectively minimizes hallucinations and produces more reliable analytical narratives. Moreover, when applied to seven novel gene sets derived from mouse B2905 melanoma cell lines, expert evaluations reveal that GeneAgent offers novel insights into gene functions, significantly expediting the process of knowledge discovery in functional genomics.

\paragraph{Reasoning with Recursive Learning}
Buehler et al. \cite{buehler2024preflexor} proposed a framework, named PRefLexOR, that fuses preference optimization with reinforcement learning concepts to enable language models to self-improve through iterative, multi-step reasoning. The approach employs a recursive learning strategy in which the model repeatedly revisits and refines intermediate reasoning steps before producing a final output, both during training and inference. Initially, the model aligns its reasoning with accurate decision paths by optimizing the log odds between preferred and non-preferred responses while constructing a dynamic knowledge graph through question generation and retrieval augmentation. In a subsequent stage, rejection sampling is employed to refine the reasoning quality by generating in-situ training data and masking intermediate steps, all within a thinking token framework that fosters iterative feedback loops.

\paragraph{Biomedical AI Scientist Agents}
Lin et al. \cite{lin2024biokgbench} introduce BioKGBench, a novel benchmark designed to evaluate biomedical AI scientist agents from the perspective of literature understanding. Unlike traditional evaluation methods that rely solely on direct QA or biomedical experiments, BioKGBench decomposes the critical ability of “understanding literature” into two atomic tasks: one that verifies scientific claims in unstructured text from research papers and another that involves interacting with structured knowledge-graph question-answering (KGQA) for literature grounding. Building on these components, the authors propose a new agent task called KGCheck, which uses domain-based retrieval-augmented generation to identify factual errors in large-scale knowledge graph databases. With a dataset of over 2,000 examples for the atomic tasks and 225 high-quality annotated samples for the agent task, the study reveals that state-of-the-art agents both in everyday and biomedical settings perform poorly or suboptimally on this benchmark. 

\begin{table*}[htbp]
\centering
\scriptsize
\caption{Overview of AI Agent Applications for Research}
\label{tab:ai_agent_apps_research}
\begin{adjustbox}{max width=\textwidth}
\begin{tabularx}{\textwidth}{@{}%
  >{\raggedright\arraybackslash}p{1.8cm}   
  >{\centering\arraybackslash}p{0.6cm}      
  >{\centering\arraybackslash}p{1.2cm}      
  >{\raggedright\arraybackslash}X           
  >{\raggedright\arraybackslash}X           
  >{\raggedright\arraybackslash}X           
  >{\centering\arraybackslash}p{1.2cm}      
  >{\centering\arraybackslash}p{1.5cm}      
  >{\centering\arraybackslash}p{1.5cm}      
@{}}
\toprule
\textbf{Agent / Tool} & \textbf{Year} & \textbf{Use Case} & \textbf{Primary Aim} & \textbf{Methodology \& Workflow} & \textbf{Key Findings \& Metrics} & \textbf{Eval.\ Framework} & \textbf{Collab.\ Platform} & \textbf{Open Sci.} \\
\midrule
AgentRxiv \cite{schmidgall2025agentrxiv} 
  & 2025 
  & Collaborative Research 
  & Share and build upon preprints across autonomous LLM labs. 
  & Upload/retrieve via shared preprint server with iterative updates. 
  & +11.4\% on MATH‑500; +3.3\% cross‑domain; +13.7\% multi‑lab. 
  & MATH‑500 benchmark 
  & AgentRxiv server 
  & Preprint sharing \\
\midrule
SurveyX \cite{liang2025surveyx} 
  & 2025 
  & Survey Generation 
  & Automate systematic literature surveys with high quality. 
  & Preparation (retrieval + AttributeTree) + Generation (repolishing). 
  & +0.259 content quality; +1.76 citation precision vs.\ baselines. 
  & Content \& citation scoring 
  & Bibliographic APIs 
  & Structured citations \\
\midrule
CoI Agent \cite{li2024chain} 
  & 2024 
  & Research Ideation 
  & Structure literature into progressive idea chains. 
  & Sequential Chain-of-Ideas + Idea Arena evaluation protocol. 
  & Expert‑comparable idea quality at \$0.50 per idea. 
  & Idea Arena 
  & CoI framework 
  & Cost‑efficient ideation \\
\midrule
Data Interpreter \cite{hong2024data} 
  & 2024 
  & Data Science Workflows 
  & Manage end‑to‑end, dynamic DS pipelines robustly. 
  & Hierarchical Graph Modeling + Programmable Node Generation. 
  & +25\% on InfiAgent‑DABench (75.9→94.9\%); ML \& MATH gains. 
  & InfiAgent DABench 
  & Pipeline APIs 
  & Reproducible workflows \\
\midrule
AI Co‑Scientist \cite{gottweis2025towards} 
  & 2025 
  & Scientific Discovery 
  & Generate and refine research hypotheses autonomously. 
  & Seven specialized agents with Elo tournaments and meta‑review. 
  & +300 Elo hypothesis quality; +27\% novelty scores. 
  & Elo \& novelty scoring 
  & Multi‑agent pipeline 
  & Hypothesis publication \\
\bottomrule
\end{tabularx}
\end{adjustbox}\\
\textbf{Eval.\ Framework}: Evaluation Framework; \textbf{Collab.\ Platform}: Collaboration Platform; \textbf{Open Sci.}: Open Science Support.
\end{table*}

\subsubsection{Research Applications}
LLM-based agents are increasingly being developed to support and automate various aspects of the scientific research process. This subsection presents a selection of recent applications, including collaborative research environments, automated survey generation, structured literature analysis for ideation, workflow management in data science, and AI-driven hypothesis generation. 

\paragraph{Collaborative Research Among LLM Agents}
Schmidgall and Moor \cite{schmidgall2025agentrxiv} introduces AgentRxiv, a framework designed to enable collaborative research among autonomous LLM agent laboratories by leveraging a shared preprint server. Recognizing that scientific discovery is inherently incremental and collaborative, AgentRxiv allows agents to upload and retrieve research reports, thereby sharing insights and building upon previous work in an iterative manner. The study demonstrates that agents with access to prior research achieve a significant performance boost an 11.4\% relative improvement on the MATH-500 dataset compared to those operating in isolation. Furthermore, the best-performing collaborative strategy generalizes to other domains with an average improvement of 3.3\%, and when multiple agent laboratories share their findings, overall accuracy increases by 13.7\% relative to the baseline. These findings highlight the potential of autonomous agents to collaborate with humans, paving the way for more efficient and accelerated scientific discovery.

\paragraph{Automated Survey Generation}
Liang et al. \cite{liang2025surveyx} developed the SurveyX platform, which leverages the exceptional comprehension and knowledge capabilities of LLMs to overcome critical limitations in automated survey generation, including finite context windows, superficial content discussions, and the lack of systematic evaluation frameworks. Inspired by human writing processes, SurveyX decomposes the survey composition process into two distinct phases: Preparation and Generation. During the preparation phase, the system incorporates online reference retrieval and applies a novel preprocessing method, AttributeTree, to effectively structure the survey’s content. In the subsequent Generation phase, a repolishing process refines the output to enhance the depth and accuracy of the study generated, particularly improving content quality and citation precision. Experimental evaluations reveal that SurveyX achieves a content quality improvement of 0.259 and a citation quality enhancement of 1.76 over existing systems, bringing its performance close to that of human experts across multiple evaluation dimensions.

\paragraph{Structuring Literature for Research Ideation}
Li et al. \cite{li2024chain} introduce the Chain-of-Ideas (CoI) agent, a novel LLM-based framework for automating research ideation by structuring relevant literature into a chain that mirrors the progressive development within a research domain. The CoI agent addresses the challenge posed by the exponential growth of scientific literature, which overwhelms traditional idea-generation methods that rely on simple prompts or expose models to raw, unfiltered text. By organizing information in a sequential chain, the CoI agent enables LLMs to capture current advancements more effectively, enhancing their ability to generate innovative research ideas. Complementing this framework is the Idea Arena, an evaluation protocol that assesses the quality of generated ideas from multiple perspectives, aligning closely with the preferences of human researchers. Experimental results indicate that the CoI agent outperforms existing methods and achieves quality comparable to human experts, all while maintaining a low cost approximately \$0.50 per candidate idea and corresponding experimental design.

\paragraph{Managing Data Science Workflows}
Hong et al. \cite{hong2024data} propose Data Interpreter, an LLM-based agent that tackles end-to-end data science workflows by addressing challenges in solving long-term, interconnected tasks and adapting to dynamic data environments. Unlike previous methods that focus on individual tasks, Data Interpreter leverages two key modules: Hierarchical Graph Modeling, which decomposes complex problems into manageable subproblems through dynamic node generation and graph optimization, and Programmable Node Generation, which iteratively refines and verifies each subproblem to boost the robustness of code generation. Extensive experiments demonstrate significant performance gains achieving up to a 25\% boost on InfiAgent-DABench (increasing accuracy from 75.9\% to 94.9\%), as well as improvements on machine learning, open-ended tasks, and the MATH dataset highlighting its superior capability in managing evolving task dependencies and real-time data adjustments.

\paragraph{Automating Scientific Discovery}
Google \cite{gottweis2025towards} introduced the AI co-scientist, a multi-agent system built on Google DeepMind Gemini 2.0, designed to automate scientific discovery by generating and refining novel research hypotheses. The framework comprises seven specialized agents Supervisor, Generation, Reflection, Ranking, Evolution, Proximity, and Meta-review that collaboratively manage tasks ranging from parsing research goals to conducting simulated debates and organizing hypotheses. For example, the system employs a Ranking Agent that uses pairwise Elo tournaments, boosting hypothesis quality by over 300 Elo points. At the same time, the Meta-review Agent’s feedback has been shown to increase hypothesis novelty scores by 27\%. In practical applications, such as drug repurposing for acute myeloid leukemia and novel target discovery for liver fibrosis, the framework demonstrates significant performance improvements, paving the way for AI systems that can generate and iteratively refine scientific hypotheses with expert-level precision.

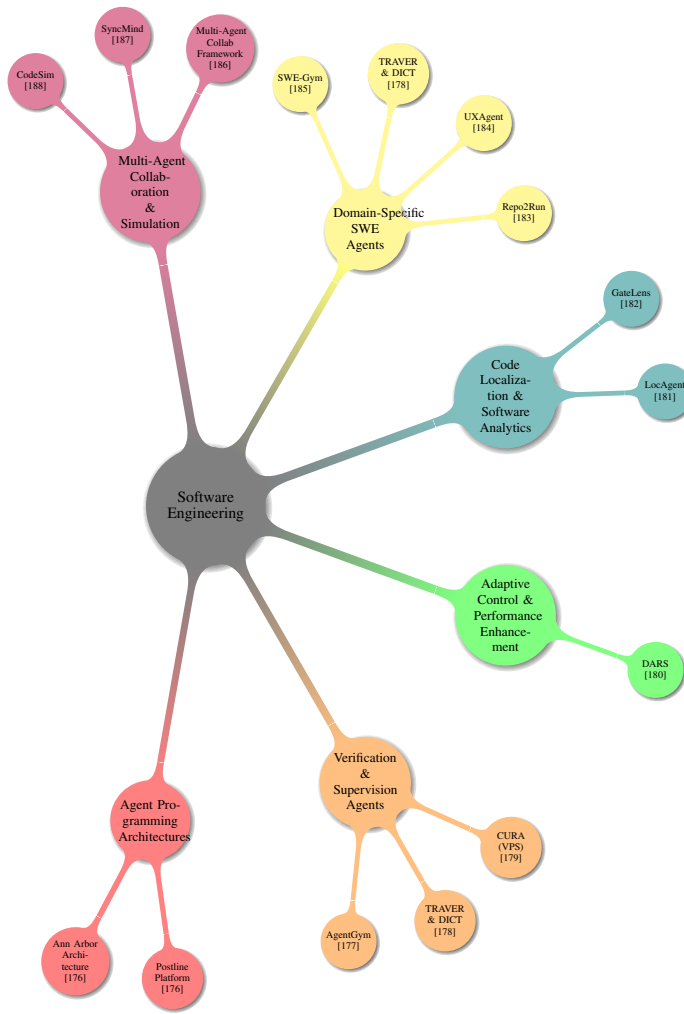
\begin{figure}[htbp]
\centering
\resizebox{0.5\textwidth}{!}{%
\begin{tikzpicture}[
  mindmap,
  every node/.style={concept, circular drop shadow, minimum size=0.1cm},
  grow cyclic, align=flush center, concept color=black!50,
  level 1/.append style={
    sibling angle=40,
    level distance=10cm,
    font=\large
  },
  level 2/.append style={
    sibling angle=36,
    level distance=5cm,
    font=\small
  }
]
\node[concept color=black!50] {\Large Software Engineering}
  child[concept color=red!50] { node[align=center] {Agent Programming\\Architectures}
    child { node[align=center] {Ann Arbor Architecture\\\cite{dong2025ann}} }
    child { node[align=center] {Postline Platform\\\cite{dong2025ann}} }
  }
  child[concept color=orange!50] { node[align=center] {Verification \&\\Supervision Agents}
    child { node[align=center] {AgentGym\\\cite{jain2025r2e}} }
    child { node[align=center] {TRAVER \& DICT\\\cite{wang2025training}} }
    child { node[align=center] {CURA (VPS)\\\cite{chen2025verbal}} }
  }
  child[concept color=green!50] { node[align=center] {Adaptive Control \&\\Performance Enhancement}
    child { node[align=center] {DARS\\\cite{aggarwal2025dars}} }
  }
  child[concept color=teal!50] { node[align=center] {Code Localization \&\\Software Analytics}
    child { node[align=center] {LocAgent\\\cite{chen2025locagent}} }
    child { node[align=center] {GateLens\\\cite{gholamzadeh2025gatelens}} }
  }
  child[concept color=yellow!50] { node[align=center] {Domain‑Specific SWE\\Agents}
    child { node[align=center] {Repo2Run\\\cite{hu2025llm}} }
    child { node[align=center] {UXAgent\\\cite{lu2025uxagent}} }
    child { node[align=center] {TRAVER \& DICT\\\cite{wang2025training}} }
    child { node[align=center] {SWE‑Gym\\\cite{pan2024training}} }
  }
  child[concept color=purple!50] { node[align=center] {Multi‑Agent Collaboration\\\& Simulation}
    child { node[align=center] {Multi‑Agent Collab Framework\\\cite{yang2025multi}} }
    child { node[align=center] {SyncMind\\\cite{guo2025syncmind}} }
    child { node[align=center] {CodeSim\\\cite{islam2025codesim}} }
  }
;
\end{tikzpicture}%
}
\caption{Agent LLM Applications in Software Engineering}
\label{fig:software_engineering_applications}
\end{figure}

\begin{table*}[htbp]
\centering
\scriptsize
\caption{Overview of AI Agent Applications for Software Engineering}
\label{tab:ai_agent_apps_sw}
\begin{adjustbox}{max width=\textwidth}
\begin{tabularx}{\textwidth}{@{}%
  >{\raggedright\arraybackslash}p{1.2cm}   
  >{\centering\arraybackslash}p{0.3cm}      
  >{\centering\arraybackslash}p{1cm}        
  >{\raggedright\arraybackslash}X           
  >{\raggedright\arraybackslash}X           
  >{\raggedright\arraybackslash}X           
  >{\centering\arraybackslash}p{0.8cm}      
  >{\centering\arraybackslash}p{0.8cm}      
  >{\centering\arraybackslash}p{0.8cm}      
@{}}
\toprule
\textbf{Agent / Tool} & \textbf{Year} & \textbf{SE Domain} & \textbf{Primary Objective} & \textbf{Architecture \& Workflow} & \textbf{Key Outcomes \& Metrics} & \textbf{Bench.} & \textbf{Intgr.} & \textbf{Std.} \\
\midrule
Ann Arbor Architecture \cite{dong2025ann} 
  & 2025 
  & Agent Programming Arch. 
  & Treat LLMs as automata, enabling programming via formal and natural languages. 
  & Introduces the Ann Arbor conceptual framework and Postline platform. 
  & Early experiments show improved in‑context learning. 
  & \halfcirc 
  & \emptycirc 
  & \emptycirc \\

\midrule
AgentGym \cite{jain2025r2e} 
  & 2025 
  & Verification \& Supervision 
  & Scalable training of SWE‑agents via SYNGEN data curation and Hybrid Test‑time Scaling. 
  & Leverages SYNGEN synthetic data and Hybrid Test‑time Scaling on SWE‑Gym; trained on SWE‑Bench Verified. 
  & Achieves 51\% pass rate on SWE‑Bench Verified. 
  & \fullcirc 
  & \halfcirc 
  & \emptycirc \\

\midrule
TRAVER\&DICT \cite{wang2025training} 
  & 2025 
  & Intelligent Tutoring 
  & Trace-and-Verify workflow for stepwise coding guidance; DICT evaluation protocol. 
  & Combines knowledge tracing with turn-by-turn verification; evaluated via DICT protocol. 
  & Significant improvements in coding‑tutoring success rates. 
  & \fullcirc 
  & \halfcirc 
  & \emptycirc \\

\midrule
CURA \cite{chen2025verbal} 
  & 2025 
  & Code Reasoning 
  & Verbal Process Supervision for code understanding and reasoning. 
  & Integrates VPS modules with LLM to guide reasoning over code. 
  & +3.65\% on BigCodeBench with o3‑mini. 
  & \fullcirc 
  & \halfcirc 
  & \emptycirc \\

\midrule
DARS \cite{aggarwal2025dars} 
  & 2025 
  & Performance Enhancement 
  & Dynamic Action Re‑Sampling to branch inference at decision points. 
  & Branches on execution feedback to explore alternative actions. 
  & 55\% pass@k and 47\% pass@1 on SWE‑Bench Lite (Claude 3.5 Sonnet V2). 
  & \fullcirc 
  & \halfcirc 
  & \emptycirc \\

\midrule
LocAgent \cite{chen2025locagent} 
  & 2025 
  & Code Localization 
  & Graph‑based code representation for multi‑hop localization. 
  & Parses code into heterogeneous graphs for reasoning over dependencies. 
  & 92.7\% file‑level accuracy; +12\% GitHub issue resolution. 
  & \fullcirc 
  & \halfcirc 
  & \emptycirc \\

\midrule
GateLens \cite{gholamzadeh2025gatelens} 
  & 2025 
  & Release Validation 
  & NL→Relational‑Algebra conversion and Python code generation for test‑data analysis. 
  & Automates query translation and optimized code for data processing. 
  & 80\% reduction in analysis time (automotive software). 
  & \fullcirc 
  & \halfcirc 
  & \emptycirc \\

\midrule
Repo2Run \cite{hu2025llm} 
  & 2025 
  & Env. Configuration 
  & Atomic Docker setup synthesis with dual‑environment rollback. 
  & Synthesizes and tests Dockerfiles; isolates failures via dual environments. 
  & 86.0\% success on 420 Python repos; +63.9\% vs. baselines. 
  & \fullcirc 
  & \halfcirc 
  & \emptycirc \\

\midrule
UXAgent \cite{lu2025uxagent} 
  & 2025 
  & Usability Testing 
  & LLM‑agent with browser connector to simulate thousands of users. 
  & Generates qualitative insights, action logs, and recordings before user studies. 
  & Accelerates UX iteration and reduces upfront user recruitment. 
  & \halfcirc 
  & \halfcirc 
  & \emptycirc \\

\midrule
SWE‑Gym \cite{pan2024training} 
  & 2024 
  & Training Environment 
  & Realistic Python tasks and unit tests for SWE‑agent training. 
  & Provides executable environments with tests and natural language descriptions. 
  & +19\% resolve rate; 32.0\% on SWE‑Bench Verified; 26.0\% on Lite. 
  & \fullcirc 
  & \halfcirc 
  & \emptycirc \\

\midrule
Qwen2.5‑xCoder \cite{yang2025multi} 
  & 2025 
  & Multi‑Agent Collaboration 
  & Multilingual instruction tuning via language‑specific agents with memory. 
  & Agents collaborate to generate and refine multilingual instructions. 
  & Outperforms on multilingual programming benchmarks. 
  & \fullcirc 
  & \halfcirc 
  & \emptycirc \\

\midrule
SyncMind \cite{guo2025syncmind} 
  & 2025 
  & Collaboration Simulation 
  & Defines and benchmarks out‑of‑sync scenarios to improve agent coordination. 
  & Introduces SyncBench with 24 k real‑world instances. 
  & Exposes performance gaps and guides improvements. 
  & \fullcirc 
  & \halfcirc 
  & \emptycirc \\

\midrule
CodeSim \cite{islam2025codesim} 
  & 2025 
  & Code Generation 
  & Plan verification and I/O simulation for multi‑agent synthesis \& debugging. 
  & Incorporates plan verification and internal debugging via input/output simulation. 
  & SOTA on HumanEval, MBPP, APPS, CodeContests. 
  & \fullcirc 
  & \halfcirc 
  & \emptycirc \\

\bottomrule
\end{tabularx}
\end{adjustbox}\\
Bench.: Benchmarking; Intgr.: Integration \& Deployment; Std.: Standards Compliance; \halfcirc: Partial; \emptycirc: Not Supported; \fullcirc: Supported.
\end{table*}

\subsubsection{Software Engineering}
Software engineering has become a significant area of application for LLM-based agents, with innovations spanning architecture design and verification systems, adaptive control, software analytics, and multi-agent collaboration. This subsection presents recent developments across a wide range of tasks, including agent programming frameworks, tutoring systems, automated environment configuration, usability testing, and multilingual code generation. Fig. \ref{fig:software_engineering_applications} presents a classification of Agent LLM Applications for Software Engineering.

\paragraph{Agent Programming Architectures}

Dong et al. \cite{dong2025ann} explore prompt engineering for large language models (LLMs) from the perspective of automata theory, arguing that LLMs can be viewed as automata. They assert that just as automata must be programmed using the languages they accept, LLMs should similarly be programmed within the scope of both natural and formal languages. This insight challenges traditional software engineering practices, which often distinguish between programming and natural languages. The paper introduces the Ann Arbor Architecture, a conceptual framework designed for agent-oriented programming of language models, which serves as a higher-level abstraction to enhance in-context learning beyond basic token generation. The authors also present Postline, their agent platform, and discuss early results from experiments conducted to train agents within this framework.

\paragraph{Verification \& Supervision Agents}

The papers by Jain et al. \cite{jain2025r2e} , Wang et al.\cite{wang2025training}, and Chen et al. \cite{chen2025verbal} contribute to advancing the use of large language models (LLMs) for real-world software engineering (SWE) tasks, intelligent tutoring, and code generation. Jain et al. \cite{jain2025r2e} introduce AgentGym, a comprehensive environment for training SWE-agents, addressing challenges in scalable curation of executable environments and test-time compute scaling. Their approach leverages SYNGEN, a synthetic data curation method, and Hybrid Test-time Scaling to improve performance on the SWE-Bench Verified benchmark, achieving a state-of-the-art pass rate of 51\%. Wang et al. \cite{wang2025training}  propose a novel coding tutoring framework, Trace-and-Verify (TRAVER), combining knowledge tracing and turn-by-turn verification to enhance tutor agents' guidance toward task completion. Their work introduces DICT, a holistic evaluation protocol for tutoring agents, demonstrating significant improvements in coding tutoring success rates. Finally, Chen et al. present CURA, a code understanding and reasoning system augmented with verbal process supervision (VPS). CURA achieves a 3.65\% improvement on benchmarks like BigCodeBench and demonstrates enhanced performance when paired with the o3-mini model. These works collectively push the boundaries of LLM applications in complex software engineering tasks, intelligent tutoring, and reasoning-driven code generation.

\paragraph{Adaptive Control \& Performance Enhancement}

Aggarwal et al. \cite{aggarwal2025dars} introduce Dynamic Action Re-Sampling (DARS), a novel approach for scaling compute during inference in coding agents, aimed at improving their decision-making capabilities. While existing methods often rely on linear trajectories or random sampling, DARS enhances agent performance by branching out at key decision points and selecting alternative actions based on the history of previous attempts and execution feedback. This enables coding agents to recover more effectively from sub-optimal decisions, leading to faster and more efficient problem-solving. The authors evaluate DARS on the SWE-Bench Lite benchmark, achieving an impressive pass@k score of 55\% with Claude 3.5 Sonnet V2 and a pass@1 rate of 47\%, surpassing current state-of-the-art open-source frameworks. This approach provides a significant advancement in optimizing coding agent performance, reducing the need for extensive manual intervention and improving overall efficiency.

\paragraph{Code Localization \& Software Analytics}

The works by Chen et al. \cite{chen2025locagent} and Gholamzadeh et al. \cite{gholamzadeh2025gatelens} contribute significant advancements in the application of Large Language Models (LLMs) to improve software engineering tasks, such as code localization and release validation. Chen et al. \cite{chen2025locagent} introduce LocAgent, a framework for code localization that utilizes graph-based representations of codebases. By parsing code into directed heterogeneous graphs, LocAgent captures the relationships between various code structures and their dependencies, enabling more efficient and accurate localization through multi-hop reasoning. Their approach, when applied to real-world benchmarks, demonstrates substantial improvements in localization accuracy, achieving up to 92.7\% on file-level localization and enhancing GitHub issue resolution success rates by 12\%. In comparison to state-of-the-art models, LocAgent provides similar performance at a significantly lower cost. On the other hand, Gholamzadeh et al. \cite{gholamzadeh2025gatelens} present GateLens, an LLM-based tool designed to improve release validation in safety-critical systems like automotive software. GateLens automates the analysis of test data by converting natural language queries into Relational Algebra expressions and generating optimized Python code, which significantly accelerates data processing. In industrial evaluations, GateLens reduced analysis time by over 80\%, demonstrating strong robustness and generalization across different query types. This tool improves decision-making in safety-critical environments by automating test result analysis, thereby enhancing the scalability and reliability of software systems in automotive applications.

\paragraph{Domain‑Specific SWE Agents}
Hu et al. \cite{hu2025llm} introduce Repo2Run, a novel LLM-based agent aimed at automating the environment configuration process in software development. Traditional methods for setting up environments often involve manual work or rely on fragile scripts, which can lead to inefficiencies and errors. Repo2Run addresses these challenges by fully automating the configuration of Docker containers for Python repositories. The key innovations of Repo2Run are its atomic configuration synthesis and a dual-environment architecture, which isolates internal and external environments to prevent contamination from failed commands. A rollback mechanism ensures that only fully executed configurations are applied, and the agent generates executable Dockerfiles from successful configurations. Evaluated on a benchmark of 420 Python repositories with unit tests, Repo2Run achieved an impressive success rate of 86.0\%, outperforming existing baselines by 63.9\%.

Lu et al. \cite{lu2025uxagent} developed UXAgent, a tool that uses LLM-Agent technology and a universal browser connector to simulate thousands of users for automated usability testing. It enables user experience (UX) researchers to quickly iterate on study designs by providing qualitative insights, quantitative action data, and video recordings before engaging participants.  Wang et al. \cite{wang2025training} introduce TRAVER (Trace-and-Verify), a novel agent workflow that combines knowledge tracing estimating a student's evolving knowledge state with turn-by-turn verification to ensure effective step-by-step guidance toward task completion. Alongside TRAVER, they propose DICT, an automatic evaluation protocol that utilizes controlled student simulation and code generation tests to assess the performance of tutoring agents holistically. SWE-Gym \cite{pan2024training} is introduced as the first dedicated environment for training real-world software engineering (SWE) agents, designed around 2,438 Python task instances that include complete codebases, executable runtime environments, unit tests, and natural language task descriptions. This realistic setup allows for training language model–based SWE agents that significantly improve performance achieving up to 19\% absolute gains in resolve rate on popular test sets like SWE-Bench Verified and Lite. Furthermore, the authors explore inference-time scaling by employing verifiers trained on agent trajectories sampled from SWE-Gym, which, when combined with their fine-tuned agents, achieve state-of-the-art performance of 32.0\% on SWE-Bench Verified and 26.0\% on SWE-Bench Lite.

\paragraph{Multi‑Agent Collaboration \& Simulation}

The works by Yang et al. \cite{yang2025multi}, Guo et al. \cite{guo2025syncmind}, and Islam et al. \cite{islam2025codesim} contribute significant advancements to the application of Large Language Models (LLMs) in code understanding, collaborative software engineering, and code generation. Yang et al. \cite{guo2025syncmind} propose a novel multi-agent collaboration framework to bridge the gap between different programming languages. By leveraging language-specific agents that collaborate and share knowledge, their approach enhances multilingual instruction tuning, enabling the efficient transfer of knowledge across languages. The Qwen2.5-xCoder model demonstrates superior performance in multilingual programming benchmarks, showcasing its potential to reduce cross-lingual gaps. Guo et al. \cite{guo2025syncmind} introduce SyncMind, a framework that defines the out-of-sync problem in collaborative software engineering. Through their SyncBench benchmark, which includes over 24,000 instances of out-of-sync scenarios from real-world codebases, they highlight performance gaps in current LLM agents and emphasize the need for better collaboration and resource-awareness in AI systems. Finally, Islam et al. \cite{islam2025codesim} present CodeSim, a multi-agent code generation framework that addresses program synthesis, coding, and debugging through a human-like perception approach. By incorporating plan verification and internal debugging via input/output simulation, CodeSim achieves state-of-the-art performance across multiple competitive benchmarks, including HumanEval, MBPP, APPS, and CodeContests. Their approach demonstrates the potential for further enhancement when coupled with external debuggers, advancing the effectiveness of code generation systems.

\subsubsection{Synthetic data generation}

Mitra et al. \cite{mitra2024agentinstruct} propose AgentInstruct, a novel framework that leverages synthetic data for post-training large language models through a process termed "Generative Teaching." Recognizing the challenges posed by the varying quality and diversity of synthetic data and the extensive manual curation typically required AgentInstruct automates the creation of high-quality instructional datasets using a multi-agent workflow. Starting from raw unstructured text and source code, the framework employs successive stages of content transformation, seed instruction generation across over 100 subcategories, and iterative instruction refinement via suggester-editor pairs. This process yields a dataset of 25 million prompt-response pairs covering diverse skills such as text editing, coding, creative writing, and reading comprehension. When applied to fine-tune a Mistral-7B model, the resulting Orca-3 model demonstrated significant performance improvements ranging from 19\% to 54\% across benchmarks like MMLU, AGIEval, GSM8K, BBH, and AlpacaEval as well as a notable reduction in hallucinations for summarization tasks. These findings underscore the potential of automated, agentic synthetic data generation to enhance model capabilities while reducing reliance on labor-intensive data curation, positioning AgentInstruct as a promising tool for advancing LLM instruction tuning.

\begin{figure}[htbp]
\centering
\resizebox{0.5\textwidth}{!}{%
\begin{tikzpicture}[
  mindmap,
  every node/.style={concept, circular drop shadow, minimum size=0.1cm},
  grow cyclic, align=flush center, concept color=black!50,
  level 1/.append style={
    sibling angle=45,
    level distance=4cm,
    font=\scriptsize
  },
  level 2/.append style={
    sibling angle=30,
    level distance=4cm,
    font=\scriptsize
  }
]
\node[concept color=black!50] {\Large Finance Applications}
  child[concept color=black!50] { node[align=center] {Structured Finance \&\\Automation}
    child { node[align=center] {Structured Finance Automation\\\cite{wan2024enhancing}} }
  }
  child[concept color=orange!50] { node[align=center] {Market\\Simulation}
    child { node[align=center] {TwinMarket\\\cite{yang2025twinmarket}} }
  }
  child[concept color=green!50] { node[align=center] {Sequential Investment\\Decision‑Making}
    child { node[align=center] {FinCon\\\cite{yu2024fincon}} }
  }
  child[concept color=teal!50] { node[align=center] {Strategic Behavior in\\Competitive Markets}
    child { node[align=center] {Strategic Behavior\\\cite{lin2024strategic}} }
  }
  child[concept color=red!50] { node[align=center] {Financial Reasoning\\\& QA}
    child { node[align=center] {Multi‑Agent Financial QA\\\cite{fatemi2024enhancing}} }
  }
  child[concept color=yellow!50] { node[align=center] {Stock Analysis \&\\Evaluation}
    child { node[align=center] {Multi‑Agent Collaboration\\\cite{han2024enhancing}} }
    child { node[align=center] {FinSphere\\\cite{han2025finsphere}} }
    child { node[align=center] {MarketSenseAI\\\cite{fatouros2025marketsenseai}} }
  }
  child[concept color=gray!50] { node[align=center] {Agentic Financial Modeling\\\& Risk Management}
    child { node[align=center] {Agentic Crews\\\cite{okpala2025agentic}} }
  }
  child[concept color=pink!50] { node[align=center] {Trustworthy Conversational\\Shopping Agents}
    child { node[align=center] {Citation‑Enhanced CSA\\\cite{zeng2025cite}} }
  }
;
\end{tikzpicture}%
}
\caption{Agent LLM Applications in Finance}
\label{fig:finance_applications}
\end{figure}
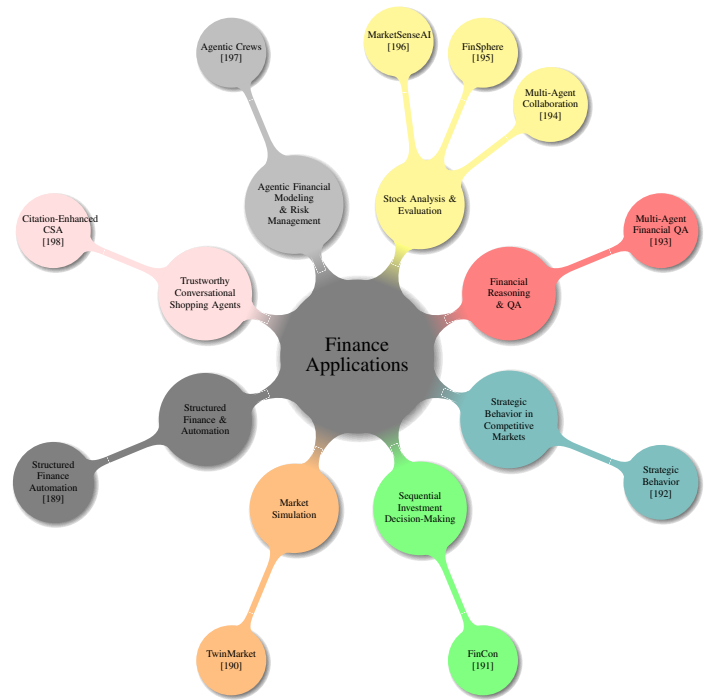

\subsubsection{Finance Applications}
Finance is a dynamic domain where the adoption of LLM-based agents has opened new avenues for automation, simulation, analysis, and decision support. This subsection presents recent innovations that span structured finance automation, market simulation, investment decision-making, financial reasoning, stock analysis, and risk management. Fig. \ref{fig:finance_applications} presents a classification of Agent LLM Applications for Finance.

\paragraph{Structured Finance and Automation}
Wan et al. \cite{wan2024enhancing} investigate the integration of artificial intelligence into structured finance, where the process of restructuring diverse assets into securities such as MBS, ABS, and CDOs presents substantial due diligence challenges. The authors demonstrate that AI, specifically large language models (LLMs), can effectively automate the verification of information between loan applications and bank statements. While close-sourced models like GPT-4 achieve superior performance, open-sourced alternatives such as LLAMA3 provide a more cost-effective option. Furthermore, implementing dual-agent systems has been shown to further increase accuracy, albeit with higher operational costs.

\paragraph{Market Simulation}
Yang et al. \cite{yang2025twinmarket} introduce TwinMarket, a multi-agent framework that harnesses large language models (LLMs) to simulate complex socio-economic systems, addressing longstanding challenges in modeling human behavior. Traditional rule-based agent-based models often fall short in capturing the irrational and emotionally driven aspects of decision-making emphasized in behavioral economics. TwinMarket leverages the cognitive biases and dynamic emotional responses inherent in LLMs to create more realistic simulations of socio-economic interactions. The study illustrates how individual agent behaviors can lead to emergent phenomena such as financial bubbles and recessions when combined through feedback mechanisms through experiments conducted in a simulated stock market environment. 

\paragraph{Sequential Investment Decision-Making}
Yu et al. \cite{yu2024fincon} propose FinCon, an LLM-based multi-agent framework designed to tackle the complexities of sequential financial investment decision-making. Recognizing that effective investment requires dynamic interaction with volatile environments, FinCon draws inspiration from real-world investment firm structures by establishing a manager-analyst communication hierarchy. This design facilitates synchronized, cross-functional collaboration through natural language interactions while endowing each agent with enhanced memory capacity. A key component is the risk-control module, which periodically triggers a self-critiquing mechanism to update systematic investment beliefs, thereby reinforcing future agent behavior and reducing unnecessary communication overhead. FinCon exhibits strong generalization across various financial tasks, such as stock trading and portfolio management, and offers a promising approach to synthesizing multi-source information for optimized decision-making in dynamic financial markets.

\paragraph{Strategic Behavior in Competitive Markets}
Li et al. \cite{lin2024strategic} investigate the strategic behavior of large language models (LLMs) when deployed as autonomous agents in multi-commodity markets within the framework of Cournot competition. The authors examine whether these models can independently engage in anti-competitive practices, such as collusion or market division, without explicit human intervention. Their findings reveal that LLMs can monopolize specific commodities by dynamically adjusting pricing and resource allocation strategies, thereby maximizing profitability through self-directed strategic decisions. These results present significant challenges and potential opportunities for businesses incorporating AI into strategic roles and regulatory bodies responsible for maintaining fair market competition. 

\paragraph{Financial Reasoning and QA}
Fatemi et al. \cite{fatemi2024enhancing} address the limitations of large language models (LLMs) in financial question-answering (QA) tasks that require complex numerical reasoning. Recognizing that multi-step reasoning is essential for extracting and processing information from tables and text, the authors propose a multi-agent framework incorporating a critical agent to evaluate the reasoning process and final answers. The framework is further enhanced with multiple critic agents specializing in distinct aspects of the answer evaluation. Experimental results show that this multi-agent approach significantly boosts performance, with an average increase of 15\% for the LLaMA3-8B model and 5\% for the LLaMA3-70B model, compared to single-agent systems. Moreover, the proposed system performs comparably to and sometimes exceeds the capabilities of much larger single-agent models such as LLaMA3.1-405B and GPT-4o-mini, although it slightly lags behind Claude-3.5 Sonnet. 

\paragraph{Stock Analysis and Evaluation}

Han et al. \cite{han2024enhancing} present a novel multi-agent collaboration system designed to enhance financial analysis and investment decision-making by leveraging the collaborative potential of multiple AI agents. Moving beyond traditional single-agent models, the system features configurable agent groups with diverse collaboration structures that dynamically adapt to varying market conditions and investment scenarios through a sub-optimal combination strategy. The study focuses on three key sub-tasks fundamentals, market sentiment, and risk analysis applied to the 2023 SEC 10-K forms of 30 companies from the Dow Jones Index. Experimental findings reveal significant performance improvements with multi-agent configurations compared to single-agent approaches, demonstrating enhanced accuracy, efficiency, and adaptability.

In a related study, Han et al. \cite{han2025finsphere} introduce FinSphere, a conversational stock analysis agent designed to overcome two major challenges faced by current financial LLMs: their insufficient depth in stock analysis and the lack of objective metrics for evaluating the quality of analysis reports. The authors make three significant contributions. First, they present Stocksis, a dataset curated by industry experts to enhance the stock analysis capabilities of LLMs. Second, they propose Analyscore, a systematic evaluation framework that objectively assesses the quality of stock analysis reports. Third, they develop FinSphere, an AI agent that leverages real-time data feeds, quantitative tools, and an instruction-tuned LLM to generate high-quality stock analysis in response to user queries. Experimental results indicate that FinSphere outperforms general and domain-specific LLMs and existing agent-based systems, even when these systems are enhanced with real-time data and few-shot guidance.

Fatouros et al. \cite{fatouros2025marketsenseai} introduce MarketSenseAI, an innovative framework for comprehensive stock analysis that harnesses large language models (LLMs) to integrate diverse financial data sources ranging from financial news, historical prices, and company fundamentals to macroeconomic indicators. Leveraging a novel architecture that combines Retrieval-Augmented Generation with LLM agents, MarketSenseAI processes SEC filings, earnings calls, and institutional reports to enhance macroeconomic analysis. The latest advancements in the framework yield significant improvements in fundamental analysis accuracy over its previous iteration. Empirical evaluations on S\&P 100 stocks (2023–2024) reveal cumulative returns of 125.9\% versus the index’s 73.5\%, while tests on S\&P 500 stocks in 2024 show a 33.8\% higher Sortino ratio, underscoring the scalability and robustness of this LLM-driven investment strategy.

\paragraph{Agentic Financial Modeling and Risk Management}

Okpala et al. \cite{okpala2025agentic} examine integrating large language models into agentic systems within the financial services industry, focusing on automating complex modeling and model risk management (MRM) tasks. The authors introduce the concept of agentic crews, where teams of specialized agents, coordinated by a manager, collaboratively execute distinct functions. The modeling crew handles tasks such as exploratory data analysis, feature engineering, model selection, hyperparameter tuning, training, evaluation, and documentation, while the MRM crew focuses on compliance checks, model replication, conceptual validation, outcome analysis, and documentation. The effectiveness and robustness of these agentic workflows are demonstrated through numerical examples applied to datasets in credit card fraud detection, credit card approval, and portfolio credit risk modeling, highlighting the potential for autonomous decision-making in financial applications.

\paragraph{Trustworthy Conversational Shopping Agents}
Zeng et al. \cite{zeng2025cite} focuses on enhancing the trustworthiness of LLM-based Conversational Shopping Agents (CSAs) by addressing two key challenges: the generation of hallucinated or unsupported claims and the lack of knowledge source attribution. To combat these issues, the authors propose a production-ready solution that integrates a "citation experience" through In-context Learning (ICL) and Multi-UX-Inference (MUI). This approach enables CSAs to include citation marks linked to relevant product information without disrupting user experience features. Additionally, the work introduces automated metrics and scalable benchmarks to evaluate the grounding and attribution capabilities of LLM responses holistically. Experimental results on real-world data indicate that incorporating this citation generation paradigm enhances response grounding by 13.83\%, ultimately improving transparency and building customer trust in conversational AI within the e-commerce domain.

\subsubsection{Chemical Reasoning}
The domain of chemical reasoning poses complex challenges for large language models, including precise information processing, task decomposition, and integrating scientific knowledge and code. In this subsection, we highlight recent advances in developing LLM-based agents for chemical reasoning and materials discovery. 

\paragraph{Chemical Reasoning \& Information Processing}

The paper by Cho et al. \cite{cho2025building} addresses the challenges of deploying large language model (LLM)–powered agents in resource-constrained environments, particularly for specialized domains and less-common languages, by introducing Tox-chat a Korean chemical toxicity information agent. It presents a context-efficient architecture utilizing hierarchical section search to reduce token consumption and a scenario-based dialogue generation methodology that distills tool-using capabilities from larger models. Experimental evaluations reveal that the fine-tuned 8B-parameter model significantly surpasses untuned models and baseline approaches in database faithfulness and user preference, offering promising strategies for developing efficient, domain-specific language agents under practical constraints.

Chemical reasoning tasks, which involve complex multi-step processes and require precise calculations, pose unique challenges for LLMs, especially in handling domain-specific formulas and integrating code accurately. ChemAgent  \cite{tang2025chemagent} addresses these challenges by decomposing chemical tasks into manageable sub-tasks and compiling them into a structured memory library that can be referenced and refined in future queries. The framework incorporates three types of memory and a library-enhanced reasoning component, enabling the system to improve over time through experience. Evaluations on four SciBench chemical reasoning datasets reveal that ChemAgent achieves performance gains of up to 46\% with GPT-4, significantly outperforming existing methods and suggesting promising applications in fields such as drug discovery and materials science.

\paragraph{Materials Discovery \& Design}

By collaborating with materials science experts, Kumbhar et al. \cite{kumbhar2025hypothesis} curate a novel dataset from recent journal publications that encapsulate real-world design goals, constraints, and methodologies. Using this dataset, they test LLM-based agents to generate viable hypotheses to achieve specified objectives under given constraints. To rigorously assess the relevance and quality of these hypotheses, a novel scalable evaluation metric is proposed that mirrors the critical assessment process of materials scientists. Together, the curated dataset, the hypothesis generation method, and the evaluation framework provide a promising foundation for future research to accelerate materials discovery and design using LLM. ChemAgent is a novel framework that aims to enhance chemical reasoning by leveraging large language models through a dynamic, self-updating library.

\begin{figure}[htbp]
\centering
\resizebox{0.5\textwidth}{!}{%
\begin{tikzpicture}[
  mindmap,
  every node/.style={concept, circular drop shadow, minimum size=0.1cm},
  grow cyclic, align=flush center, concept color=black!50,
  level 1/.append style={
    sibling angle=120,
    level distance=4cm,
    font=\scriptsize
  },
  level 2/.append style={
    sibling angle=30,
    level distance=4cm,
    font=\scriptsize
  }
]
\node[concept color=black!50] {\Large Solving Mathematical Problems}
  child[concept color=red!50] { node[align=center] {Mathematical Reasoning\\\& Problem Solving}
    child { node[align=center] {MACM\\\cite{lei2024macm}} }
    child { node[align=center] {MathLearner\\\cite{xie2024mathlearner}} }
    child { node[align=center] {Prompt Sampling\\\cite{lee2024expanding}} }
    child { node[align=center] {Flows\\\cite{deng2024flow}} }
    child { node[align=center] {KG‑Proofs\\\cite{li2025automating}} }
    child { node[align=center] {MA‑LoT\\\cite{wang2025ma}} }
  }
  child[concept color=orange!50] { node[align=center] {Educational \&\\Tutoring Applications}
    child { node[align=center] {MATHVC\\\cite{yue2024mathvc}} }
    child { node[align=center] {PACE\\\cite{liu2025sizedoesntfitall}} }
  }
  child[concept color=green!50] { node[align=center] {Numerical\\Reasoning}
    child { node[align=center] {Agent Trading Arena\\\cite{ma2025llm}} }
  }
;
\end{tikzpicture}%
}
\caption{Agent LLM Applications in Solving Mathematical Problems}
\label{fig:math_problem_solving}
\end{figure}
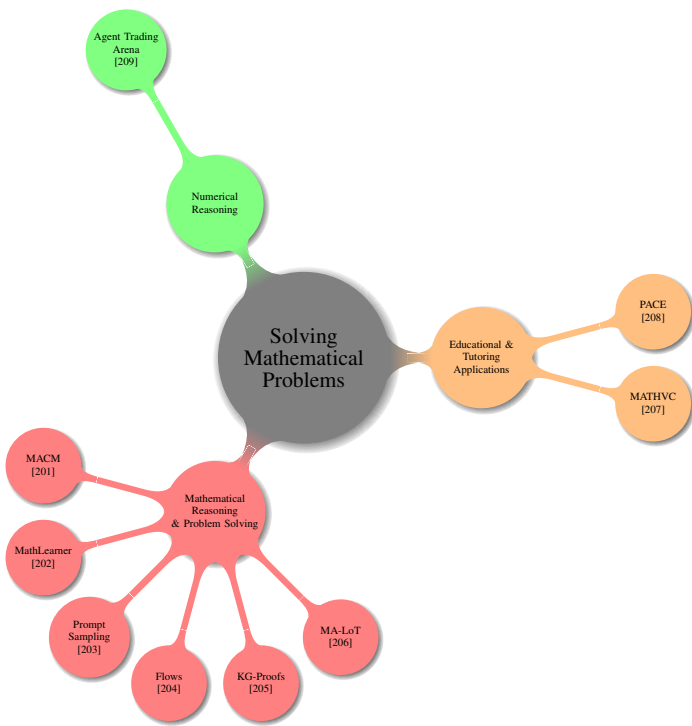

\begin{table*}[htbp]
\centering
\scriptsize
\caption{Overview of AI Agent Applications for Mathematical Problem Solving}
\label{tab:ai_agent_apps_math}
\begin{adjustbox}{max width=\textwidth}
\begin{tabularx}{\textwidth}{@{}%
  >{\raggedright\arraybackslash}p{1.8cm}   
  >{\centering\arraybackslash}p{0.6cm}      
  >{\centering\arraybackslash}p{1.2cm}      
  >{\raggedright\arraybackslash}X           
  >{\raggedright\arraybackslash}X           
  >{\raggedright\arraybackslash}X           
  >{\centering\arraybackslash}p{1.2cm}      
  >{\centering\arraybackslash}p{1.5cm}      
  >{\centering\arraybackslash}p{1.5cm}      
@{}}
\toprule
\textbf{Agent / Tool} & \textbf{Year} & \textbf{Math Task} & \textbf{Primary Objective} & \textbf{Architecture \& Workflow} & \textbf{Key Outcomes \& Metrics} & \textbf{Proof Val.} & \textbf{Solver Integr.} & \textbf{Notation Sup.} \\
\midrule
MACM \cite{lei2024macm} 
  & 2024 
  & Advanced Reasoning 
  & Solve multi-step math problems with robust generalization. 
  & Multi-Agent Conditional Mining prompting for iterative refinement. 
  & MATH level 5 accuracy increase from 54.68\% to 76.73\% on GPT-4 Turbo. 
  & \emptycirc 
  & \emptycirc 
  & \emptycirc \\
\midrule
MathLearner \cite{xie2024mathlearner} 
  & 2024 
  & Inductive Reasoning 
  & Enhance LLM reasoning via inductive retrieval and application. 
  & Retrieval module plus procedural knowledge injection in inductive loop. 
  & +20.96\% global accuracy; solves 17.54\% previously unsolved problems. 
  & \emptycirc 
  & \emptycirc 
  & \emptycirc \\
\midrule
Prompt Sampling \cite{lee2024expanding} 
  & 2024 
  & Search Space Expansion 
  & Combine diverse prompting methods to expand search space efficiently. 
  & Uniform sampling over multiple prompt strategies; fewer inference runs. 
  & 43\% fewer runs for MATH-hard with maximal coverage. 
  & \emptycirc 
  & \emptycirc 
  & \emptycirc \\
\midrule
Flows \cite{deng2024flow} 
  & 2024 
  & Reasoning Trace 
  & Generate detailed math reasoning traces online. 
  & Collaborative LLM ensemble with online DPO and rollouts. 
  & Significant improvement in reasoning quality versus direct inference. 
  & \emptycirc 
  & \emptycirc 
  & \emptycirc \\
\midrule
KG-Proof Agent \cite{li2025automating} 
  & 2025 
  & Proof Construction 
  & Automate formalization of proofs using knowledge graphs. 
  & Integrates concept KG with LLM to structure lemmas and steps. 
  & 34\% success on MUSTARDSAUCE; 2--11\% improvement over baselines. 
  & \halfcirc 
  & \emptycirc 
  & \halfcirc \\
\midrule
MA-LoT \cite{wang2025ma} 
  & 2025 
  & Theorem Proving 
  & Synergize natural-language reasoning with Lean4 verification feedback. 
  & Multi-agent chain-of-thought plus LoT-Transfer pipeline in Lean4. 
  & 61.07\% on MiniF2F-Test (Lean4) versus 22.95\% for GPT-4. 
  & \fullcirc 
  & \fullcirc 
  & \fullcirc \\
\midrule
MATHVC \cite{yue2024mathvc} 
  & 2024 
  & Educational Modeling 
  & Simulate group discussions for mathematical modeling skills. 
  & Virtual classroom with diverse student-agents and meta planning. 
  & Realistic dialog; improves modeling task performance. 
  & \emptycirc 
  & \emptycirc 
  & \emptycirc \\
\midrule
PACE \cite{liu2025sizedoesntfitall} 
  & 2025 
  & Personalized Tutoring 
  & Tailor math instruction to learning styles with Socratic feedback. 
  & Felder-Silverman personas plus Socratic method and tailored data. 
  & Higher engagement and outcomes versus traditional tutors. 
  & \emptycirc 
  & \emptycirc 
  & \emptycirc \\
\midrule
Agent Trading Arena \cite{ma2025llm} 
  & 2025 
  & Numerical Reasoning 
  & Improve numeric inference with visual data and reflection. 
  & Virtual stock game plus analysis over plots and charts. 
  & Enhanced geometric reasoning; validated on NASDAQ dataset. 
  & \emptycirc 
  & \emptycirc 
  & \halfcirc \\
\bottomrule
\end{tabularx}
\end{adjustbox}\\
\textbf{Proof Val.}: Proof Validation; 
\textbf{Solver Integr.}: Solver \& Assistant Integration; 
\textbf{Notation Sup.}: Notation \& Formalism Support: \halfcirc: Partial; \emptycirc: Not Supported; \fullcirc: Supported.
\end{table*}

\subsubsection{Solving mathematical problems}
Mathematical problem-solving remains a fundamental challenge for large language models due to the need for structured reasoning, formal logic, and precise numerical computation. In this subsection, we present recent efforts to enhance the mathematical capabilities of LLM-based agents through novel prompting strategies, collaborative agent systems, theorem proving, and knowledge integration. Fig. \ref{fig:math_problem_solving} presents a classification of agent LLM applications for solving mathematical problems.

\paragraph{Mathematical Reasoning and Problem Solving}

The paper by Lei et al. \cite{lei2024macm} tackles the challenge of advanced mathematical problem-solving in large language models (LLMs), where performance significantly declines despite recent advancements like GPT-4. While methods such as Tree of Thought and Graph of Thought have been explored to enhance logical reasoning, they face notable limitations: their effectiveness on complex problems is limited, and the need for custom prompts for each problem restricts generalizability. In response, the authors introduce the Multi-Agent System for Conditional Mining (MACM) prompting method. MACM successfully addresses intricate, multi-step mathematical challenges and exhibits robust generalization across diverse mathematical contexts. Notably, using MACM, the accuracy of GPT-4 Turbo on level five problems in the MATH dataset improves markedly from 54.68\% to 76.73\%, demonstrating its potential to elevate LLM inferential capabilities substantially. 

Xie et al. \cite{xie2024mathlearner} present an agent framework designed to enhance the mathematical reasoning abilities of large language models (LLMs) through inductive reasoning. Drawing inspiration from the human learning process of generalizing information and applying prior knowledge to new tasks, the framework significantly outperforms traditional chain-of-thought approaches. Specifically, it improves global accuracy by 20.96\%  and can solve 17.54\% of mathematical problems that the baseline fails to address. A key framework component is its efficient retrieval method, which enables the model to effectively incorporate external knowledge and support mathematical computations based on explicit written procedures. 

Lee et al. \cite{lee2024expanding} investigate the limitations of traditional single prompting methods in large language models (LLMs) for mathematical reasoning and explore alternative prompting strategies. It experimentally demonstrates that distinct prompting methods each probe unique search spaces, a differentiation that becomes more pronounced with increased problem complexity. To capitalize on this diversity, the study introduces an efficient sampling process that uniformly combines outputs from these varied methods, thereby expanding the overall search space and achieving improved performance with fewer inference runs. Notably, for the particularly challenging problems in the MATH-hard subset, the approach reached maximal search space utilization with approximately 43\% fewer runs compared to individual methods.

Deng et al. \cite{deng2024flow} introduce a novel approach to enhance the generation of detailed and accurate reasoning traces in large language models (LLMs), particularly for mathematical reasoning tasks. The authors propose an online learning framework termed "Flows," where component LLMs work collaboratively and iteratively, engaging in incremental output production to build coherent solutions. Central to the approach is online Direct Preference Optimization (DPO) with rollouts, which generates DPO pairs for each training example and updates the models in real-time. By directly comparing the quality of reasoning traces produced by this method against those generated by standard direct model inference, the study demonstrates that the proposed Flow framework significantly improves LLM performance in mathematical reasoning.

Li et al. \cite{li2025automating} introduce a novel framework that augments large language models (LLMs) with knowledge graphs to improve the construction and formalization of mathematical proofs. The proposed approach tackles persistent challenges in automating the identification of key mathematical concepts, understanding their relationships, and embedding them within rigorous logical frameworks. Experimental results show significant performance gains, with the framework achieving up to a 34\% success rate on the MUSTARDSAUCE dataset on o1-mini and consistently outperforming baseline models by 2–11\% across various benchmarks. 

Wang et al. \cite{wang2025ma} introduce MA-LoT, a novel multi-agent framework designed for the Lean4 theorem proving that it synergizes high-level natural language reasoning with formal language verification feedback. Unlike traditional single-agent approaches that either generate complete proofs or perform tree searches, MA-LoT leverages structured interactions among multiple agents to maintain long-term coherence and deeper insight during proof generation. The framework employs a novel LoT-Transfer Learning training-inference pipeline that harnesses long chain-of-thought processes' emergent formal reasoning abilities. Extensive experiments demonstrate that MA-LoT achieves a 61.07\% accuracy on the Lean4 version of the MiniF2F-Test dataset, significantly outperforming baselines such as GPT-4 (22.95\%), single-agent tree search methods (50.70\%), and whole-proof generation techniques (55.33\%). These results underscore the potential of integrating long chain-of-thought reasoning with formal verification to enhance automated theorem proving.

\paragraph{Educational and Tutoring Applications}

Yue et al. \cite{yue2024mathvc} introduce MATHVC, a pioneering virtual classroom powered by large language models (LLMs) designed to enhance students' mathematical modeling (MM) skills through collaborative group discussions. Recognizing that traditional MM practice often suffers from uneven access to qualified teachers and resources, the authors leverage LLMs' capabilities to simulate diverse student characters, each embodying distinct math-relevant properties. To ensure that these simulated interactions mirror authentic student discussions, the framework incorporates three key innovations: integrating domain-specific MM knowledge into the simulation, defining a symbolic schema to ground character behaviors, and employing a meta planner to guide the conversational flow.

Liu et al. \cite{liu2025sizedoesntfitall} introduce the Personalized Conversational Tutoring Agent (PACE) for mathematics instruction, addressing a critical gap in intelligent educational systems by adapting to individual learner characteristics. PACE leverages the Felder and Silverman learning style model to simulate distinct student personas, enabling the system to tailor teaching strategies to diverse learning styles, a crucial factor for enhancing engagement and comprehension in mathematics. Integrating the Socratic teaching method, PACE provides instant, reflective feedback that encourages deeper cognitive processing and critical thinking. The framework also involves constructing personalized teaching datasets and training specialized models, which facilitate identifying and adapting each student’s unique needs. Extensive evaluations using multi-aspect criteria demonstrate that PACE outperforms traditional methods in personalizing the educational experience and boosting student motivation and learning outcomes.

\paragraph{Numerical Reasoning}

Ma et al. \cite{ma2025llm} investigate the limitations of large language models (LLMs) in handling dynamic and unseen numerical reasoning tasks, mainly when operating on plain-text data. To address this, the authors introduce the Agent Trading Arena, a virtual numerical game simulating complex economic systems via zero-sum stock portfolio investments, which better reflects real-world scenarios where optimal solutions are not clearly defined. Experimental results indicate that LLMs, including GPT-4o, face challenges with algebraic reasoning in textual formats, often focusing on local details at the expense of broader trends. In contrast, when LLMs are provided with visual data representations, such as scatter plots or K-line charts, they exhibit significantly enhanced geometric reasoning capabilities. This improvement is further enhanced by incorporating a reflection module that facilitates the analysis and interpretation of complex data. These findings are validated using the NASDAQ Stock dataset, underscoring the value of visual inputs for bolstering numerical reasoning in LLMs.

\subsubsection{Geography Applications}

Yu et al. \cite{yu2024mineagent} introduce MineAgent, a modular framework designed to enhance the capabilities of multimodal large language models (MLLMs) in the domain of remote-sensing mineral exploration. This field presents significant challenges, including the need for domain-specific geological knowledge and the complexity of reasoning across multiple remote-sensing images, which is further complicated by long-context issues. MineAgent addresses these challenges by incorporating hierarchical judging and decision-making modules to improve multi-image reasoning and spatial-spectral integration. In addition, the authors propose MineBench, a specialized benchmark to evaluate MLLMs on mineral exploration tasks using geological and hyperspectral data. Extensive experiments demonstrate the effectiveness of MineAgent, showcasing its potential to significantly advance the use of MLLMs in the critical area of remote-sensing mineral exploration

Ning et al. \cite{ning2025autonomous} introduce an autonomous geographic information system (GIS) agent framework that utilizes large language models (LLMs) to perform spatial analyses and cartographic tasks. A significant research gap in the field has been the ability of these agents to autonomously discover and retrieve the necessary geospatial data. The proposed framework addresses this by generating, executing, and debugging programs to select data sources from a predefined list, using source-specific handbooks that document metadata and retrieval details. The framework is designed in a plug-and-play style, allowing users or automated crawlers to easily add new data sources by creating additional handbooks. A prototype of the agent has been developed as a QGIS plugin and Python program. Experimental results demonstrate its capability to retrieve data from various sources, including OpenStreetMap, U.S. Census Bureau demographic data, satellite basemaps from ESRI, global digital elevation models from OpenTopography, weather data, and COVID-19 case data from the NYTimes GitHub. This work is one of the first efforts to create an autonomous GIS agent for geospatial data retrieval, marking a significant advancement in spatial data automation.

\begin{table*}[htbp]
\centering
\scriptsize
\caption{Overview of AI Agent Applications for Multimedia}
\label{tab:ai_agent_apps_mm}
\begin{adjustbox}{max width=\textwidth}
\begin{tabularx}{\textwidth}{@{}%
  >{\raggedright\arraybackslash}p{1.8cm}   
  >{\centering\arraybackslash}p{0.6cm}      
  >{\centering\arraybackslash}p{1.2cm}      
  >{\raggedright\arraybackslash}X           
  >{\raggedright\arraybackslash}X           
  >{\raggedright\arraybackslash}X           
  >{\raggedright\arraybackslash}p{1.5cm}    
  >{\raggedright\arraybackslash}p{1.5cm}    
  >{\raggedright\arraybackslash}p{1.5cm}    
@{}}
\toprule
\textbf{Agent/Tool} & \textbf{Year} & \textbf{Domain} & \textbf{Primary Objective} & \textbf{Architecture \& Workflow} & \textbf{Key Outcomes \& Metrics} & \textbf{Eval.\ Metrics} & \textbf{Pipeline Integr.} & \textbf{Fmt.\ Compat.} \\
\midrule
FilmAgent \cite{xu2025filmagent} 
  & 2025 
  & Film Automation 
  & Fully automate end‑to‑end 3D virtual film production. 
  & Multi‑agent roles (director, screenwriter, actors, cinematographer) with iterative feedback loops. 
  & Outperforms single‑agent baselines with coherent video across 15 scenarios. 
  & Mean user score 3.98/5 
  & Virtual studio pipeline support 
  & Exports MP4/WebM \\
\midrule
AesopAgent \cite{wang2024aesopagent} 
  & 2024 
  & Video 
  & Convert story drafts into scripts, images, audio, and video. 
  & Two‑layer RAG‑evolutionary workflow plus utility layer for image/audio/effects. 
  & Rich, coherent multimodal outputs with continuous optimization. 
  & Workflow convergence rate $\approx$\ 85 \% 
  & Integrates with AIGC asset generators 
  & Supports PNG, WAV, MP4 \\
\midrule
IBSEN \cite{han2024ibsen} 
  & 2024 
  & Drama Scripts 
  & Generate coherent drama scripts via director–actor coordination. 
  & Director agent outlines plot; actor agents role‑play and adjust narrative. 
  & Diverse, complete scripts preserving character traits. 
  & Narrative coherence > 90\% (human eval) 
  & Scriptwriting toolchain compatible 
  & Plain‑text script output \\
\midrule
Fashion‑Agent \cite{maronikolakis2024should} 
  & 2024 
  & Conversational Retail 
  & Enhance online fashion discovery with LLM dialogue agents. 
  & LLM front‑end connects to search \& recommendation backends. 
  & 4 000‑dialog dataset; improves retrieval relevance by 18 \%. 
  & Precision@5: 78\% 
  & E‑commerce API integration 
  & JSON / HTML widget \\
\midrule
ComposerX \cite{deng2024composerx} 
  & 2024 
  & Music Composition 
  & Multi‑agent symbolic music generation with harmony constraints. 
  & Agents specialize in melody, harmony, and structure using LLM reasoning. 
  & Coherent polyphonic pieces rated high on musicality. 
  & Subjective rating 4.2/5 
  & MIDI pipeline plugin 
  & Standard MIDI files \\
\midrule
MusicAgent \cite{yu2023musicagent} 
  & 2023 
  & Music Processing 
  & Orchestrate diverse music tasks via unified LLM agent. 
  & Autonomous task decomposition and tool invocation over HF/GitHub/APIs. 
  & Simplifies tool use; reduces development effort by 40 \%. 
  & Task completion time ↓ 40 \% 
  & Integrates FFmpeg, Librosa, Web APIs 
  & WAV, MP3, MIDI \\
\midrule
PoetryAgents \cite{zhang2024llm} 
  & 2024 
  & Poetry Generation 
  & Boost diversity \& novelty in LLM‑generated poetry via multi‑agent social learning. 
  & Cooperative \& non‑cooperative agent interactions on GPT‑2/3/4. 
  & +3.0–3.7 pp diversity; +5.6–11.3 pp novelty. 
  & Distinct n‑gram ↑ 11\% 
  & Text pipeline integration 
  & UTF‑8 text \\
\midrule
LyricAgents \cite{liu2024agent} 
  & 2024 
  & Lyric Generation 
  & Melody‑to‑lyric alignment in tonal languages with multi‑agent sub‑tasks. 
  & Agents for rhyme, syllable, alignment \& consistency; evaluated via singing synth. 
  & Listening test accuracy 85 \%. 
  & Alignment score 0.87 
  & Singing‑synth pipeline ready 
  & LRC / JSON lyric files \\
\bottomrule
\end{tabularx}
\end{adjustbox}\\
\textbf{Eval.\ Metrics}: Evaluation Metrics; \textbf{Pipeline Integr.}: Pipeline Integration; \textbf{Fmt.\ Compat.}: Format Compatibility.
\end{table*}

\begin{figure}[htbp]
\centering
\resizebox{0.5\textwidth}{!}{%
\begin{tikzpicture}[
  mindmap,
  every node/.style={concept, circular drop shadow, minimum size=0.1cm},
  grow cyclic, align=flush center, concept color=black!50,
  level 1/.append style={
    sibling angle=20,
    level distance=10cm,
    font=\large
  },
  level 2/.append style={
    sibling angle=36,
    level distance=5cm,
    font=\large
  }
]
\node[concept color=black!50] {\Large Multimedia Applications}
  child[concept color=red!50] { node[align=center] {Film Automation\\Agents}
    child { node[align=center] {FilmAgent\\\cite{xu2025filmagent}} }
  }
  child[concept color=orange!50] { node[align=center] {Story‑to‑Video Production\\Agents}
    child { node[align=center] {AesopAgent\\\cite{wang2024aesopagent}} }
  }
  child[concept color=green!50] { node[align=center] {Drama Script\\Generation Agents}
    child { node[align=center] {IBSEN\\\cite{han2024ibsen}} }
  }
  child[concept color=teal!50] { node[align=center] {Fashion‑Domain Conversational\\Agents}
    child { node[align=center] {Fashion Assistant Eval.\\\cite{maronikolakis2024should}} }
  }
  child[concept color=yellow!50] { node[align=center] {Symbolic Music\\Composition Agents}
    child { node[align=center] {ComposerX\\\cite{deng2024composerx}} }
  }
  child[concept color=purple!50] { node[align=center] {Music Understanding \&\\Generation Agents}
    child { node[align=center] {MusicAgent\\\cite{yu2023musicagent}} }
  }
  child[concept color=gray!50] { node[align=center] {Poetry Generation\\Agents}
    child { node[align=center] {Multi‑Agent Poetry\\Framework\\\cite{zhang2024llm}} }
  }
  child[concept color=pink!50] { node[align=center] {Lyric Generation\\Agents}
    child { node[align=center] {Melody‑Lyric Agents\\\cite{liu2024agent}} }
  }
;
\end{tikzpicture}%
}
\caption{Agent LLM Applications in Multimedia}
\label{fig:multimedia_applications}
\end{figure}
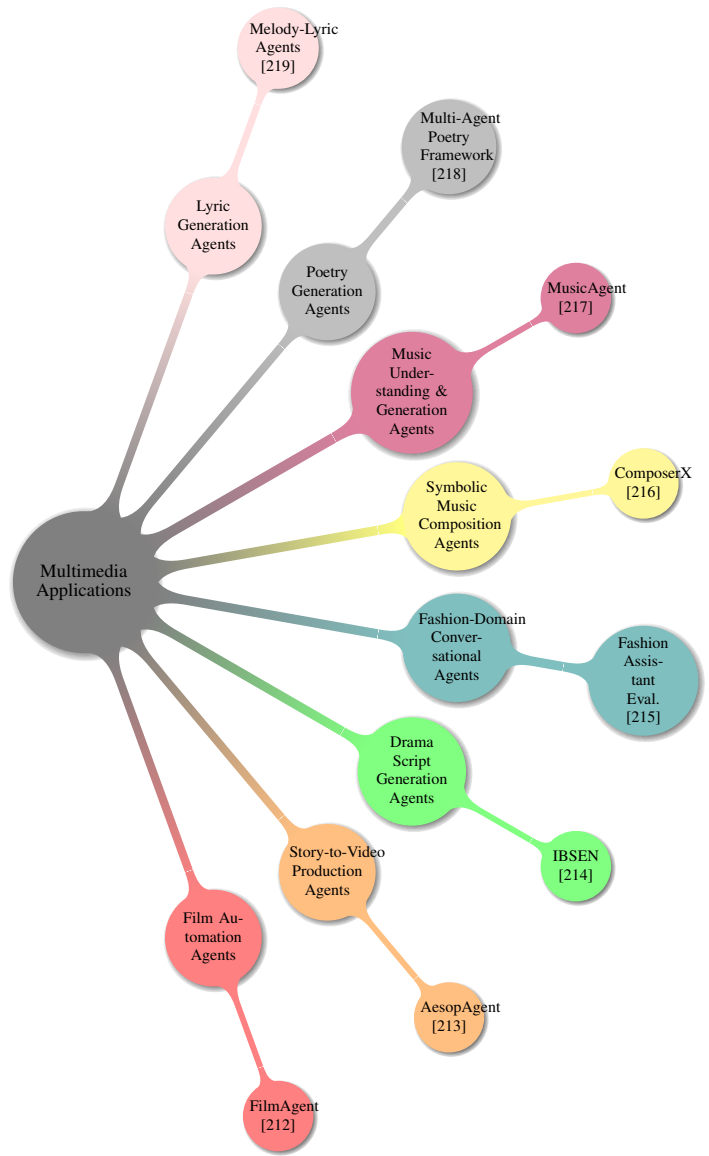

\subsubsection{Multimedia Applications}
Multimedia is an emerging frontier for LLM-based agents, where creative and interpretive tasks require coordination across diverse modalities, including text, audio, image, and video. In this subsection, we present recent advancements in applying agent-based language learning and machine learning (LLM) systems to domains such as film production, music and poetry generation, drama scripting, fashion assistance, and lyric composition. Fig. \ref{fig:multimedia_applications} presents a classification of agent LLM applications for Multimedia.

\paragraph{Film Automation Agents}

Xu et al. \cite{xu2025filmagent} introduce FilmAgent, an innovative LLM-based multi-agent collaborative framework designed to automate end-to-end film production within 3D virtual spaces. Virtual film production involves complex decision-making, including scriptwriting, cinematography, and actor positioning. FilmAgent simulates various crew roles such as directors, screenwriters, actors, and cinematographers, covering crucial stages of the film production process. These stages include idea development, where brainstormed ideas are transformed into structured story outlines; scriptwriting, which generates dialogues and character actions; and cinematography, which determines the camera setups for each shot. The agents collaborate iteratively, providing feedback and revisions to verify intermediate scripts and reduce hallucinations. Evaluations of the generated videos on 15 ideas across four key aspects show that FilmAgent outperforms all baselines, achieving an average score of 3.98 out of 5. Despite using the GPT-4o model, FilmAgent surpasses the single-agent o1, demonstrating the benefits of a coordinated multi-agent system.

\paragraph{Story‑to‑Video Production Agents}
Wang et al. \cite{wang2024aesopagent} introduce AesopAgent, an Agent-driven Evolutionary System designed for story-to-video production, leveraging the advancements in Agent and Artificial Intelligence Generated Content (AIGC) technologies. AesopAgent integrates multiple generative capabilities within a unified framework, enabling users to easily convert story proposals into scripts, images, audio, and videos. The system orchestrates the entire video generation workflow, ensuring that the generated content is both rich and coherent. The system consists of two layers: the Horizontal Layer and the Utility Layer. The Horizontal Layer incorporates a novel RAG-based evolutionary system that continuously optimizes the video production process by accumulating expert knowledge and refining workflow steps, such as LLM prompt optimization. The Utility Layer provides essential tools for consistent image generation, ensuring visual coherence in terms of composition, characters, and style, while also integrating audio and special effects.

\paragraph{Drama Script Generation Agents}
Han et al. \cite{han2024ibsen} introduce IBSEN, a director-actor coordination agent framework designed to generate drama scripts and provide greater control over the plot development, especially in scenarios where human players are involved. While current language model agents excel at creating individual behaviors for characters, they often struggle with maintaining consistency and coherence at the storyline level. IBSEN addresses this by introducing a director agent that writes plot outlines based on user input, instructs actor agents to role-play their respective characters, and adjusts the plot as needed to ensure that the narrative progresses toward the intended objective. The framework was evaluated using a novel drama plot involving multiple actor agents, where the director agent guided the interactions. The results demonstrate that IBSEN is capable of generating diverse and complete drama scripts from a rough plot outline, while preserving the unique characteristics of each character, showing the effectiveness of the framework in producing controlled, dynamic narrative content.

\paragraph{Fashion‑Domain Conversational Agents}
Maronikolakis et al. \cite{maronikolakis2024should}  focus on the potential of Large Language Models (LLMs) to revolutionize online fashion retail by enhancing customer experiences and improving product discovery through conversational agents. These LLM-powered agents allow customers to interact naturally, refining their needs and receiving personalized fashion and shopping advice. For tasks like finding specific products, conversational agents must translate customer interactions into calls to various backend systems, such as search engines, to display relevant product options. The authors emphasize the importance of evaluating the capabilities of LLMs in these tasks, particularly in integrating with backend systems. However, existing evaluations are often complex due to the lack of high-quality, relevant datasets that align with business needs. To address this, the authors developed a multilingual evaluation dataset comprising 4,000 conversations between customers and a fashion assistant on a large e-commerce platform.

\paragraph{Symbolic Music Composition Agents}
Deng et al. \cite{deng2024composerx} introduce ComposerX, an agent-based symbolic music generation framework designed to enhance the music composition capabilities of Large Language Models (LLMs). While LLMs have demonstrated impressive performance in STEM domains, they often struggle with music composition, particularly when dealing with long dependencies and harmony constraints. Even when equipped with advanced techniques like In-Context Learning and Chain-of-Thought, LLMs typically generate poorly structured music. ComposerX aims to address this by leveraging the reasoning abilities of LLMs and their extensive knowledge of music history and theory. By employing a multi-agent approach, the framework significantly enhances the music composition quality of GPT-4. The results show that ComposerX is capable of generating coherent, polyphonic music compositions with engaging melodies that follow user instructions, marking a substantial improvement in the application of LLMs to creative music composition tasks.

\paragraph{Music Understanding \& Generation Agents}
Yu et al. \cite{yu2023musicagent} present MusicAgent, a system designed to streamline AI-powered music processing by organizing and integrating diverse music-related tasks. Music processing spans a wide range of activities, from generation tasks like timbre synthesis to comprehension tasks like music classification. However, developers and amateurs often struggle to navigate the complexity of these tasks, particularly due to the varying representations of music data and the applicability of different models across platforms. MusicAgent addresses this challenge by offering an integrated solution that simplifies the process for users. The system includes a comprehensive toolset that gathers music tools from diverse sources such as Hugging Face, GitHub, and Web APIs. Additionally, it incorporates an autonomous workflow powered by Large Language Models (LLMs), like ChatGPT, which organizes these tools and automatically decomposes user requests into sub-tasks, invoking the appropriate tools. The primary goal of MusicAgent is to alleviate users from the technicalities of using AI-based music tools, allowing them to focus on the creative aspects of music.

\paragraph{Poetry Generation Agents}
Zhang et al. \cite{zhang2024llm} introduces a framework for enhancing the diversity and novelty of poetry generated by Large Language Models (LLMs) by incorporating social learning principles. While LLMs have made significant strides in automatic poetry generation, their outputs often lack the diversity and creativity seen in human-generated poetry. The proposed framework emphasizes both cooperative and non-cooperative interactions among multiple agents to foster diversity in generated poetry. This is the first attempt to apply multi-agent systems in non-cooperative environments for poetry generation, utilizing both TRAINING-BASED agents (GPT-2) and PROMPTING-BASED agents (GPT-3 and GPT-4). Experiments based on 96k generated poems show significant improvements, particularly for TRAINING-BASED agents, with a 3.0–3.7 percentage point increase in diversity and a 5.6–11.3 percentage point increase in novelty, as measured by distinct and novel n-grams. The results also reveal that poetry generated by these agents shows increased divergence in terms of lexicons, styles, and semantics. For PROMPTING-BASED agents, the non-cooperative environment helps enhance diversity, with an increase of 7.0–17.5 percentage points, though these agents showed a decrease in lexical diversity over time and did not exhibit the desired group-based divergence.

\paragraph{Lyric Generation Agents}
Liu et al. \cite{liu2024agent} address the challenges of melody-to-lyric generation by leveraging Generative Large Language Models (LLMs) and multi-agent systems. Previous research in this area has been constrained by limited high-quality aligned data and unclear standards for creativity. Many studies focused on broad themes or emotions, which have limited value given the advanced capabilities of current language models. In tonal contour languages like Mandarin, where pitch contours are influenced by both melody and tone, achieving a good fit between lyrics and melody becomes more complex. The study, validated by the Mpop600 dataset, demonstrates that lyricists and melody writers carefully consider this fit during their composition process. To tackle this, the authors developed a multi-agent system that decomposes the melody-to-lyric task into specific sub-tasks, with individual agents managing aspects such as rhyme, syllable count, lyric-melody alignment, and consistency. The quality of the generated lyrics was evaluated through listening tests using a diffusion-based singing voice synthesizer, assessing how different agent groups performed in terms of lyric creation. This work introduces a more structured approach to melody-to-lyric generation, offering a deeper understanding of the interaction between melody and lyrics in tonal languages.

\begin{figure*}[t]
    \centering
    \includegraphics[width=1\linewidth]{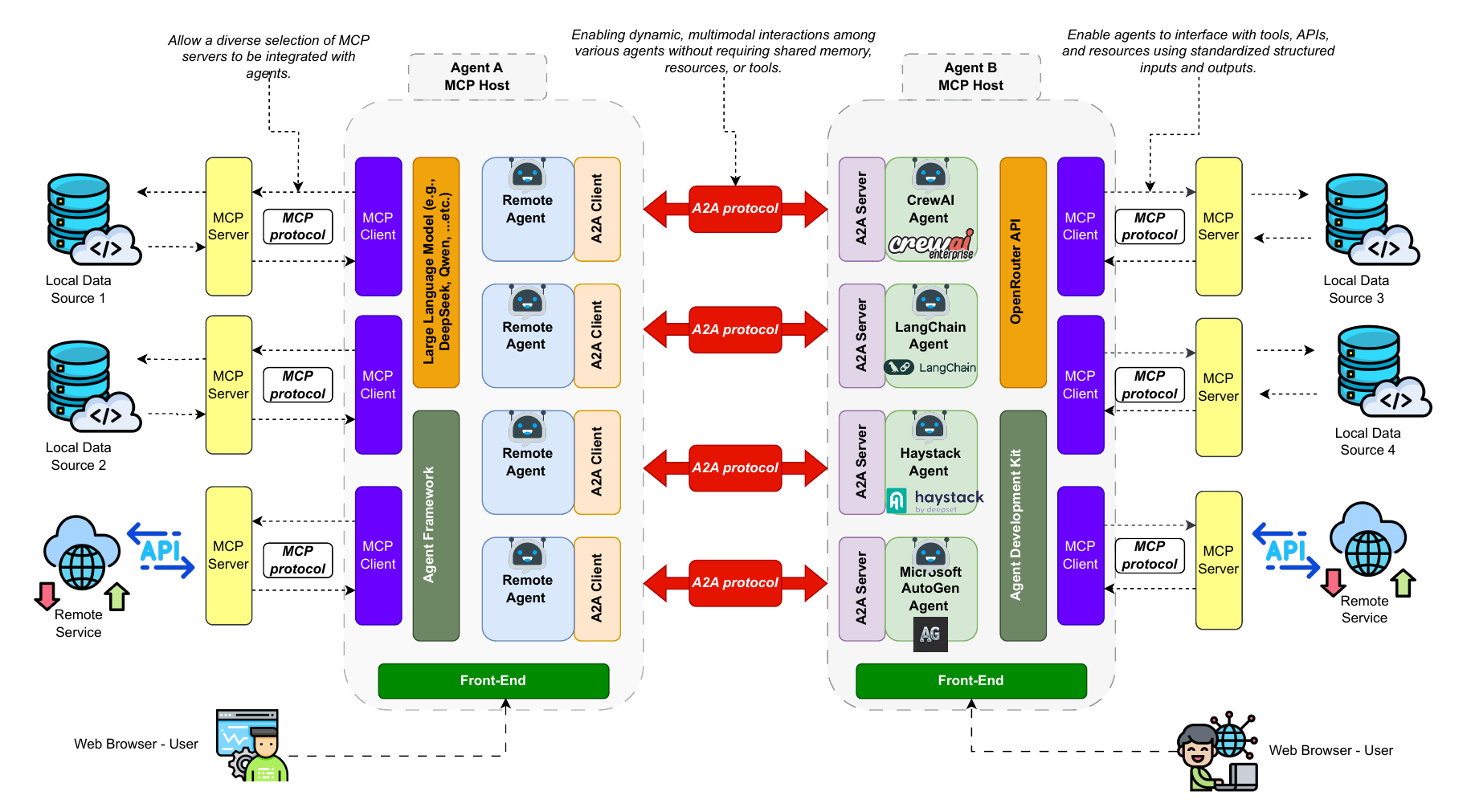}
    \caption{Multi-Agent Integration Framework: Enabling dynamic collaboration through the A2A and MCP Protocols.}
    \label{fig:protocols}
\end{figure*}

\begin{table*}[ht]
\centering
\caption{Comparison of MCP, ACP, and A2A Protocols}
\label{tab:protocolscomp}
\begin{tabular}{lp{4cm}p{5cm}p{4cm}} 
\hline
\textbf{Feature}                       & \textbf{MCP (Model Context Protocol) \cite{modelcontextprotocol2025}}                             & \textbf{ACP (Agent Communication Protocol) \cite{acp2025}}                          & \textbf{A2A (Agent-to-Agent Protocol) \cite{googleblog2025}}                                \\ 
\hline
Main Purpose                   & Facilitates access to context and data for LLMs                   & Enables communication between multiple agents within BeeAI           & Facilitates communication and task-sharing between agents across frameworks \\ 
\hline
Common Setup                   & Distributed servers providing specific data, connected via an MCP hub  & BeeAI Server coordinates and manages multiple agents within a local environment & Agents from different frameworks discover and connect through HTTP interfaces \\ 
\hline
Key Capabilities               & Standardized interface for connecting data and services to LLMs   & Simplifies agent deployment, discovery, and offers deep telemetry within BeeAI & Allows agents to discover each other’s capabilities and share tasks with updates \\ 
\hline
Typical Application            & Managing context for LLMs and integrating data streams           & Managing multiple agents within BeeAI’s environment                  & Enabling interaction and collaboration between agents from diverse systems \\ 
\hline
Core Objective                 & Uniformly managing how LLMs receive context and external tools    & Standardizing communication between BeeAI agents and external systems & Creating a standardized method for agents from different systems to communicate and collaborate \\ 
\hline
Architecture                   & Client-server model where LLMs hook into servers for data and tools  & BeeAI Server orchestrates the interaction of local agents and integrates external frameworks & Agents connect through agent cards and HTTP for task execution and communication \\ 
\hline
Key Differences                & Focuses on integrating tools and data into a single LLM process  & Primarily focused on internal coordination of agents within BeeAI    & Aims at linking agents across different ecosystems to collaborate effectively \\ 
\hline
Ideal Usage Scenario          & Integrating multiple data sources or services into an LLM workflow & Running and managing various agents within BeeAI’s environment       & Connecting agents from different platforms to enable collaboration and task-sharing \\ 
\hline
Common Use Cases              & Implementing controlled, secure LLM workflows with external data  & Orchestrating multi-agent environments with BeeAI’s platform         & Enabling task sharing and agent communication across different vendor systems \\ 
\hline
\textcolor{black}{Message Structure}
    & \textcolor{black}{JSON-RPC 2.0 requests/responses with fields \texttt{"jsonrpc"}, \texttt{"id"}, \texttt{"method"}, \texttt{"params"}, and \texttt{"result"}/\texttt{"error"}} 
    & \textcolor{black}{Custom JSON objects with top‐level \texttt{"role"} and an array of \texttt{"parts"} supporting MIME types, content URLs, encoding, and metadata}
    & \textcolor{black}{JSON-RPC extension in TypeScript with \texttt{id}, \texttt{method}, \texttt{params}, \texttt{result}/\texttt{error} and optional SSE streaming} \\ 
\hline
\textcolor{black}{Context Sharing}
    & \textcolor{black}{Host‐mediated, stateful sessions aggregating full conversation and forwarding only declared context fragments}
    & \textcolor{black}{Stateful sessions via a \texttt{session\_id}; full history is prepended on each call when the same ID is reused}
    & \textcolor{black}{Task ID grouping in both \texttt{id} and \texttt{params.id}; context streamed via Server-Sent Events with resubscription support} \\ 
\hline
\textcolor{black}{Failure Recovery}
    & \textcolor{black}{Standard JSON-RPC error objects plus built‐in cancellation and structured error‐reporting utilities}
    & \textcolor{black}{Structured error payloads (\texttt{"code"}, \texttt{"message"}) delivered via HTTP bodies, run results, or stream events; clients implement retry/fallback}
    & \textcolor{black}{Standard JSON-RPC error codes (e.g.\ -32700, -32001) and explicit \texttt{tasks/resubscribe} method to recover dropped streams} \\ 
\hline
\end{tabular}
\end{table*}

\subsection{AI Agents Protocols}

Recent advances in autonomous AI systems have underscored the importance of standardized communication protocols in facilitating seamless interaction among agents, tools, and external services. In this subsection, we present three prominent protocols developed between 2024 and 2025: the Agent Communication Protocol (ACP), the Model Context Protocol (MCP), and the Agent-to-Agent Protocol (A2A). 

\subsubsection{Agent Communication Protocol (ACP)}

In 2025, IBM Research proposed the agent-to-agent communication protocol named ACP, which is central to the operation of BeeAI\footnote{https://github.com/i-am-bee/beeai-framework}, an experimental platform designed to streamline the orchestration and execution of open-source AI agents, regardless of their underlying framework or code base. The primary goal of ACP is to standardize communication between agents, addressing challenges posed by inconsistent interfaces and enabling seamless interaction across diverse agents and client systems. Inspired by Anthropic’s MCP, ACP initially aimed to connect agents to data and tools but has since evolved to include advanced features such as discovery, delegation, and multi-agent orchestration. Core components of BeeAI include the BeeAI Server, which orchestrates agent processes in a local-first environment and provides a unified REST endpoint for external apps and UIs, and the ACP SDKs, which offer libraries in Python and TypeScript, along with a command-line interface and UI for easy agent discovery and launch \cite{acp2025}.

\textcolor{black}{ACP messages are custom JSON objects containing a top-level \texttt{"role"} field and a \texttt{"parts"} array, where each part encodes MIME-type metadata along with either inline content or content URLs, optional encoding hints, and arbitrary metadata fields. Context is maintained via a \texttt{session\_id} issued on the first call; reusing that same ID causes the full turn history to be prepended to each request, enabling stateful conversational sessions. In the event of errors, ACP delivers structured payloads with \texttt{"code"} and \texttt{"message"} in either HTTP response bodies, within run-result envelopes, or as real-time stream events; clients are responsible for implementing retry or fallback logic.}

\subsubsection{Model Context Protocol (MCP)}
In late 2024, Anthropic introduced the Model Context Protocol (MCP), an open and flexible protocol that standardizes how AI systems interact with external tools and data sources, much like a USB-C port provides a universal connection for devices. Inspired by the Language Server Protocol, MCP enables AI agents to autonomously identify, select, and manage a wide range of services based on the specific context of each task. The protocol facilitates the development of complex workflows by offering a growing catalog of pre-built integrations, the flexibility to switch between different LLM providers, and best practices for securing data within an organization's infrastructure \cite{hou2025model}.  

An expanding ecosystem of servers highlights the protocol’s potential. For example, official reference servers demonstrate MCP’s core capabilities through secure file management and database access, utilizing PostgreSQL, SQLite, and Google Drive. At the same time, development environments benefit from integration with tools such as Git, GitHub, and GitLab. Moreover, MCP supports productivity and communication enhancements via integrations with platforms like Slack and Google Maps and even extends to specialized AI tools, including image generators and sophisticated search APIs\footnote{https://github.com/modelcontextprotocol/servers}.

MCP is designed around a client-server architecture in which host applications connect to multiple lightweight servers \cite{modelcontextprotocol2025}. This allows secure access to local data sources such as files and databases and remote services available through web APIs. By unifying these interfaces, MCP transforms everyday platforms into versatile, multi-modal AI agents, simplifying the creation of AI-native applications and accelerating innovation across diverse domains.

\textcolor{black}{MCP messages conform to the JSON-RPC 2.0 standard, with each request carrying the fields  
\texttt{"jsonrpc":"2.0"}, \texttt{"id"}, \texttt{"method"} and an optional \texttt{"params"} object, and each response echoing the same \texttt{"id"} alongside either a \texttt{"result"} or an \texttt{"error"}. Context sharing is host-mediated and stateful: the MCP hub aggregates the full conversation history. It selectively forwards only those context fragments that each server has declared it can consume, ensuring isolation between adapters. For failure recovery, MCP builds on JSON-RPC’s native error objects by providing built-in cancellation requests and structured error-reporting utilities, allowing clients to cancel operations or receive detailed diagnostics when calls fail.}

\subsubsection{Agent-to-Agent Protocol (A2A)}

In 2025, Google introduced the Agent2Agent (A2A) protocol to usher in a new era of seamless interoperability among AI agents, significantly enhancing workplace productivity and automation \cite{googleblog2025}. The protocol is designed to facilitate dynamic collaboration between autonomous agents, enabling them to work together across isolated data systems and diverse applications regardless of their underlying frameworks or vendors. Using familiar standards such as HTTP, SSE, and JSON-RPC, A2A simplifies integration with existing IT infrastructures while also ensuring robust enterprise-grade security through proven authentication and authorization practices. A2A supports both swift and long-duration tasks by allowing agents to exchange real-time updates, negotiate user interface requirements, and perform capability discovery via structured "Agent Cards.

\textcolor{black}{A2A extends JSON-RPC 2.0 with a TypeScript interface that includes optional fields \texttt{"jsonrpc"}, \texttt{"id"}, \texttt{"method"}, \texttt{"params"}, \texttt{"result"}, and \texttt{"error"}. A shared task ID groups interactions passed both as the JSON-RPC \texttt{"id"} and within \texttt{"params.id"}, and agents consume context updates by subscribing to Server-Sent Events streams (e.g.\ via \texttt{tasks/sendSubscribe}). For failure recovery, A2A defines standard JSON-RPC error codes (such as \(-32700\) for parse errors) and an explicit \texttt{tasks/resubscribe} method, enabling clients to reattach to ongoing task streams and resume processing seamlessly.}

MCP is designed to connect agents with tools, APIs, and resources through structured inputs and outputs. It is fully supported by Google’s ADK, which enables a wide range of MCP servers to be seamlessly integrated with AI agents. In parallel, A2A \footnote{https://google.github.io/A2A/} provides a dynamic, multimodal framework for agent-to-agent communication, allowing different agents to collaborate without sharing memory, resources, or tools. Fig. \ref{fig:protocols} presents a sophisticated multi-agent integration framework that leverages two key protocols, A2A and MCP, to enable seamless interactions among diverse agents and services. It depicts multiple remote agents, including those branded as CrewAI Agent, LangChain Agent, Haystack Agent, and Microsoft AutoGen, which communicate via the A2A protocol. This communication method allows agents to collaborate dynamically without sharing internal memories, resources, or tools, ensuring secure and efficient inter-agent exchanges. In parallel, the framework utilizes the MCP protocol to standardize interactions with various tools, APIs, and resources, enabling agents to connect with both local data sources and remote services through structured inputs and outputs.

Tab. \ref{tab:protocolscomp} provides a comparative analysis of three agent communication protocols: MCP, ACP, and A2A. It highlights their primary purpose, typical setup, core features, and ideal use cases. MCP (Model Context Protocol) focuses on integrating data and tools into LLM workflows, providing a standardized interface for delivering context. ACP (Agent Communication Protocol), a component of the BeeAI platform, enables communication among multiple agents in a local-first setup, providing tools for agent discovery and telemetry. In contrast, A2A (Agent-to-Agent Protocol) enables interoperability between agents across different frameworks, allowing them to exchange tasks and collaborate. The table highlights the distinct roles these protocols play in agent-based systems, with MCP focusing on data integration for LLMs, ACP concentrating on local agent orchestration, and A2A facilitating cross-platform collaboration among agents.

\subsection{Training datasets}

High-quality training datasets are crucial for enhancing the reasoning, multilingual understanding, and instruction-following abilities of large language models. In this subsection, we present three recently developed datasets: NaturalReasoning, FineWeb2, and MagPie-Ultra. Each dataset addresses different aspects of model training, ranging from expanding reasoning across multiple domains to enhancing multilingual capabilities and advancing the generation of synthetic instructions. 

\subsubsection{NaturalReasoning dataset}
Scaling reasoning capabilities beyond traditional domains such as math and coding has been challenging due to the scarcity of diverse, high-quality questions. In response, \cite{yuan2025naturalreasoning} introduces NaturalReasoning a comprehensive dataset comprising 2.8 million questions that span multiple domains, including STEM fields (like Physics and Computer Science), Economics, and Social Sciences, complete with reference answers. The dataset is designed not only to serve as a resource for knowledge distillation experiments, where it effectively transfers reasoning capabilities from a strong teacher model, but also for unsupervised self-training using external reward models. When training the Llama3.1-8B-Instruct model, NaturalReasoning demonstrates superior scaling effects, achieving notably higher average performance on benchmarks such as MATH, GPQA, and MMLU-Pro compared to other datasets. This work highlights the potential of a large, diverse question dataset to expand the boundaries of LLM reasoning across a broader range of fields.

\subsubsection{FineWeb2 dataset}

Hugging Face's team introduced \cite{penedo2024fineweb2} FineWeb2, a groundbreaking multilingual dataset comprising 8TB of meticulously cleaned text data with over 3 trillion non-English words drawn from more than 1,000 languages. FineWeb2 supports a total of 1,893 languages, with substantial coverage of 486 languages, including more than 1MB of data, and 80 languages boast over 1GB each, demonstrating its extensive linguistic diversity. Built upon 96 snapshots of CommonCrawl data spanning 2013 to 2024 and processed using the "datatrove" alongside sophisticated filtering code and configurations, FineWeb2 employs innovative techniques such as "re-hydration," deduplication, and language-specific filtering to ensure high data quality. Extensive ablation experiments, conducted with a 1.45 billion-parameter model trained on 30 billion tokens, further validate the dataset’s robustness. In comparative evaluations against established datasets like CC-100, mC4, CulturaX, and HPLT, FineWeb2 consistently outperforms across diverse languages. Additionally, specialized evaluations using the FineTasks benchmark on 9 varied languages underscore its potential for advancing multilingual natural language processing and retrieval-augmented generation applications.

\subsubsection{MagPie-Ultra dataset}

MagPie-Ultra \cite{argilla} is a synthetic dataset generated using Meta Llama 3.1 405 B-Instruct FP8, representing the first open dataset of its kind. It comprises 50,000 synthetic instruction pairs, created by prompting the language model with minimal "empty" prompts (only initial special tokens) that allow it to generate both user queries and corresponding responses auto-regressively. These generated pairs, filtered according to the MagPie recipe and refined via Argilla distilabel, cover a diverse range of challenging tasks, including coding, mathematics, data analysis, creative writing, advice seeking, and brainstorming. In addition to the raw instruction pairs, the dataset includes detailed metadata quality and difficulty scores, embeddings, topic labels, and safety assessments from tools like ArmorRM and LlamaGuard, which further support its use in training and evaluating large language models across complex instruction-following scenarios.

\subsection{\textcolor{black}{Human-in-the-Loop AI Agents}}

\textcolor{black}{Human-in-the-loop (HITL) interactions provide essential safeguards for autonomous agents, particularly in business-critical scenarios where incorrect tool selection or parameter misinterpretation can lead to unintended side effects. HITL mechanisms maintain human oversight throughout the orchestration of an agent by interposing confirmation steps between the decision-making of the model and the execution of sensitive operations.}

\textcolor{black}{Recently, the Amazon team proposed Amazon Bedrock Agents \cite{perrot2025implement}, which supports two primary HITL frameworks:}

\begin{itemize}

\item \textcolor{black}{User Confirmation. This mechanism pauses the orchestration to expose the intended function call and parameter values to the end user for approval. Non-critical read operations—such as retrieving PTO balances—execute automatically, while write operations—such as creating, updating, or canceling a PTO request—are gated by a Boolean confirm/reject prompt. Upon confirmation, the agent proceeds; upon rejection, the workflow either aborts or generates an alternative plan. Developers can configure at a per-action level which tools require confirmation, striking a balance between automation and safety.}

\item \textcolor{black}{Return of Control (ROC). ROC elevates human involvement by returning entire action groups to the application layer, allowing users not only to approve but also to edit parameters or provide additional context before execution. Configured at the action-group level, ROC enables more nuanced workflows—such as adjusting PTO dates via an interactive form—where a simple yes/no decision is insufficient. Final validation and execution are performed by the application’s API, ensuring compliance with business rules and regulatory requirements.}
\end{itemize}

\textcolor{black}{By integrating these HITL patterns, developers can harness the efficiency of autonomous agents while preserving accuracy, accountability, and user trust.}

\subsection{\textcolor{black}{Collaborative LLM frameworks}}

\textcolor{black}{Wang et al. \cite{wang2025talk} proposed TalkHier, which is a collaborative framework that integrates a rigorously structured messaging protocol with a hierarchical refinement process to improve inter-agent communication. It defines three message types, raw input, intermediate output, and background context, to organize information flow, improving clarity, and reducing misunderstandings common in unstructured exchanges. The hierarchical refinement mechanism allows agents to iteratively improve their contributions at successive levels, balancing diverse perspectives and mitigating biases introduced by feedback ordering. Experimental results on complex subdomain problem solving, open question answering, and Japanese text generation show that TalkHier outperforms current approaches and that each protocol element and refinement layer contributes meaningfully to its overall success.}

\textcolor{black}{Chen et al. \cite{chen2024comm} tackles the challenge of using large language models for complex scientific problem solving by introducing CoMM, a collaborative multi-agent, multi-reasoning-path prompting framework. In CoMM, each agent assumes a distinct expert role and follows its reasoning trajectory, which together foster a richer exploration of the problem space than single-chain prompts allow. By distributing few-shot exemplars across these role-play agents, the framework balances diverse analytical approaches and mitigates biases that arise when feedback is processed sequentially. Evaluation of two undergraduate-level science questions shows that CoMM not only exceeds competitive baselines, but also that independently prompting agents as different specialists is critical to achieving its improved performance.}

\section{Challenges and Open problems}\label{sec:5}

As the capabilities of AI agents and large language models continue to grow, new challenges and open problems emerge that limit their effectiveness, reliability, and security \cite{costello2025thinkprunetrainimprove}. In this section, we highlight several critical research directions, including advancing the reasoning abilities of AI agents, understanding the failure modes of multi-agent systems, supporting automated scientific discovery, enabling dynamic tool integration, reinforcing autonomous search capabilities, and addressing the vulnerabilities of emerging communication protocols.
\subsection{AI Agents Reasoning}

The primary challenge addressed in \cite{xiang2025towards} is the inherent limitation of traditional Chain-of-Thought (CoT) methods, which only reveal the final reasoning steps without explicitly modeling the underlying cognitive process that leads to those steps. Meta Chain-of-Thought (Meta-CoT) aims to fill this gap by capturing and formalizing the latent reasoning that underlies a Chain-of-Thought (CoT). This involves not only generating the visible chain of thought but also understanding the in-context search behavior and iterative reasoning steps that contribute to it. To overcome these challenges, the authors explore innovative approaches, including process supervision, synthetic data generation, and search algorithms, to produce robust Meta-CoTs. Moreover, they propose a concrete training pipeline that integrates instruction tuning with linearized search traces and reinforcement learning post-training. Open research questions remain regarding scaling laws, the role of verifiers, and the discovery of novel reasoning algorithms, underscoring the complexity and potential of advancing more human-like reasoning in large language models.

\subsection{Why Do Multi-Agent LLM Systems Fail?}
Pan et al. \cite{pan2025multiagent} present a critical examination of why multi-agent LLM systems, despite the theoretical benefits of collaboration, continue to underperform compared to their single-agent counterparts. Through a rigorous study of five open-source frameworks across 150 tasks, the authors enlist expert human annotators to identify fourteen distinct failure modes ranging from ignoring task or role specifications and unnecessary repetition, to lapses in memory and flawed verification processes. These issues are systematically grouped into three categories: design and specification shortcomings, inter-agent misalignment, and challenges in task verification and termination. Moreover, the study explores interventions such as refining agent role definitions and orchestration strategies, but finds that these measures alone are insufficient; thereby, it outlines a clear roadmap for future research to address the intricate challenges inherent in multi-agent coordination.

\subsection{AI Agents in Automated Scientific Discovery}

Liu et al. \cite{liu2025researchbenchbenchmarkingllmsscientific} introduce a large-scale benchmark for evaluating the capability of large language models (LLMs) in generating high-quality scientific research hypotheses. It tackles this gap by focusing on three pivotal sub-tasks: inspiration retrieval, hypothesis composition, and hypothesis ranking. The authors have developed an automated framework that extracts key components from scientific papers, including research questions, background surveys, inspirations, and hypotheses, across 12 disciplines. Expert validation ensures the reliability of this framework. By exclusively using papers published in 2024, the study minimizes data contamination from large language model (LLM) pretraining datasets, revealing that LLMs perform notably well in retrieving novel inspirations. This positions LLMs as promising “research hypothesis mines” that can facilitate the automation of scientific discovery by generating innovative hypotheses at scale.

Despite these advances, significant challenges remain for AI agents employing LLMs to automate scientific discovery. One key obstacle is ensuring that these agents generate novel and scientifically valid hypotheses, as they must navigate the risk of producing biased or spurious associations without sufficient human oversight. Furthermore, the complexity and diversity of scientific literature across various disciplines demand that these agents not only understand domain-specific nuances but also adapt dynamically to evolving research contexts. The risk of data contamination, particularly when recent publications might overlap with pretraining data, further complicates the extraction of truly innovative insights. In addition, scaling these systems while preserving transparency, interpretability, and ethical standards poses a multifaceted challenge that must be addressed to harness the potential of AI-driven scientific discovery fully.

\subsection{Dynamic Tool Integration for Autonomous AI Agents}

Wu et al. \cite{wu2025chain} introduce Chain-of-Tools, a novel tool learning approach that leverages the robust semantic representation capabilities of frozen large language models (LLMs) to perform tool calling as part of a chain-of-thought reasoning process. By utilizing a vast and flexible tool pool that can include previously unseen tools, this method addresses the inefficiencies and highlights key challenges, including managing vast prompt-based demonstrations. The authors validate their approach on a range of datasets, including a newly constructed dataset, SimpleToolQuestions, as well as GSM8K-XL, FuncQA, and KAMEL, demonstrating that Chain-of-Tools outperforms conventional baselines. Additionally, the method holds promise for enhancing autonomous AI agents by enabling them to select and utilize external tools dynamically, thereby broadening their capability to solve complex, multi-step tasks independently. This work prompts several questions: How can the integration of unseen tools further enhance LLM adaptability in diverse scenarios? What critical dimensions of the model output influence effective tool selection, and how can they be optimized for greater interpretability? Moreover, how might this methodology be extended to enable more robust autonomous decision-making in AI agents facing increasingly complex reasoning challenges? Notably, these questions also underscore key challenges such as managing a huge tool pool, ensuring efficient tool selection, enhancing model interpretability, and integrating autonomous AI agents capable of dynamic, independent operation.

\subsection{Empowering LLM Agents with Integrated Search via Reinforcement Learning}

ReSearch \cite{chen2025learning} represents a significant step toward endowing LLM-based agents with the ability to decide autonomously when and how to consult external knowledge sources, seamlessly weaving search operations into their reasoning chains via reinforcement learning. By framing search as an actionable tokenized operation rather than a separate retrieval step ReSearch trains models like Qwen2.5 through a reward signal that emphasizes final-answer accuracy and adherence to a structured think/search/result format. This paradigm eliminates the need for painstakingly annotated reasoning traces and yields strong multi-hop question–answering performance and cross-domain generalization. Yet, several challenges remain for deploying such agents in the wild: how to scale the approach to richer, real-time toolsets (e.g., calculators, databases, code execution environments) without blowing up action spaces; how to design more nuanced reward functions that capture partial credit for intermediate reasoning or mitigate reward hacking; how to ensure robustness and interpretability when agents autonomously interleave reasoning and tool use; and how to balance exploration of novel tool sequences against exploitation of known effective patterns. Addressing these questions will be crucial for realizing truly versatile, trustworthy LLM agents capable of complex, multi-step problem-solving.

\subsection{Vulnerabilities of AI Agents Protocols}

MCP protocol standardizes how AI applications provide context to LLMs. The MCP protocol faces critical vulnerabilities in Agent AI communications due to its fundamentally decentralized design \cite{hou2025model}. Without a central authority overseeing security, disparate implementation practices can lead to uneven defenses, making it easier for attackers to exploit weak links. In particular, the absence of a standardized authentication mechanism across different nodes hinders reliable identity verification, thereby increasing the risk of unauthorized access and potential data breaches. Moreover, deficiencies in robust logging and debugging tools further complicate the timely detection of anomalies and errors, which is vital for preventing and mitigating attacks. Additionally, the complexity inherent in managing multi-step, distributed workflows can lead to state inconsistencies and operational glitches, amplifying the potential impact of a security compromise across interconnected systems.

\section{Conclusion}
\label{sec:6}

\textcolor{black}{In this paper, we have surveyed recent advances in large language model reasoning and autonomous AI agents, demonstrating that multi-step, intermediate processing, exposed through models such as DeepSeek-R1, OpenAI o1 and o3, and GPT-4o, yields marked improvements in accuracy and reliability across complex tasks in mathematics, code generation and logical reasoning. We examined a range of training and inference strategies, including inference-time scaling, pure reinforcement learning (for example, DeepSeek-R1-Zero), supervised fine-tuning combined with reinforcement learning, and distillation-based fine-tuning. We demonstrated that hybrids of these methods, applied to Qwen-32B and Llama-based architectures, foster emergent reasoning behaviors while curbing overthinking and verbosity. Our unified comparison of some 60 benchmarks from 2019 to 2025, organized into a structured taxonomy, highlights the impact of mixture-of-experts, retrieval-augmented generation, and reinforcement learning frameworks, as well as key architectural enhancements, on model performance. We also reviewed AI agent frameworks introduced between 2023 and 2025, illustrating their applications in materials science, biomedical research, synthetic data generation, and financial forecasting. Despite these successes, significant challenges remain, most notably automating multi-step reasoning without human oversight, balancing structured guidance with model flexibility, and integrating long-context retrieval at scale. We anticipate that future work will increasingly focus on domain- and application-specific optimizations, as early systems such as DeepSeek-R1-Distill, Sky-T1, and TinyZero begin to demonstrate how specialized reasoning systems can achieve optimal trade-offs between computational cost and accuracy.}

\bibliographystyle{IEEEtran}
\bibliography{bibliography} 

\end{document}